\documentclass{article}

\usepackage{arxiv}

\usepackage[utf8]{inputenc} 
\usepackage[T1]{fontenc}    
\usepackage{hyperref}       
\usepackage{url}            
\usepackage{booktabs}       
\usepackage{amsfonts}       
\usepackage{nicefrac}       
\usepackage{microtype}      
\usepackage{lipsum}		
\usepackage{graphicx}
\usepackage{natbib}
\usepackage{doi}

\usepackage{pifont}

\usepackage{forest}
\usetikzlibrary{matrix}

\usepackage{amsmath}
\usepackage{algorithm}
\usepackage{algorithmicx}
\usepackage{algpseudocode}
\usepackage{mathrsfs}
\usepackage{multirow}
\usepackage{amsthm}
\usepackage{bm}
\usepackage{booktabs}
\usepackage{scalerel}
\usepackage{makecell}

\usepackage{hyperref}

\usepackage{caption}
\usepackage{subcaption}

\usepackage{bbm}
\usepackage{url}

\title{An Interdisciplinary and Cross-Task Review on Missing Data Imputation}

\author{Jicong Fan \\
	School of Data Science\\
	The Chinese University of Hong Kong, Shenzhen\\
	\texttt{fanjicong@cuhk.edu.cn} \\
}

\renewcommand{\shorttitle}{An Interdisciplinary and Cross-Task Review on Missing Data Imputation}

\begin{document}

\maketitle

\begin{abstract}

Missing data represents a fundamental and pervasive challenge in modern data science, significantly impeding analytical capabilities and decision-making processes across an exceptionally broad spectrum of disciplines including healthcare, bioinformatics, social science, e-commerce, and industrial monitoring systems. Despite decades of research and the development of numerous imputation methodologies, existing literature remains fragmented across disciplinary boundaries, creating a critical need for a comprehensive, interdisciplinary synthesis that bridges statistical foundations with contemporary machine learning advances. 
This work systematically covers fundamental concepts—including missingness mechanisms, single vs. multiple imputation, and varying imputation goals—and explores problem characteristics across different domains. The review extensively categorizes imputation methods, spanning classical techniques (e.g., regression, EM algorithm) to modern approaches such as low-rank and high-rank matrix completion, deep learning models (autoencoders, GANs, diffusion models, graph neural networks), and large language models. Special consideration is given to methods tailored for complex data types including tensor data, time series, graph-structured data, categorical data, and multimodal data, acknowledging their unique challenges and solution approaches. Beyond methodological considerations, we investigate the crucial integration of imputation with downstream machine learning tasks including classification, clustering, and anomaly detection, examining both sequential pipelines and joint optimization frameworks. The review also assesses theoretical guarantees for various methods, available benchmarking resources, and comprehensive evaluation metrics. Finally, we identify critical challenges and future directions, emphasizing the complexities of model selection and hyperparameter optimization, the growing importance of privacy-preserving imputation through federated learning approaches, and the ambitious pursuit of generalizable or universal imputation models that can adapt across domains and data types, thereby providing a roadmap for advancing this vital field of research.

\end{abstract}


\tableofcontents

\section{Introduction}\label{sec:introduction}

Missing data is a prevalent challenge across numerous fields, including social science, e-commerce, healthcare, bioinformatics, and communications. For instance, in social science surveys, respondents may refuse to answer certain questions, leading to non-responses \citep{little1989analysis}. In bioinformatics, gene expression matrices often contain significant missing values, or `dropouts', which occur when expressed transcripts are not fully detected and are consequently recorded as zeros \citep{troyanskaya2001missing,huang2018saver}. Similarly, in sensor networks, node failures can result in incomplete data collection \citep{faizin2019review}. In e-commerce, the user-item interaction matrix is typically highly sparse, and predicting its missing values forms the basis of recommendation systems \citep{koren09}. Even in computer vision, tasks such as watermark and occlusion removal can be framed as missing data imputation problems.
Beyond imputation, two other primary approaches for handling missing data are deletion and ignorance. The deletion method involves discarding any data instances or features that contain missing values, proceeding with analysis only on the remaining complete dataset. A common implementation is to remove rows or columns with missing entries from a data matrix. A significant drawback of this approach is that it can drastically reduce the dataset size, potentially compromising the reliability of subsequent analysis. Moreover, it fails to leverage the partial information available in the incomplete observations.
The ignorance approach, conversely, overlooks the presence of missing values during analysis, often by replacing them with a simple placeholder like zero. This method presupposes that the analytical or learning algorithms are robust to such missingness. Intuitively, this strategy is only viable when the proportion of missing data is sufficiently small.

Research on handling missing data dates back to the 1930s \citep{wilks1932moments, yates1933analysis}. For instance, \citet{wilks1932moments} investigated the estimation of means, variances, and covariances for a bivariate normal population from incomplete samples, comparing two methods: maximum likelihood estimation and an approach based on systems of independent estimates. This foundational work was later expanded by \citep{lord1955estimation} and \citep{10.2307/2280845}, who addressed the more general problem of estimating parameters for multivariate normal populations with incomplete data. A key literature review on this topic in multivariate statistics was subsequently compiled by \citep{afifi1966missing}.
A significant theoretical advance was made by \citep{rubin1976inference}, who established that directly ignoring the missing data process is inappropriate for likelihood-based inference unless the data are Missing at Random. Around the same time, \citet{dempster1977maximum} introduced the Expectation-Maximization (EM) algorithm for obtaining maximum likelihood estimates from incomplete data, which has since become one of the most influential methods in the field. Parallel to these methodological developments, researchers such as \citet{kalton1982imputing}, \citet{little1988missing}, and \citet{little1989analysis} focused on missing data problems in the social sciences. Among them, \citet{little1989analysis} provided a detailed analysis of different missingness patterns, including general and special cases like univariate and monotone missingness. From the perspective of psychological research, \citet{roth1994missing} later reviewed classical methods for missing data imputation.

Since real-world data are often represented in matrix form, several scholars have proposed performing missing data imputation by leveraging the underlying structure of the matrix, a problem known as matrix completion \citep{CandesRecht2009, mazumder2010spectral}. A particularly important and useful structure is low-rankness, which arises from the fact that real data matrices often lie in a low-dimensional subspace \citep{udell2019big}. This low-rank prior enables the effective recovery of missing values. While numerous approaches to matrix completion exist, some researchers have extended the low-rank assumption to more general structures, such as unions of subspaces or manifolds, which can lead to high-rank or even full-rank matrices \citep{ErikssonBalzanoNowak2011, FANNLMC, pmlr-v70-ongie17a, fan2020polynomial}. For instance, \citet{fan2020polynomial} assumed that the data are drawn from a union of polynomials. Matrix completion methods have found applications in data pre-processing, image inpainting, collaborative filtering (and recommendation systems more broadly) \citep{koren09}, and link prediction \citep{menon2011link}.

Higher-order tensors, as natural extensions of matrices, are also ubiquitous in science and engineering. For example, a color image with three channels can be represented as a third-order tensor, and fMRI scans across multiple time points can form a fourth-order tensor. To impute missing values in such data, tensor completion has been extensively studied as an extension of matrix completion \citep{gandy2011tensor, acar2011scalable, liu2012tensor, qin2022low}. Tensor completion has applications in recommendation systems, image and video inpainting, and knowledge graph completion, among others. Compared to matrix completion, tensor completion typically involves higher time and space complexity, and the landscape of tensor decomposition models is more diverse. However, both theoretical and empirical studies have shown that tensor completion often significantly outperforms matrix completion applied to matricized tensors in terms of recovery accuracy.

Recently, deep learning \citep{lecun2015deep,goodfellow2016deep}, a prominent branch of machine learning, has been increasingly applied to missing data imputation. Deep learning typically refers to deep artificial neural networks, which are capable of approximating highly complex functions, as established by the universal approximation theorems \citep{pinkus1999approximation,sonoda2017neural,lu2017expressive}. This capacity enables deep learning to capture the intricate structures inherent in real-world data, leading to impressive performance in imputation tasks.
Perhaps the most well-known deep learning model for this purpose is the autoencoder \citep{hinton2006reducing}. In this architecture, the input typically consists of incomplete data, and the output is the reconstructed, complete data \citep{FAN2017540,gondara2018mida,8891716}.
Beyond autoencoders, the widely recognized generative adversarial network (GAN) \citep{goodfellow2014generative} has also been adapted for imputation \citep{yoon2018gain,li2019misgan,yoon2020gamin}. For instance, GAIN, introduced by \citet{yoon2018gain}, incorporates additional `hint' information into the discriminator to enable its operation on incomplete data. These hints guide the generator to produce imputations that adhere to the true underlying data distribution. Diffusion models \citep{sohl2015deep,ho2020denoising,yang2023diffusion} have also been leveraged for missing data imputation \citep{tashiro2021csdi}. More recently, the potential of Large Language Models (LLMs) for missing data imputation has been investigated. More details will be provided in Section \ref{sec_LLM}.

\begin{figure}
    \centering
    \includegraphics[width=0.9\linewidth]{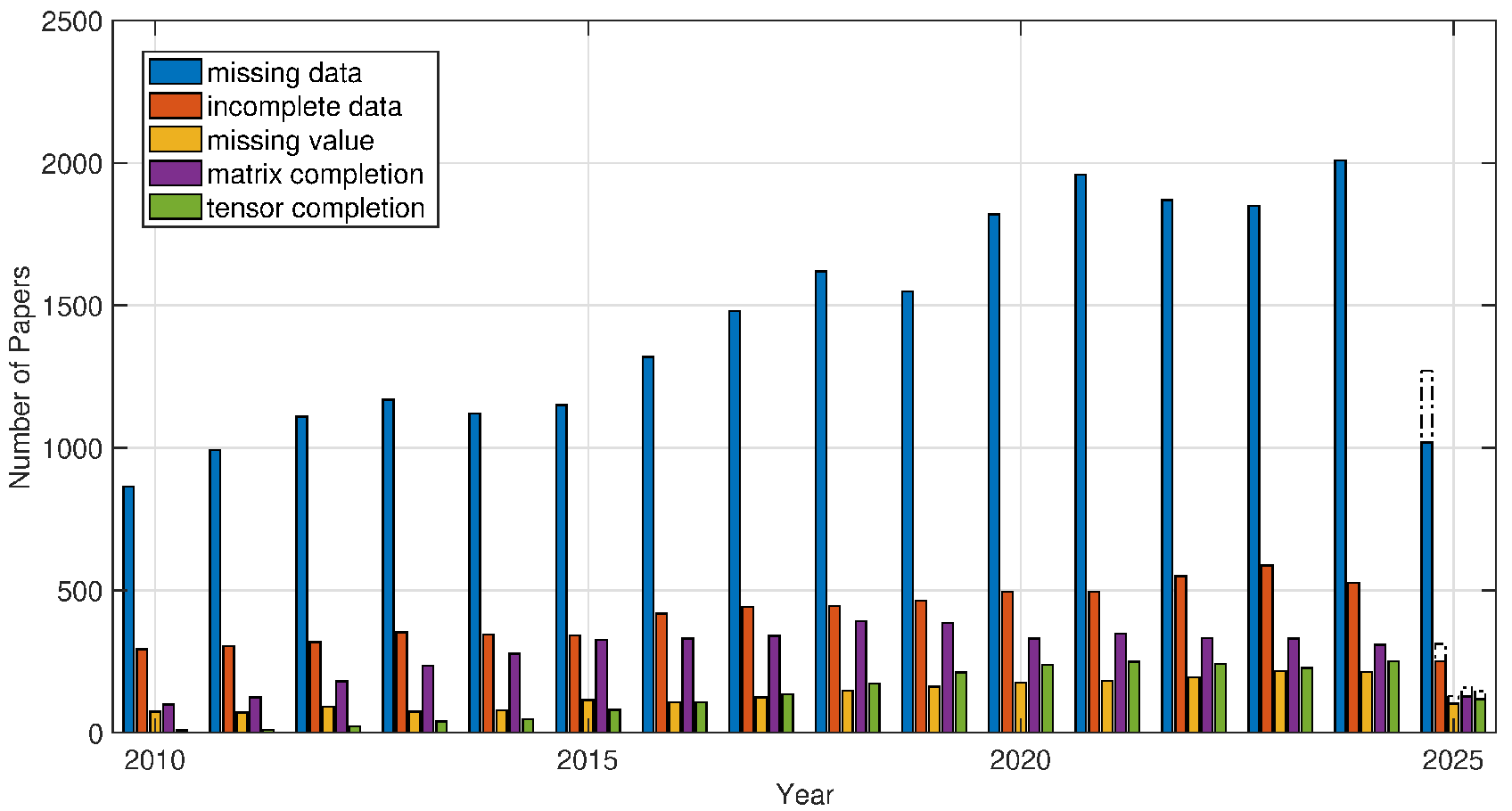}
    \caption{Number of publications on missing data (2010–2025) from Google Scholar, retrieved by searching for five specific keywords in paper titles. The dotted line for 2025 is based on a simple prediction according to the survey time of this work.}
    \label{fig_papercounts}
\end{figure}

Figure \ref{fig_papercounts} depicts the annual publication count in the field of missing data imputation between 2010 and 2025. The data, retrieved from Google Scholar, were obtained by querying five specific keywords in paper titles: `missing data', `incomplete data', `missing value', `matrix completion', and 'tensor completion' were used to retrieve the papers. The figure reveals a substantial and gradually growing body of literature, particularly during the 2010-2020 period. Given this scale, a comprehensive inclusion of all pertinent papers in this review is not practicable. However, it is both valuable and feasible to develop a systematic taxonomy of missing data problems and solutions by incorporating a sufficient number of representative studies. 

Missing data exhibit diverse characteristics across different fields, such as social science, bioinformatics, healthcare, and e-commerce. While imputation methods vary accordingly, they are often motivated by shared principles. Table \ref{tab_survey} summarizes relevant survey papers categorized by domain. As observed, existing reviews are often dated and lack a cross-disciplinary perspective. To address this gap, this paper begins by introducing fundamental concepts related to missing data in Section \ref{sec_pre}. Section \ref{sec_areas} then provides a comprehensive review of missing data problems and methodologies across different fields, aiming to help readers recognize the diversity of challenges and solutions. By facilitating cross-disciplinary insight, we hope to enable researchers in one area to draw inspiration from others and identify novel problems or methods. In Section \ref{sec_methods}, we categorize and introduce various imputation approaches—such as the EM algorithm, matrix and tensor completion, deep learning and LLM-based imputation methods, and time series imputation—covering state-of-the-art methodologies for diverse data types. Section \ref{sec_theory} discusses theoretical advances and guarantees for missing data imputation, while Section \ref{sec_benchmarks} reviews available benchmarks and evaluation metrics. Section \ref{sec_challenges} outlines key challenges and future directions. Finally, Section \ref{sec_conclude} concludes the paper. Figure \ref{fig_tree} shows a taxonomy of methods for handling missing data summarized in this paper.

\begin{figure}[t]
    \centering
\begin{forest}
for tree={
    grow'=east,
    draw=red!5,
    fill=red!5,
    thick,
    align=center,
    anchor=west,
    parent anchor=east,
    child anchor=west,
    minimum height=0.8cm,
    l sep=1.0cm,
    s sep=0.1cm,
    edge={red, thick, ->}
}
[Handling\\Missing\\Data
    [Imputation\\Methods
        [General \\Imputation\\ Methods
            [Simple Imputation]
            [Hot-Deck Imputation]
            [Regression Imputation]
            [Likelihood Imputation]
            [Matrix Completion
                [Low-Rank Matrix Completion]
                [High-Rank Matrix Completion]
            ]
            [Deep Imp.
                [Autoencoder]
                [Deep Factorization]
                [Generative Models
                    [VAE]
                    [GAN]
                    [Flow]
                    [Diffusion]
                ]
            ]
            [LLM Imp.]
            [Others] 
        ]
        [Imputation for\\Special Data Format
            [Tensor Completion]
            [Graph Completion]
            [Time Series Imputation]
            [Online Imputation for Streaming Data]
            [Categorical Data Imputation]
            [Multimodal Data Imputation]
        ]
    ]
    [Learning on\\Incomplete Data
        [Classification and Regression]
        [Clustering]
        [Anomaly/Outlier/Novelty Detection]
        [Others]
    ]
]
\end{forest}
    \caption{Taxonomy of methods for handling missing data}
    \label{fig_tree}
\end{figure}

\section{Related Work}
The period from 2000 to 2010 yielded several influential surveys on missing data. Specifically, \citet{pigott2001review} summarized common methods, including complete-case analysis, available-case analysis, single imputation, and model-based methods for multivariate normal data, as well as maximum likelihood via the EM algorithm. \citet{schafer2002missing} provided a detailed discussion of the Missing at Random assumption and evaluated the effectiveness of maximum likelihood and Bayesian multiple imputation. During this time, \citet{little2002statistical} also published a seminal book on the statistical analysis of missing data, the most recent edition of which was released in 2019. Meanwhile, \citet{donders2006gentle} offered a concise review that compared single and multiple imputation through numerical experiments. \citet{graham2009missing} presented a practical overview of the missing data literature, covering theoretical foundations and methods like normal-model multiple imputation and maximum likelihood, while also addressing practical issues such as using auxiliary variables to enhance power and reduce bias.
The subsequent decade saw a continuation of this review effort. In 2010, Enders C.K. authored an applied book on handling missing data, with a latest edition published in 2022 \citep{enders2022applied}. That same year, \citet{baraldi2010introduction} explored the theoretical foundations of missing data analysis, summarized conventional techniques, and provided accessible explanations of maximum likelihood and multiple imputation. 
After 2010, there are also a few review papers on missing data \citep{nakai2011review,eekhout2012missing,silva2014brief,shen2015missing,pratama2016review,gabrio2017handling,wu2019imputation}. For instance, 
\citet{pratama2016review} briefly reviewed a few methods for handling missing data in time series.

A number of recent reviews have explored imputation methods across various fields. \citet{zhang2022handling} surveyed methods for environmental monitoring, while \citet{sethia2023review} provided a brief overview that categorizes techniques into simple and multiple imputation. In a 2024 review of 46 studies on electronic health records, \citet{ren2024moving} concluded that while machine learning methods show significant promise, no single approach offers a universally generalizable solution. More recently, \citet{chourib2025missing} presented a short review that classified strategies into three principal categories: preprocessing techniques, graph-based imputation, and algorithms inherently tolerant to missing values, accompanied by a numerical evaluation of five methods.
It is worth noting that numerous other publications are titled as reviews or surveys on missing data, but most of them have a very narrow focus (e.g., \citep{moorthy2014review,faizin2019review,wu2019imputation,zhang2025missing}) and will not be discussed here.

\newcommand{\cmark}{\ding{51}}
\newcommand{\xmark}{\ding{55}}

\begin{table}[h!]
\caption{Comparison of review papers of missing data imputation (DL: deep learning; MC: matrix completion; GC: graph completion; TC: tensor completion; MD: multimodal data; DI: downstream integration)}\label{tab_survey}
\begin{tabular}{l|l|lllllllll} \hline
paper & area & DL &MC & GC& TC &MD & LLM & DI & theory\\ \hline
\citep{roth1994missing} &  psychology &  \xmark{}&\xmark{}&\xmark{}&\xmark{}&\xmark{}&\xmark{}&\xmark{}&\xmark{}\\
\citep{pigott2001review} & statistics & \xmark{}&\xmark{}&\xmark{}&\xmark{}&\xmark{}&\xmark{}&\xmark{}&\xmark{}\\
\citep{troyanskaya2001missing} & bioinformatics &\xmark{}& \xmark{}&\xmark{}&\xmark{}&\xmark{}&\xmark{}&\xmark{}&\xmark{}\\
\citep{schafer2002missing} & statistics & \xmark{}&\xmark{}&\xmark{}&\xmark{}&\xmark{}&\xmark{}&\xmark{}&\xmark{}\\
\citep{little2002statistical} & statistics & \xmark{}&\xmark{}&\xmark{}&\xmark{}&\xmark{}&\xmark{}&\xmark{}&\xmark{}\\
\citep{graham2009missing} & psychology & \xmark{}&\xmark{}&\xmark{}&\xmark{}&\xmark{}&\xmark{}&\xmark{}&\xmark{}\\
\citep{porter2012missing} & social science & \xmark{}&\xmark{}&\xmark{}&\xmark{}&\xmark{}&\xmark{}&\xmark{}&\xmark{}\\
\citep{cheema2014review} & education & \xmark{}&\xmark{}&\xmark{}&\xmark{}&\xmark{}&\xmark{}&\xmark{}&\xmark{}\\
\citep{silva2014brief} & data analytics & \xmark{}&\xmark{}&\xmark{}&\xmark{}&\xmark{}&\xmark{}&\xmark{}&\xmark{}\\
\citep{shen2015missing}& remote sensing & \xmark{}&\xmark{}&\xmark{}&\xmark{}&\xmark{}&\xmark{}&\xmark{}&\xmark{}\\
\citep{pratama2016review}& time series & \xmark{}&\xmark{}&\xmark{}&\xmark{}&\xmark{}&\xmark{}&\xmark{}&\xmark{}\\
\citep{du2020missing} & monitoring systems & \cmark{}&\cmark{}&\xmark{}&\cmark{}&\xmark{}&\xmark{}&\xmark{}&\xmark{}\\
\citep{emmanuel2021survey} & machine learning & \cmark{}&\xmark{}&\xmark{}&\xmark{}&\xmark{}&\xmark{}&\xmark{}&\xmark{}\\
\citep{thomas2021systematic} & machine learning & \cmark{}&\xmark{}&\xmark{}&\xmark{}&\xmark{}&\xmark{}&\xmark{}&\xmark{}\\
\citep{joel2022review} & machine learning & \cmark{}&\cmark{}&\xmark{}&\xmark{}&\xmark{}&\xmark{}&\xmark{}&\xmark{}\\
\citep{adhikari2022comprehensive} & internet of things & \cmark{}&\cmark{}&\xmark{}&\xmark{}&\xmark{}&\xmark{}&\xmark{}&\xmark{}\\
\citep{nijman2022missing}& medical & \xmark{}&\xmark{}&\xmark{}&\xmark{}&\xmark{}&\xmark{}&\xmark{}&\xmark{}\\
\citep{liu2022handling} & healthcare \& machine learning & \cmark{}&\cmark{}&\xmark{}&\xmark{}&\xmark{}&\xmark{}&\xmark{}&\xmark{}\\
\citep{sun2023deep} & machine learning & \cmark{}&\xmark{}&\xmark{}&\xmark{}&\xmark{}&\xmark{}&\xmark{}&\xmark{}\\
\citep{ren2024moving} & healthcare \& machine learning & \cmark{}&\xmark{}&\xmark{}&\xmark{}&\xmark{}&\xmark{}&\xmark{}&\xmark{}\\
\citep{alwateer2024missing} & machine learning & \cmark{}&\xmark{}&\xmark{}&\xmark{}&\xmark{}&\xmark{}&\xmark{}&\xmark{}\\
\citep{zhang2024comprehensive} & transportation & \cmark{}&\cmark{}&\xmark{}&\cmark{}&\xmark{}&\xmark{}&\xmark{}&\xmark{}\\
\citep{little2024missing} & psychology & \xmark{}&\xmark{}&\xmark{}&\xmark{}&\xmark{}&\xmark{}&\xmark{}&\xmark{}\\
\citep{LE2025100720} & healthcare & \cmark{}&\cmark{}&\xmark{}&\xmark{}&\xmark{}&\xmark{}&\xmark{}&\xmark{}\\
\citep{benhamza2025comprehensive} & healthcare & \cmark{}&\xmark{}&\xmark{}&\xmark{}&\xmark{}&\xmark{}&\xmark{}&\xmark{}\\ 
\citep{chourib2025missing} & healthcare & \cmark{}&\xmark{}&\xmark{}&\xmark{}&\xmark{}&\xmark{}&\xmark{}&\xmark{}\\ 
\midrule
Ours & interdiscipline & \cmark{}& \cmark{}& \cmark{}& \cmark{}& \cmark{}& \cmark{}&\cmark{}&\cmark{}\\
\hline
\end{tabular}
\end{table}

\textbf{Notation} Throughout this paper, we use $x$ or $X$ to denote a scalar, use $\mathbf{x}$ and $\mathbf{X}$ to denote vector and matrix respectively, use $\mathcal{X}$ to denote a set, and use $\boldsymbol{\mathcal{X}}$ to denote a tensor. Moreover, $\mathbf{x}$ may also denote a general data sample, corresponding to an image, a text, or even a multivariate time series. An incomplete $\mathbf{x}$ is denoted as $\tilde{\mathbf{x}}$, while the imputed $\tilde{\mathbf{x}}$ is denoted as $\hat{\mathbf{x}}$. This convention also applies to $\mathbf{X}$ and $\boldsymbol{\mathcal{X}}$. 

\section{Basic Notions about Missing Data}\label{sec_pre}


\subsection{Mechanisms of Missingness}\label{sec_miss_mechan}

Missing data mechanisms are categorized into three types: missing completely at random (MCAR), missing at random (MAR), and missing not at random (MNAR). The specifics of each mechanism are detailed below.

\paragraph{Missing Completely at Random}
Data values are classified as missing completely at random (MCAR) if the circumstances causing a specific data point to be missing are entirely random and independent of both observable variables and unobservable parameters of interest. 
Formally, let $\mathbf{X}$ be a complete data matrix of size $n\times d$ with missing values and $\mathbf{M}$ be a binary indicator such that $m_{ij}=1$ if $x_{ij}$ is observed and $m_{ij}=0$ if $x_{ij}$ is missing. The data are MCAR if $P(\mathbf{M}\mid\mathbf{X})=P(\mathbf{M})$.
In the case of MCAR, the missing data decreases the study's analyzable population, thereby reducing the statistical power. However, they do not introduce any bias. Specifically, when data are MCAR, the remaining data can be viewed as a simple random sample from the complete dataset of interest. It's important to note that MCAR is generally regarded as a strong and often unrealistic assumption, while real data are rarely MCAR.

\paragraph{Missing at Random} When data are classified as missing at random (MAR), the occurrence of missing data is systematically linked to the observed data, but not to the unobserved data \citep{rubin1976inference,heitjan1996distinguishing,little2019statistical}. 
More formally, following the previous definition of $\mathbf{X}$ and $\mathbf{M}$, we split $\mathbf{X}$ into two parts, i.e., $\mathbf{X}=(\mathbf{X}_{\mathrm{obs}},\mathbf{X}_{\mathrm{mis}})$, where $\mathbf{X}_{\mathrm{obs}}$ and $\mathbf{X}_{\mathrm{mis}}$ denote the observed values and missing values respectively. The data are MAR if $P(\mathbf{M}\mid\mathbf{X})=P(\mathbf{M}\mid\mathbf{X}_{\mathrm{obs}})$.
For instance, in a survey, if men are more likely to withhold their weight than women, but this non-disclosure is not connected to the actual weights, then such data would be considered MAR. In other words, the likelihood of survey completion is associated with their gender (which is completely observed) but not their weight. MAR is a weaker assumption than MCAR and is more common in real-world data. However, handling MAR data is more complex and usually requires more sophisticated statistical techniques to avoid bias, such as multiple imputation or maximum likelihood estimation.

\paragraph{Missing Not at Random}
Missing not at random (MNAR), also referred to as non-ignorable nonresponse, represents a type of data that is neither MAR nor MCAR. In instances where data is MNAR, the absence of data is systematically connected to the unseen data, meaning that the missingness is linked to events or factors that the researcher has not measured.
To elaborate on the previous example, a weight registry might face MNAR data if participants who are overweight are more inclined to decline completing a weight survey. Another instance could be a survey including sensitive queries like personal income. If individuals with higher earnings are more prone to withhold their income information (resulting in missing data), the missingness of the income data correlates with the actual missing income values. This situation exemplifies MNAR. More details and subtypes of MNAR can be found in \citep{little2019statistical,pereira2024imputation}.
Addressing MNAR is a challenging task and may yield skewed results if not appropriately managed.

\subsection{Single Imputation and Multiple Imputation}

In single imputation, each missing value is replaced with a single estimated value. The methods used to estimate the missing value can vary, including mean imputation, median imputation, regression imputation, and many others that will be discussed in the subsequent sections of this paper. While single imputation is simple and easy to implement, it has significant limitations. It doesn't reflect the uncertainty about the imputations (since each missing value is replaced by a single value), which can lead to underestimated standard errors and overly confident statistical inferences \citep{rubin1987multiple}.

Multiple imputation addresses the limitations of single imputation by replacing each missing value with a set of plausible values, creating multiple complete datasets. These datasets are then analyzed separately, and the results are combined to produce estimates and confidence intervals that incorporate missing-data uncertainty. This process reflects the uncertainty about the right model to impute missing data and the randomness inherent in the data itself. As a result, multiple imputation provides more accurate and statistically valid results than single imputation \citep{rubin2018multiple,rubin1996multiple,murray2018multiple}. The book \citep{rubin2018multiple} provided a very detailed introduction, application, and discussion of multiple imputation. Compared to single imputation, the major limitation of multiple imputation is the higher computation cost, especially when advanced imputation algorithms are used. 

\subsection{Goals of Missing Data Imputation}

Missing data imputation not only plays an important role in data pre-processing, but also is the final goal of many real scenarios such as recommendation systems. Moreover, missing data imputation can be used to improve efficiency and reduce the cost of data acquisition and analysis.
 
\paragraph{Imputation as Data Preprocessing}
In many data analysis tasks—such as classification, regression, clustering, novelty detection, and causal inference—algorithms typically require complete data. The performance of these algorithms is often directly influenced by the quality of the imputed values. For example, a support vector machine trained on data where missing values are zero-filled might achieve 60\% classification accuracy on a test set; this accuracy can improve to 80\% when a reliable imputation algorithm is used.
The importance of imputation is further demonstrated across various domains. \cite{dietterich2018anomaly} showed that anomaly detection methods combined with data imputation techniques achieve significantly better performance than those without imputation. Similarly, \cite{fan2017sparse} demonstrated that clustering incomplete data can also benefit from missing data imputation. Furthermore, \cite{pmlr-v89-tu19a} studied the problem of causal discovery in the presence of missing data, highlighting its relevance in causal inference.

\paragraph{Imputation as Final Goal}
Certain problems can be naturally formulated as missing data imputation tasks. A primary example is found in recommendation systems, where an incomplete user-item interaction matrix (see Figure \ref{example_recommendation}) is used; predicting the missing interactions directly yields user or item recommendations. Another example is image or video inpainting, which involves replacing corrupted or unwanted pixels or patches with plausible values. This technique is widely used in multimedia entertainment and film-making. A further instance is transductive classification \citep{NIPS2010_3932}, particularly when feature sets contain missing values. In this context, \cite{NIPS2010_3932} treated unknown labels as missing data and employed low-rank matrix completion to simultaneously predict the labels and impute the missing features.

\paragraph{Imputation as Cost Reduction}
Data acquisition is often a time-consuming and costly process. For example, in questionnaire surveys, respondents may be presented with hundreds of questions, making it impractical for any individual to complete them all. To mitigate this, each respondent can be assigned a random subset of questions. Similarly, in chemical engineering, frequent measurement of certain components can be prohibitively expensive or hazardous; a more feasible strategy is to collect a limited number of samples and impute the missing values, thereby significantly reducing costs. This principle also applies to computing similarity or distance matrices, where measuring the pairwise relationship between every two objects is often infeasible. In such cases, one can measure a subset of the relationships and impute the remainder to reconstruct the full matrix.

\section{Missing Data Problem in Different Areas}\label{sec_areas}

\subsection{Missing Data in Social Science}

\begin{figure}
\centering
\includegraphics[width=0.7\textwidth]{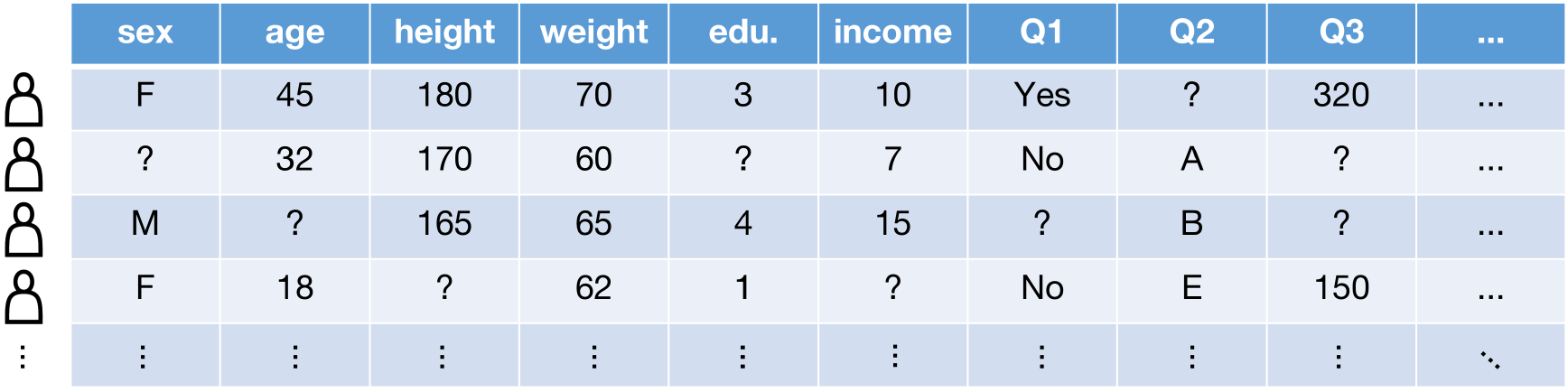}
\caption{Missing values (marked as "$?$") in survey data (row: subject; column: question).}
\label{fig_data_survey}
\end{figure}

Data in social science often suffers from missing values \citep{kalton1982imputing,little1988missing,little1989analysis}. For instance, questionnaire responses may be incomplete because respondents refuse to answer certain questions. Fig. \ref{fig_data_survey} shows an example of a questionnaire with missing values. Indeed, as people usually face numerous spam emails and pop-up advertisements and pay more attention to their privacy, it becomes increasingly difficult for social scientists to collect complete surveys. \citet{little1989analysis} reviewed methods for handling missing data in social science datasets, outlining three primary strategies. The first is imputation, which replaces missing values with appropriate estimates, allowing standard complete-data methods to be applied. The second is weighting, which involves dropping incomplete cases and assigning new weights to the complete cases to compensate for the lost information. The last strategy is direct analysis of the incomplete data that, for example, calculates the covariances using only observed pairs of variables. The review by \citet{porter2012missing} emphasized the importance of understanding the mechanisms driving missing data patterns.

It is worth mentioning that a questionnaire often contains multiple types of data, such as numerical data, categorical data, ordinal data, and even unstructured text. Missing data imputation for categorical data, ordinal data, and text is much more difficult than that for numerical data \citep{quintero2018missing}. In Section \ref{sec_categorical}, we will detail the imputation methods for non-numerical data. 

\subsection{Missing Data in Bioinformatics}

Bioinformatics is an interdisciplinary field that applies methods from biology, computer science, and statistics to analyze biological data such as gene expression and network data. However, the prevalence of missing values in such datasets often prevents the direct application of standard analytical methods. For instance, DNA microarrays—a classical technology for gene expression profiling—frequently contain missing values due to various factors, including insufficient resolution or image corruption. The need for imputation in this context is well-documented. For instance, \citet{troyanskaya2001missing} reviewed methods for DNA microarrays and demonstrated the effectiveness of k-nearest neighbor (kNN) and singular value decomposition approaches. Similarly, \citet{moorthy2014review} provided a review of several imputation algorithms for microarray gene expression data.

\begin{figure}
\centering
\includegraphics[width=0.7\textwidth]{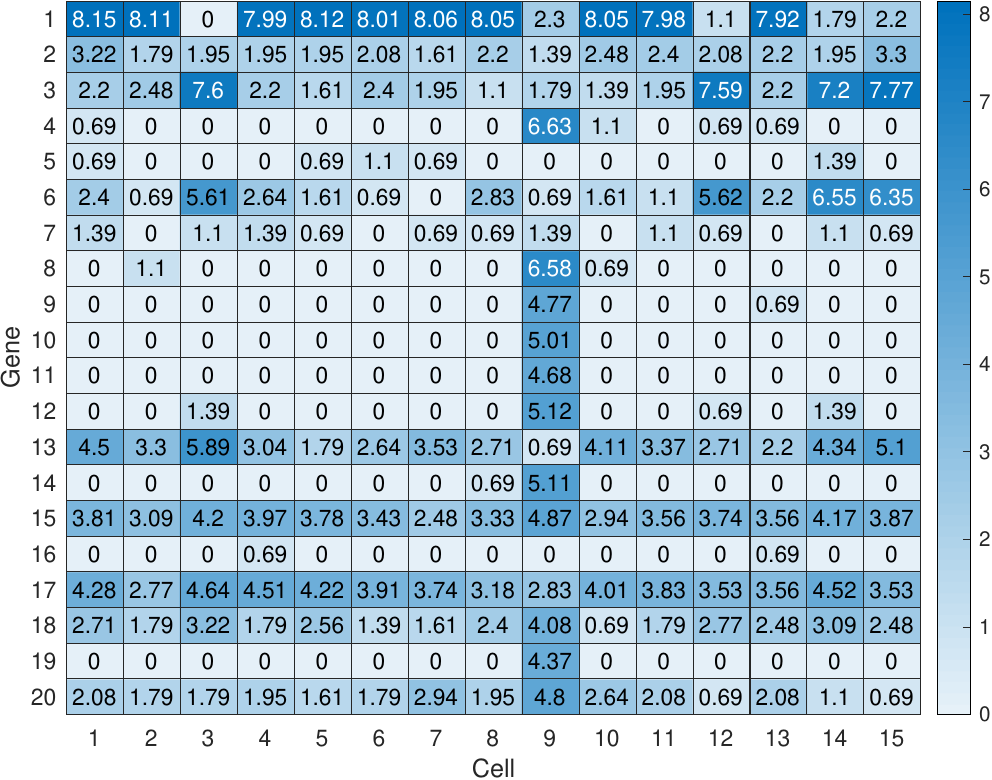}
\caption{Possible missing values (marked as zero) in scRNA-Seq data (log-transformed).}\label{fig_data_scrna}
\end{figure}

In the past decades, gene expression profiling technology has evolved a lot. For example, RNA sequencing (RNA-Seq), compared with microarray, allows for full sequencing of the whole transcriptome and has been widely used in the community of bioinformatics. Particularly, single-cell RNA sequencing (scRNA-Seq) provides the expression profiles of individual cells and can be used to uncover rare or new cell types within a cell population. 
As shown in Figure \ref{fig_data_scrna}, scRNA-Seq data often contains a lot of missing values.
Missing data imputation for scRNA-Seq has become a hot topic for a few years \citep{van2018recovering,huang2018saver,li2018accurate,lopez2018deep,linderman2018zero,chen2018viper,gong2018drimpute,huang2018saver,
eraslan2019single,mongia2019mcimpute,arisdakessian2019deepimpute,xu2020cmf,mongia2020deepmc,
xu2020scigans,wang2021scgnn,rao2021imputing,dai2022scimc}. These methods can be organized into two categories: model-based methods (e.g. \cite{van2018recovering,huang2018saver,li2018accurate}) and machine learning-based methods (e.g. \citep{mongia2019mcimpute,mongia2020deepmc,arisdakessian2019deepimpute,wang2021scgnn,xu2020scigans,rao2021imputing}). 
\citet{zhang2018comparison} conducted comprehensive experiments to compare many computational methods for imputing scRNA-Seq data.
\citet{dai2022scimc} provided a platform for comparing different missing data imputation methods for scRNA-Seq.

One challenge in scRNA-Seq imputation is that it is difficult to distinguish those genes not expressed at all from those produced by dropout \citep{miao2019screcover}.

\subsection{Missing Data in Healthcare}
In healthcare, a critical data source is the Electronic Health Record (EHR), a systematized collection of a patient's health information in digital format. EHRs typically encompass a wide range of data, including personal statistics, medical history, vital signs, medications, allergies, immunization status, laboratory test results, and radiology images. Originally used primarily for documenting patient information, EHRs are now leveraged with statistical and machine learning methods to improve healthcare outcomes \citep{shickel2017deep, yadav2018mining}. For instance, \citet{miotto2016deep} used deep learning to predict patients' future diseases from EHRs.

However, EHRs often contain a substantial amount of missing data. Consider physical examination data as an example: the total number of potential measurements—including urine tests, blood tests, liver and gallbladder tests, tumor marker tests, electrocardiograms, and X-rays—can amount to hundreds of items. A single patient typically undergoes only a subset of these tests. Consequently, from the perspective of the full suite of examination items, the resulting data table contains numerous missing values, with missingness rates for many items reaching as high as 80\%. This pervasive missingness complicates the analysis and utilization of EHRs. Conversely, accurately predicting missing data in EHRs could effectively simulate the results of a physical or medical test without performing it, potentially reducing both financial and labor costs.

Several studies have focused on missing data imputation for EHRs \citep{beaulieu2017missing, beaulieu2018characterizing, XU2020103576, ALAMOODI2021111236, kabir2022non}. For example, \citet{beaulieu2017missing} proposed a deep autoencoder to recover missing values. \citet{zhang2020predicting} used Extreme Gradient Boosting (XGBoost) \citep{chen2015xgboost} to predict missing values in the MIMIC-III dataset \citep{johnson2016mimic}, a well-known, freely accessible clinical database. \citet{zhou2021imputehr} developed a graphical tool called ImputEHR for implementing various imputation methods, from simple techniques to sophisticated gradient-boosted tree-based and neural network approaches. Furthermore, \citet{liu2022handling} provided a review of deep learning-based imputation techniques for healthcare data, and \citet{luo2022evaluating} evaluated state-of-the-art imputation methods specifically for clinical data from MIMIC-III.

\subsection{Missing Data in Image Science and Computer Vision}
\begin{figure}
     \centering
     \begin{subfigure}[b]{0.3\textwidth}
         \centering
         \includegraphics[width=\textwidth]{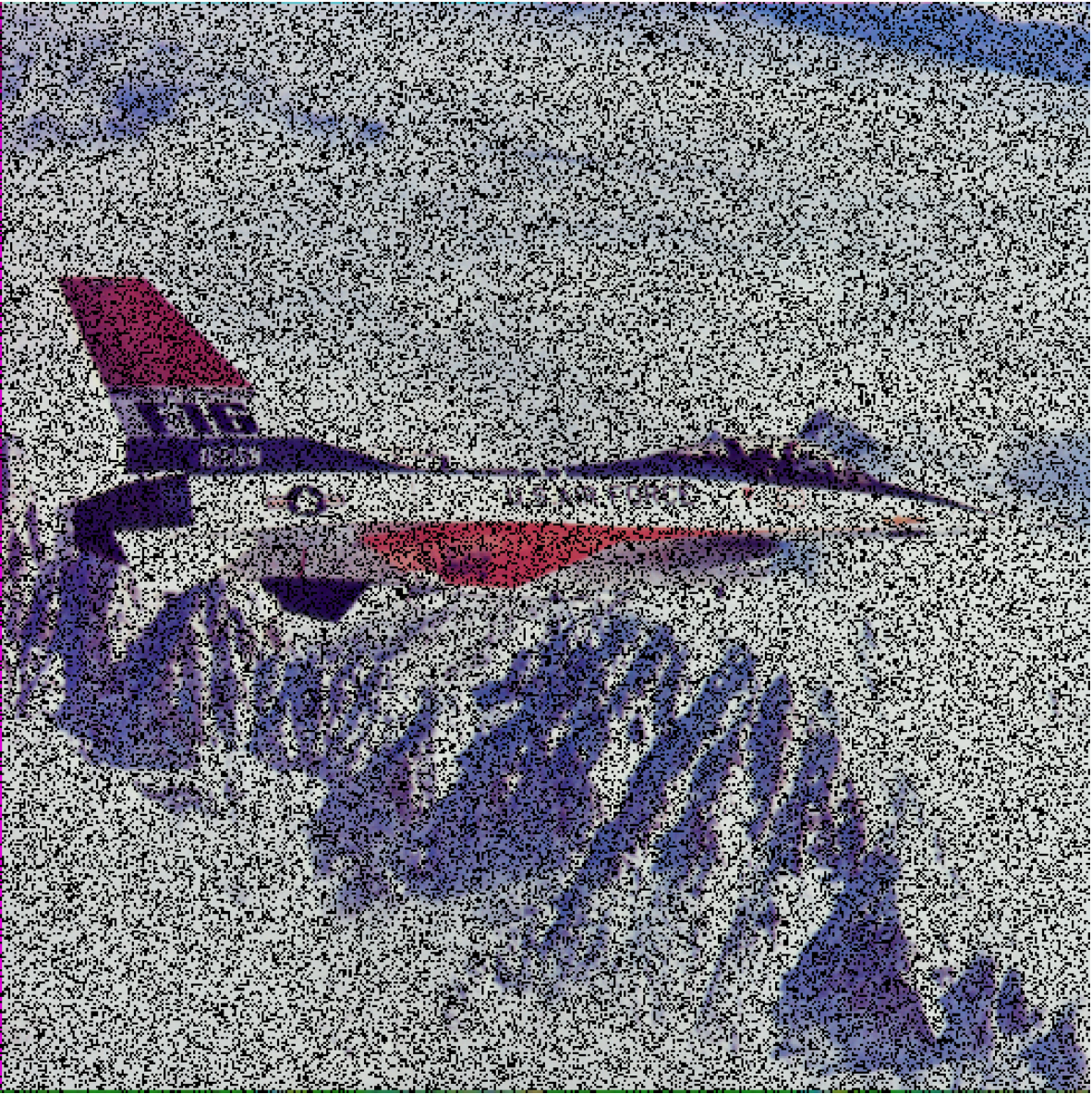}
         \caption{random missing (30$\%$ of the pixels)}
         \label{fig:y equals x}
     \end{subfigure}
     \hspace{50pt}
     \begin{subfigure}[b]{0.3\textwidth}
         \centering
         \includegraphics[width=\textwidth]{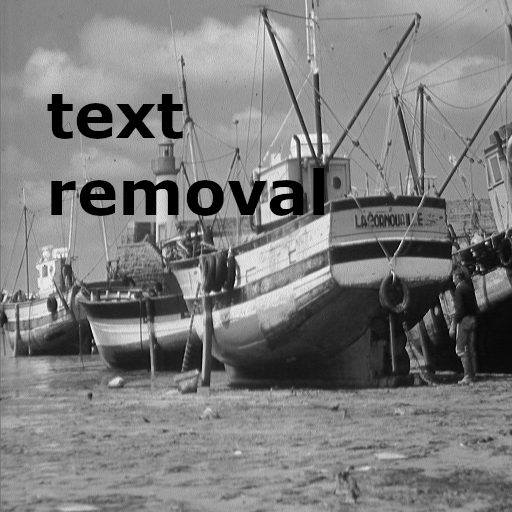}
         \caption{non-random missing (text removal)}
         \label{fig:three sin x}
     \end{subfigure}
     \caption{Two examples of the missing data problem of images. The original complete images are from \url{https://sipi.usc.edu/database/}.}\label{fig_eg_image_video}
\end{figure}

Images and videos can also contain missing data. Unlike other data types, the effects of missing data in these media are often visually apparent. For instance, missing pixels may occur during the signal acquisition process. Alternatively, pixels that are heavily corrupted are often removed, which also results in missing data, as illustrated in Figure \ref{fig_eg_image_video}(a). In other scenarios, the goal is to intentionally remove or replace specific pixels or regions with plausible content, a task known as image inpainting (see Figure \ref{fig_eg_image_video}(b)). When applied to video, this is referred to as video inpainting. These techniques are not confined to natural images and videos but are also applicable to medical imaging, such as Computed Tomography (CT) and Magnetic Resonance Imaging (MRI).

Unlike tabular data, images possess a strong local structure, meaning pixels within a small region or patch often have similar values. Furthermore, images, particularly hyperspectral images, exhibit channel-wise correlations. Consequently, effective imputation of missing values in images must leverage both the local and channel structure. Numerous image inpainting methods have been developed, which we categorize as follows: exemplar-based methods \citep{1323101}, total-variation based methods \citep{chan2006total,benning2013higher}, dictionary learning based methods \citep{5995592, LI20142862}, matrix completion based methods \citep{wen2012solving, hu2012fast, FANNLMC}, and deep learning based methods \citep{NIPS2012_6cdd60ea, fan2018matrix, XIANG2023109046}. It is noteworthy that most deep learning-based methods require a set of training images, whereas the other categories can typically handle single images without prior model training. More detailed literature reviews can be found in \citep{guillemot2013image, shen2015missing, elharrouss2020image, qin2021image, jam2021comprehensive, XIANG2023109046}.

Video inpainting \cite{10.1145/2980179.2982398,patwardhan2007video,yu2018generative,xu2019deep} involves replenishing missing segments in a video sequence with content that maintains spatial and temporal coherence. Also known as video completion, this technique has numerous practical applications, such as the removal of unwanted objects and video restoration. A recent survey by \citet{quan2024deep} comprehensively reviews deep learning methods for image and video inpainting.

\subsection{Missing Data in E-commerce and Social Media}

\begin{figure}[h]
     \centering
     \begin{subfigure}[b]{0.4\textwidth}
         \centering
         \includegraphics[width=\textwidth]{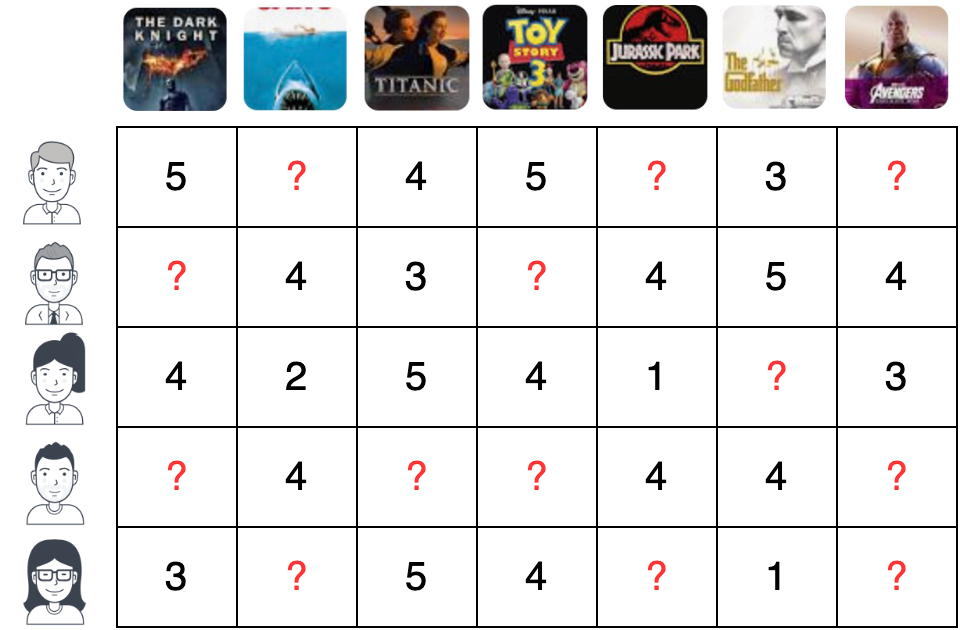}
         \caption{User-item interaction matrix}
         \label{fig:y equals x}
     \end{subfigure}
     \hspace{50pt}
     \begin{subfigure}[b]{0.4\textwidth}
         \centering
         \includegraphics[width=\textwidth]{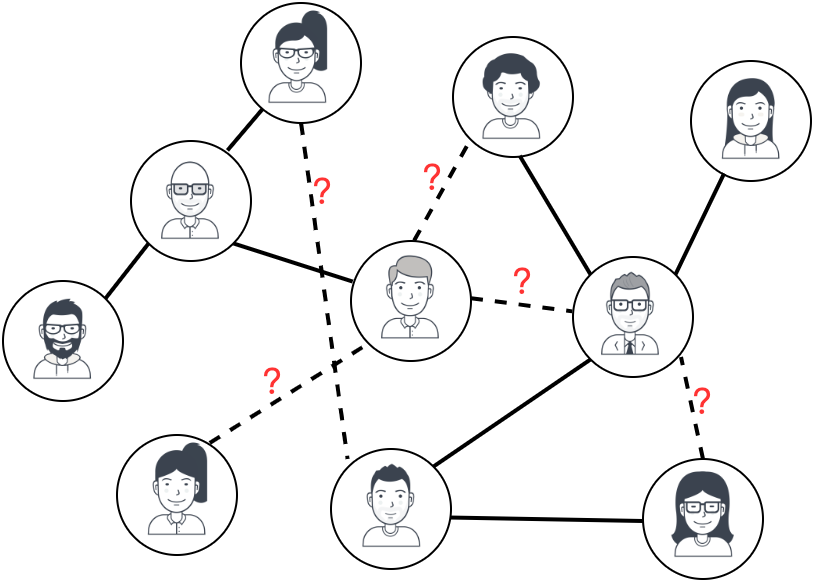}
         \caption{Link prediction for social network}
         \label{fig:three sin x}
     \end{subfigure}
     \caption{Examples of collaborative filtering (a) and link prediction (b). The question marks indicate missing values.}\label{example_recommendation}
\end{figure}

In e-commerce platforms like Netflix, Amazon, and Alibaba, user-item interaction data is inherently incomplete. This is because any single user typically interacts with only a small subset of items, and any single item is only rated or used by a small fraction of users. For example, Figure \ref{example_recommendation}(a) illustrates an incomplete rating matrix where a few users have rated a few movies. Predicting the missing entries in such a matrix forms the basis of a recommendation system.
Beyond explicit feedback like ratings, user-item interactions can also be measured through implicit feedback, such as time spent viewing content or the number of clicks on a product. The task of predicting these unknown user preferences is known as collaborative filtering. Recommendation systems, which perform this task, play an increasingly vital role in daily life, influencing user behavior both explicitly and implicitly.

Recommendation systems are generally categorized into three types \citep{breeseCF1998,sarwar2001item,2005CFsurvey,Zhang2019}: content-based methods, collaborative filtering, and hybrid methods. Content-based methods recommend items that are similar to those a user has liked, or target users who are similar to those who have liked an item. This similarity is typically derived from side information, such as item attributes (e.g., genre) or user profiles (e.g., sex, age, and occupation). In contrast, collaborative filtering leverages latent correlations among users and items to predict unknown preferences. Hybrid methods, as the name suggests, combine content-based and collaborative filtering approaches. Notably, collaborative filtering is closely related to missing data imputation, while content-based methods more closely resemble regression with incomplete response variables. Given this paper's focus, we will concentrate primarily on collaborative filtering (CF).

We simply organize CF methods into the following four categories: similarity-based methods \citep{knncf,FKIH20227645,sarwar2002recommender,beregovskaya2021review}, matrix factorization-based methods, neural network-based methods, and other methods \citep{markov1,breeseCF1998,bayes2}.
Particularly, in the past decades, matrix factorization-based methods \citep{billsus98,Salakhutdinov07,koren09,fan2019factor,rendle2019difficulty}
have been extensively studied and used in CF. These methods usually exploit the potential low-rank structure of the incomplete user-item utility matrix via embedding items and users into a latent space of reduced dimension, where the observed utilities are approximated by the inner products of the user feature vectors and item feature vectors. More details will be introduced in Section \ref{sec_lrmc}. In recent years, neural networks and deep learning have shown promising performance in CF \citep{RBMCF,nnmf,Sedhain2015,CDAE,Zheng2016,he2017neural,Berg2017a,muller18a,fan2018matrix,Yi2020}. 
For instance, \cite{Sedhain2015} proposed an autoencoder \citep{hinton2006reducing} based CF method called AutoRec. It predicts unknown ratings by an encoder-decoder model, though the input of the encoder contains missing values (usually filled with zeros). AutoRec outperformed a few matrix factorization-based methods, such as SVD++ \citep{koren09} and LLORMA \citep{LLORMA}, on several benchmark datasets \citep{Sedhain2015}.
More about collaborative filtering can be found in the following review papers  \citep{burke2002hybrid,su2009survey,bobadilla2013recommender,wang2021survey}.

Similar to the user-item utility matrix, interaction networks between users on social media platforms (e.g., Facebook and WeChat) can be leveraged for friend recommendations, targeted advertising, and product suggestions. These networks are typically represented as graphs where nodes represent users and edges represent connections. Such graphs are often incomplete due to the impracticality of mapping every possible connection, resulting in missing edges. The task of predicting these unknown connections is known as link prediction. An intuitive example of this process in a social network is illustrated in Figure \ref{example_recommendation}(b).

\subsection{Missing Data in Manufacturing Industries}
\begin{figure}
\includegraphics[width=1\textwidth]{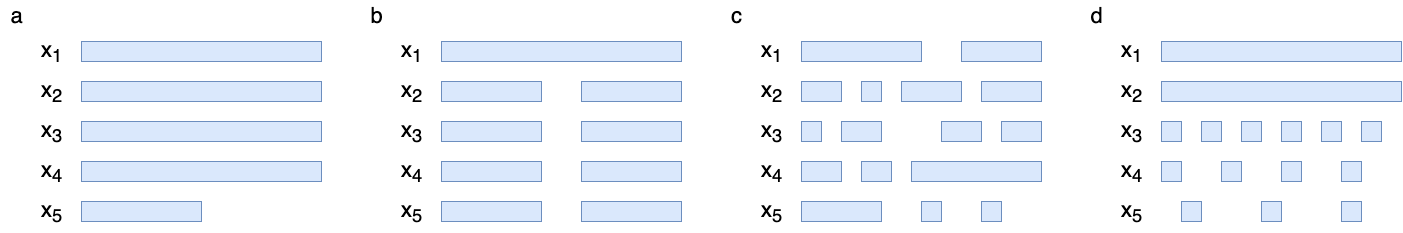}
\caption{Examples of missing data patterns in industrial process: (a) sensor breakdown; (b) process shutdown; (c) general pattern (i.e., outlier combined with sensor breakdown); (d) multi-rate data. In each plot, the rows correspond to the variables (or sensors), while the columns correspond to observations; each small square indicates a scalar.}
\label{fig_miss_pattern_process}
\end{figure}

In manufacturing industries, such as chemicals and consumer electronics, failures in numerous equipment and sensors can lead to missing data \citep{lakshminarayan1999imputation, imtiaz2008treatment, 8846984}. The data in these contexts are often multivariate time series, making it essential to account for temporal dynamics when recovering missing values. \cite{imtiaz2008treatment} focused on process data and categorized missing data patterns in chemical processes into four types (shown in Figure \ref{fig_miss_pattern_process}): (a) sensor breakdown; (b) process shutdown; (c) general patterns (e.g., outliers combined with sensor breakdown); and (d) multi-rate data. These patterns are typically classified as Missing Not at Random (MNAR).

Several studies have addressed missing data problems in manufacturing industries \citep{imtiaz2008treatment, ZIAEIHALIMEJANI2021107549, 8891716, carbery2022missingness, fantii2022, JEONG2023103937}. For example, \cite{8846984} described the challenges associated with missing data in industrial data analytics, while \cite{du2020missing} reviewed the missing data problem in sensor network-based monitoring systems.

\section{General Missing Data Imputation}\label{sec_methods}

To handle incomplete data, a naive approach is to replace the missing values with the mean (or median) of the observed values. This approach, however,
distorts the variances and covariances that are crucial to inference. In this section, we review the methods of missing data imputation.

\subsection{Basic Imputation via Filling with Zero, Mean, Median, or Mode}

The simplest strategy for handling missing data is single-value imputation, which replaces missing entries with easily computed, statistically plausible values such as zero, the mean, median, or mode.

Filling missing values with zeros often yields statistical bias because the means of the variables are not necessarily zero. Note that in count data such as scRNA-Seq data, a zero could be either an observed value or a missing value, which means in many scenarios the missing values are naturally filled with zeros. On the other hand, zero filling can be a useful initialization for some missing data imputation algorithms such as low-rank matrix completion (see Section \ref{sec_lrmc}) and autoencoder-based imputation (see Section \ref{sec_deep_impute}). An appealing property of zero filling is that for regression models, especially neural networks, the contribution of the corresponding missing value is prohibited, which may improve the robustness or generalization ability of the models \citep{fan2024neuronenhanced}. A few researchers showed that zero filling has an adverse effect on model performances \citep{smieja2018processing,yi2019not}. 

\begin{figure}[h]
     \centering
     \begin{subfigure}[b]{0.45\textwidth}
         \centering
         \includegraphics[width=\textwidth]{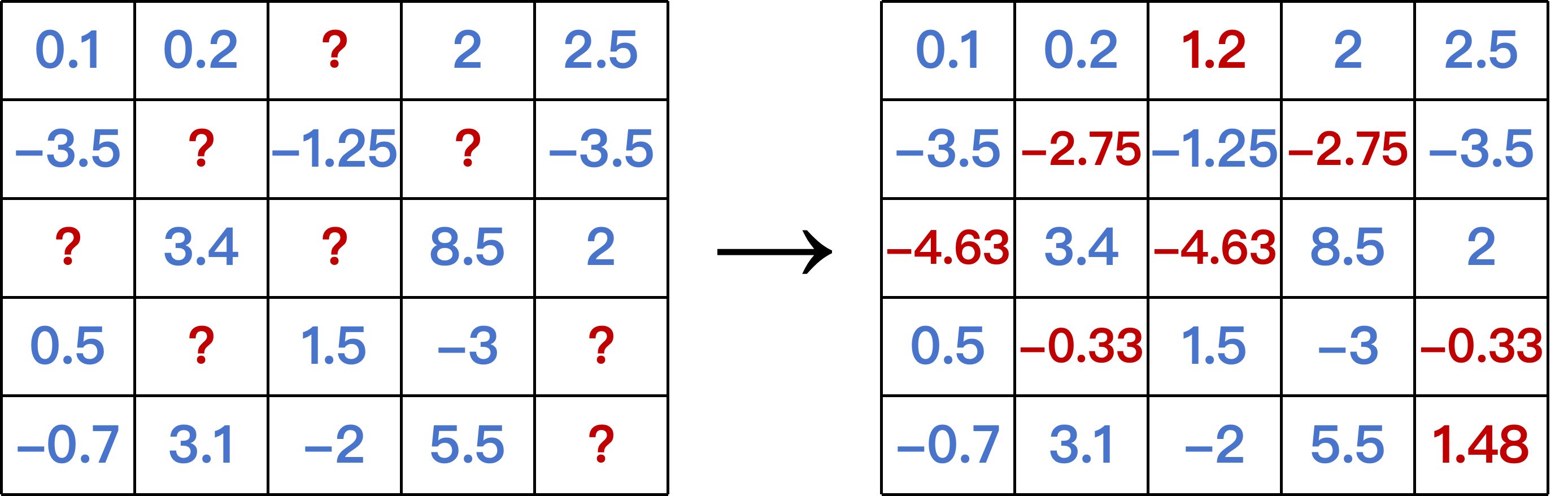}
         \caption{Mean (row-wise) imputation}
     \end{subfigure}
     \hspace{20pt}
     \begin{subfigure}[b]{0.45\textwidth}
         \centering
         \includegraphics[width=\textwidth]{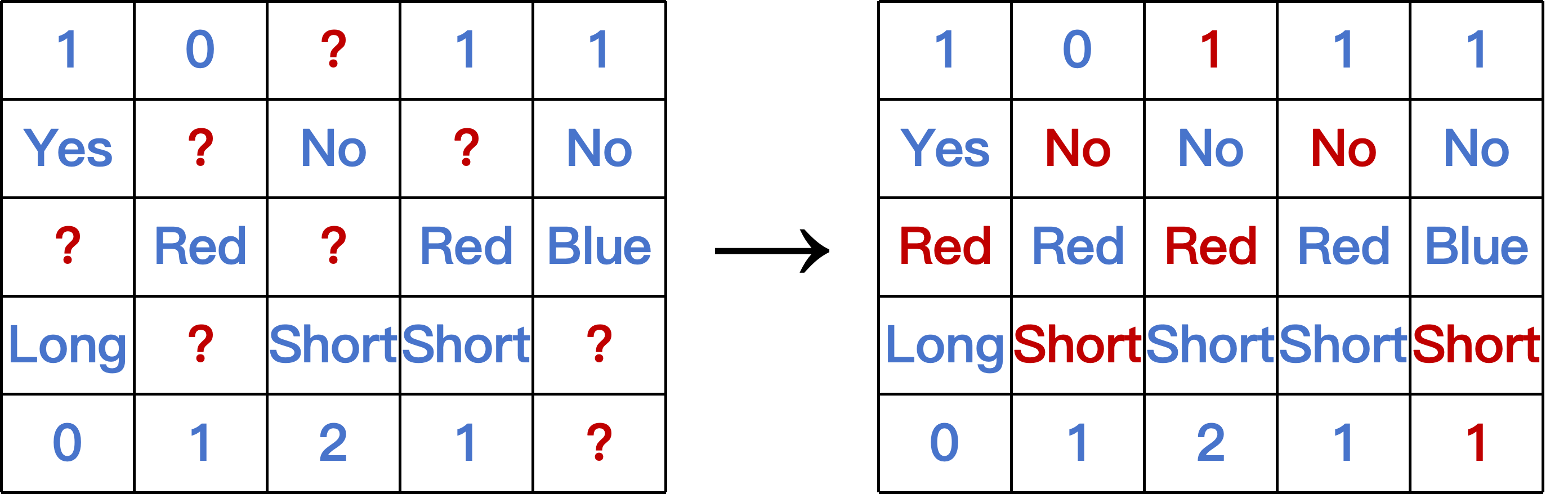}
         \caption{Mode (row-wise) imputation}
     \end{subfigure}
\caption{Toy examples of mean imputation and mode imputation.}\label{fig_mean_mode}
\end{figure}

Mean filling, or mean imputation or substitution, is filling in missing data in a dataset by replacing the missing values with the mean (average) value of the complete cases for that variable. Mean filling is simple and efficient, and preserves the overall mean of the dataset, which can be important for statistical analysis. A simple example is shown in Figure \ref{fig_mean_mode}(a). However, mean filling has the following disadvantages. First, mean imputation tends to underestimate the variance of the data. Second, it only considers the average of the observed values for the variable with missing values and does not take into account possible correlations with other variables. Third, mean imputation can lead to biased parameter estimates, like means and correlations, especially if the data is MNAR. Lastly, mean imputation does not preserve the relationships among variables and can distort the original data distribution. Median and mode imputations share similar advantages and disadvantages with mean imputation. Particularly, for categorical variables and ordinal variables, we should use mode imputation rather than mean imputation. See the example in Figure \ref{fig_mean_mode}(b).

\subsection{Regression Imputation}\label{sec_regimpu}

The aforementioned zero, mean, median, and mode imputations only consider the distribution of the values of the variable that has missing entries. However, if there is a correlation between the target variable and other variables, which is often the case in practice, regression of the missing variable on these other variables often provides more accurate estimations, especially when data are MNAR \citep{little1988missing,roth1994missing}. For instance, in a bivariate analysis where one variable has missing data, the regression equation is estimated using only the cases with complete data. In this equation, the variable with incomplete data is treated as the outcome, while the variable with complete data serves as the predictor. \cite{baraldi2010introduction} provided an intuitive example that used the math score to predict course grade.
More generally, suppose we have $d$ variables $x_1,x_2,\ldots,x_d$ and we want to impute the missing values for $x_j$. Regression imputation first uses the complete observations to build a regression model, such as 
\begin{equation}\label{eq_reg_imp_0}
x_j=f^{(j)}(x_1,\ldots,x_{j-1},x_{j+1},\ldots,x_d)+\varepsilon
\end{equation}
where $\varepsilon$ denotes the error term and $f^{(j)}$ could be a linear function or nonlinear function. An example of $f^{(j)}$ is $\sum_{i\neq j}w_ix_i+b$, where $w_i$ are the coefficients and $b$ is a bias term. 

An important variation of regression imputation is the iterative regression method \citep{roth1994missing}. It begins by calculating an initial correlation matrix using a straightforward imputation method, like mean imputation, or simply using the observed values. Following this, regression equations are derived from this matrix, and the missing values are estimated and replaced in the data matrix. This cycle continues until the regression coefficients exhibit minimal changes.

Perhaps the best-known iterative regression imputation algorithm is the missForest proposed by \cite{MissForest}. In missForest, the $f^{(j)}$ in \eqref{eq_reg_imp_0} is a random forest, and there are $d$ random forests in total. The main steps of missForest are summarized as follows:
\begin{itemize}
\item An initial guess is made for the missing values. For instance, this could be done by the mean or median imputation for numerical variables or mode imputation for categorical variables.
\item For each variable $x_j$, a random forest model $f^{(j)}$ is trained on the observed data, namely the complete observations of $x_j$ and the corresponding values of $x_i$, $i\neq j$, where $x_j$ is treated as the response variable, and the other variables are used as predictors. The trained model $f^{(j)}$ is used to predict the missing values of $x_j$. And then update the missing values of $x_j$.
\item The step above is repeated until the imputations converge to a stable solution.
\end{itemize}
One of the advantages of missForest is that it can handle non-linear relationships and interactions between variables, since $f^{(j)}$ are random forests. It also provides an out-of-bag estimate of the imputation error, which can be useful for assessing the quality of the imputations. Additionally, missForest is effective in handling mixed-type data, including numerical and categorical variables.

The random forest within the missForest algorithm can be replaced with other regression models, such as polynomial regression, support vector regression \citep{drucker1996support}, Gaussian processes \citep{williams1995gaussian}, or neural networks \citep{lecun2015deep}. This substitution can create new regression-based imputation algorithms, though potentially at the cost of higher computational complexity and reduced effectiveness with categorical data. In practice, missForest remains a stable and effective benchmark. To the best of our knowledge, no variant proposed in the past decade has substantially surpassed its imputation accuracy. For instance, while \cite{liu2023improving} incorporated a Markov Blanket discovery method to refine the feature selection process in missForest, it achieved only a marginal improvement in accuracy.

Regression imputation has several potential drawbacks. It can artificially reduce the variance of imputed variables and introduce bias if the underlying model assumptions are violated \citep{roth1994missing,baraldi2010introduction}. Furthermore, if the proportion of missing data is very high (e.g., 90\%), estimating the regression function $f^{(j)}$ 
may be infeasible or highly inaccurate, leading to imputation failure. Although iterative regression methods can mitigate this issue, they remain sensitive to the initial estimates, and an unreliable starting correlation matrix can still yield poor final results.

Notably, the core principle of regression imputation—leveraging associations among variables to predict missing values—is shared by many modern techniques, including low-rank matrix completion and certain deep learning approaches. This connection will be further explored in the corresponding sections.

\subsection{Hot-Deck Imputation}

Hot-deck imputation \citep{andridge2010review} replaces the missing values of one or several variables for a non-respondent (known as the recipient) with observed values from a respondent (referred to as the donor) who shares similar characteristics with the non-respondent. Some variations of this method involve randomly selecting the donor from a group of potential donors, also known as the donor pool. These are termed as random hot deck methods. Conversely, some versions identify a single donor, typically the nearest neighbor (NN) based on a certain metric, and impute values from that particular case. These are referred to as deterministic hot-deck methods, as they don't involve any randomness in the selection of the donor. There are also methods that impute summaries of values from a group of donors, like the mean or weighted average, instead of individual values. 
The process of hot-deck imputation typically involves the following steps:
\begin{itemize}
\item Define a metric for determining similarity or distance between cases. This could be based on variables that are believed to be related to the variable with the missing data.
\item For each case with missing data, find a donor case that is similar based on the defined metric but does not have missing data on the variable of interest.
\item Replace the missing value with the observed value from the donor case.
\end{itemize}
This method generally offers greater accuracy than basic imputation strategies, as it replaces missing data with plausible values derived from similar, observed cases rather than a single statistic like the mean, which can distort the underlying variable distributions. A key advantage is that the imputed values preserve the original data's distributional characteristics because they are conditioned on the matching (categorization) variables. This approach is particularly beneficial when data is missing in specific patterns \citep{roth1994missing}. The methodology, implementation, and applications of this technique, known as hot-deck imputation, are detailed in \citep{andridge2010review}.

For hot-deck imputation, the nearest neighbors or the donor pool are found by using a distance or similarity metric. There are many choices for the distance or similarity metric. For numerical data, one may use the Euclidean distance, $\ell_1$ distance,  cosine similarity, Mahalanobis distance, Chebyshev distance, etc \citep{FKIH20227645}. For categorical data, one may use the overlap measure (or Hamming distance), Eskin \citep{eskin2002geometric}, Goodall's measures \citep{goodall1966new}, Gambaryan measure, inverse occurrence frequency, etc. A review of similarity measures for categorical data can be found in \citep{chandola2007similarity}. Consequently, for mixed-type data, one needs to combine different distance or similarity measures.

\begin{figure}[h]
     \centering
         \includegraphics[width=0.75\textwidth]{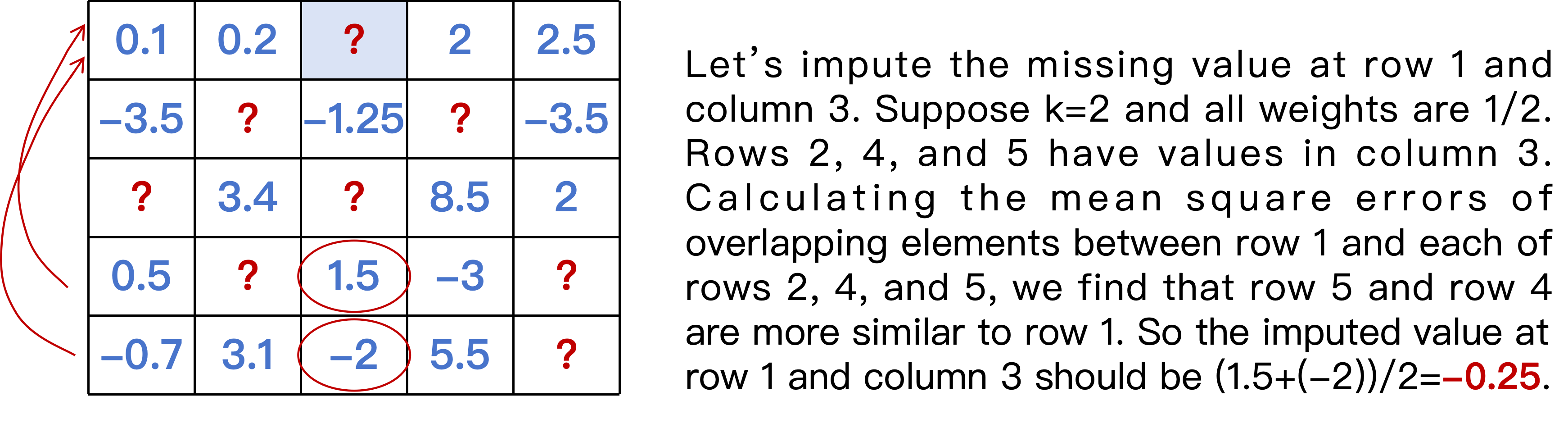}
\caption{A simple example of kNN imputation.}\label{fig_knn}
\end{figure}

The most representative example of hot-deck imputation is kNN imputation \citep{chen2000nearest,beretta2016nearest}, which is a deterministic hot-deck method. Suppose ${x}_{ij}$, the variable $j$ at observation $i$, is a missing value. kNN imputation make prediction for ${x}_{ij}$ by the following weighted sum
\begin{equation}
\hat{x}_{ij}=\sum_{v \in \mathcal{N}_k}x_{vj}w_{v},
\end{equation}
where $\mathcal{N}_k$ denotes the index set of the $k$ nearest neighbors of observation $i$ without missing values of variable $j$. $w_{v}$ are the weights of the neighbors and satisfy $\sum_{v \in \mathcal{N}_k}w_v=1$. If $w_v$, $v \in \mathcal{N}_k$, are all $1/k$, $\hat{x}_{ij}$ is just the average of the $k$ nearest neighbors. In recommendation systems, $w_v$ are usually set to $w_v=s_{vi}/\sum_{l\in \mathcal{N}_k}s_{li}$ \citep{breeseCF1998,sarwar2001item}, where $s_{li}$ denotes the similarity between observations $l$ and $i$. Note that $s_{li}$ is computed from only the overlapped observed values of the two samples. An illustrative example is shown in Figure \ref{fig_knn}.

Hot-deck imputation is widely used due to its relative simplicity and its non-parametric nature, as it does not require assumptions about the underlying data distribution. However, the method has several limitations. It can underestimate data variability by redistributing existing values without introducing new information. Furthermore, imputation quality is highly dependent on the chosen similarity metric; a poor metric that fails to capture true associations between cases can lead to inaccurate imputations. This issue is exacerbated by high missingness rates or noisy data.

\subsection{Likelihood Methods}\label{sec_likelihood}
Maximum likelihood is a strategy for estimating the unknown parameters of a model from data. When data is complete, it's straightforward to calculate quantities like the mean, covariance, and linear regression coefficients, which are maximum likelihood estimates derived from maximizing the likelihood of the observed data. This same principle applies even when there are missing values in the data, as the estimation can still be based on the likelihood of the observed values. However, when missing data is present, the likelihood of the observed data becomes more complex compared to the typical data analysis scenarios, and finding an estimate that maximizes this likelihood doesn't have a straightforward solution. To tackle the challenge,  \cite{dempster1977maximum} proposed the Expectation-Maximization (EM) algorithm for incomplete data. The main idea and algorithm are explained in the following context.

Following the notations defined by Section \ref{sec_miss_mechan} and letting $\theta$ and $\phi$ be the parameters of the data model and the parameters of the missing mechanism, respectively, the full likelihood can be formulated as
\begin{equation}\label{eq_likelihood_0}
L_{\text {full}}\left(\theta, \phi \mid \mathbf{Y}_{\mathrm{obs}}, \mathbf{M}\right)=c\times\int f\left(\mathbf{X}_{\mathrm{obs}}, \mathbf{X}_{\mathrm{mis}} \mid \theta\right) f\left(\mathbf{M} \mid \mathbf{X}_{\mathrm{obs}}, \mathbf{X}_{\mathrm{mis}}, \phi\right) d \mathbf{X}_{\mathrm{mis}},
\end{equation}
where $c$ is some constant independent of $\theta$ and $\phi$ and $f$ denotes the density function. When the missing mechanism is MAR (or MCAR), the log-likelihood is
\begin{equation}\label{eq_likelihood_1}
\log L_{\text {full}}\left(\theta, \phi \mid \mathbf{Y}_{\mathrm{obs}}, \mathbf{M}\right)=\log c+\log f\left(\mathbf{M} \mid \mathbf{X}_{\mathrm{obs}}, \phi\right)+\log f\left(\mathbf{X}_{\mathrm{obs}}\mid \theta\right),
\end{equation}
where $f\left(\mathbf{X}_{\mathrm{obs}}\mid \theta\right)=\int f\left(\mathbf{X}_{\mathrm{obs}}, \mathbf{X}_{\mathrm{mis}}\mid \theta\right) d \mathbf{X}_{\mathrm{mis}}$.
Note that the prior distributions of $\theta$ and $\phi$, if available, can also be incorporated into \eqref{eq_likelihood_1}. If we do not care about $\phi$, we only need to maximize $f\left(\mathbf{X}_{\mathrm{obs}}\mid \theta\right)$, that is
\begin{equation}\label{eq_EM_obj}
\mathop{\text{maximize}}_{\theta}~ \log\int f\left(\mathbf{X}_{\mathrm{obs}}, \mathbf{X}_{\mathrm{mis}}\mid \theta\right) d \mathbf{X}_{\mathrm{mis}}.
\end{equation}
The optimization is often difficult and is usually handled by the EM algorithm. The E-step is to compute the following expectation
\begin{equation}\label{eq_EM_E}
Q\left(\theta, \theta^{\prime}\right)=\mathbb{E}_{\theta^{\prime}}\left[\log f\left(\mathbf{X}_{\mathrm{obs}}, \mathbf{X}_{\mathrm{mis}}\mid \theta\right) \mid \mathbf{X}_{\mathrm{obs}}\right],
\end{equation}
which is a lower bound of the objective in \eqref{eq_EM_obj} plus some constant.
The M-step is to solve
\begin{equation}\label{eq_EM_M}
\mathop{\text{maximize}}_{\theta} ~Q\left(\theta, \theta^{\prime}\right)
\end{equation}
The two steps are repeated until some convergence conditions are met. Finally, the missing values are filled by using the estimated $\theta$. See the example of multivariate Gaussian in Section 11.6 of \citep{murphy2012machine}.

Despite its effectiveness, the EM algorithm has several major limitations. First, it relies on the MAR assumption and is not suitable for data that are MNAR. Second, its performance is contingent on a pre-specified data distribution (e.g., Gaussian); violations of this assumption can lead to inaccurate imputations. Finally, the algorithm is computationally intensive, and its results can be sensitive to the initial parameter values.

\subsection{Matrix Completion}
\subsubsection{Low-Rank Matrix Completion}\label{sec_lrmc}
\begin{figure}[h]
     \centering
     \begin{subfigure}[b]{0.52\textwidth}
         \centering
         \includegraphics[width=\textwidth]{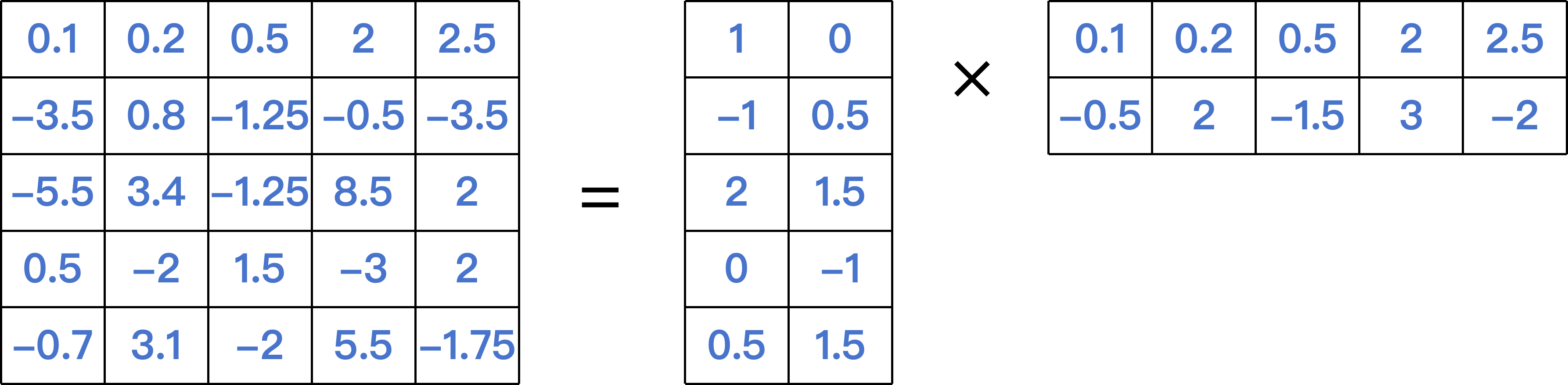}
         \caption{A rank-2 low-rank matrix of size $5\times 5$}
     \end{subfigure}
     \hspace{20pt}
     \begin{subfigure}[b]{0.40\textwidth}
         \centering
         \includegraphics[width=\textwidth]{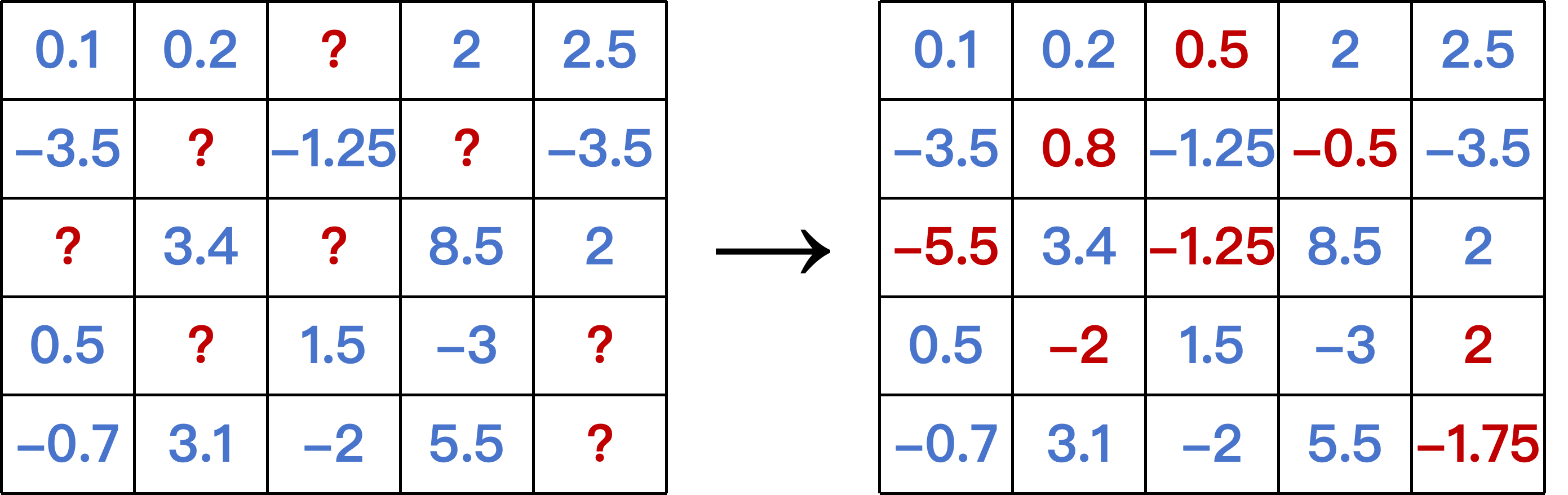}
         \caption{Completion of the low-rank matrix}
     \end{subfigure}
\caption{A toy example of LRMC.}\label{fig_lrmf}
\end{figure}

Low-rank matrix completion (LRMC) \citep{srebro2005rank,CandesRecht2009,wen2012solving,hardt2014understanding} addresses the problem of recovering missing entries in a matrix under the assumption that the true matrix is low-rank. Formally, a matrix is low-rank if its rank (the dimension of the vector space spanned by its rows or columns) is significantly smaller than its dimensions. An equivalent characterization is that the matrix has a rapidly decaying singular value spectrum, with only a few non-zero or significant singular values. Consider a matrix $\mathbf{X}\in\mathbb{R}^{m\times n}$, where $m<n$, the singular value decomposition (SVD) is denoted as $\mathbf{X}=\mathbf{U}\boldsymbol{\Sigma}\mathbf{V}^\top$, where $\mathbf{U}\in\mathbb{R}^{m\times m}$ and $\mathbf{V}\in\mathbb{R}^{n\times n}$ are orthonormal matrices, i.e., $\mathbf{U}^\top\mathbf{U}=\mathbf{I}_m$ and $\mathbf{V}^\top\mathbf{V}=\mathbf{I}_n$, $\boldsymbol{\Sigma}=(\text{diag}(\sigma_1,\sigma_2,\ldots,\sigma_m), \boldsymbol{0})$, and $\sigma_1\geq \sigma_2\geq\cdots\geq \sigma_m$ are singular values. The rank of $\mathbf{X}$ is defined as $\text{rank}(\mathbf{X}):=\sum_{i=1}^m\mathbbm{1}(\sigma_i\neq 0)$. 
It has been found that many data matrices in real applications are low-rank ($\text{rank}(\mathbf{X})\ll m$) or can be well approximated by a low-rank matrix, i.e., many singular values are close to zero. Figure \ref{fig_lrmf}(a) shows a low-rank matrix of size $5\times 5$, where the rank is 2. Figure \ref{fig_lrmf}(b) shows the corresponding LRMC task.

Over the past decades, numerous LRMC methods have been developed, which generally fall into two categories: low-rank regularization and low-rank factorization. Given an incomplete data matrix $\tilde{\mathbf{X}}\in\mathbb{R}^{m\times n}$, where the set of locations of observed entries is denoted as $\Omega$, the low-rank regularization methods aim to solve the following problem
\begin{equation}\label{eq_lrmc_rank}
\begin{aligned}
    \underset{\hat{\mathbf{X}}}{\operatorname{minimize}}~ &\text{rank}(\hat{\mathbf{X}}), \\
    \text{subject to}~ &\mathcal{P}_{\Omega}(\hat{\mathbf{X}})=\mathcal{P}_{\Omega}(\tilde{\mathbf{X}}),
\end{aligned}
\end{equation}
where the operator $\mathcal{P}_{\Omega}: \mathbb{R}^{m \times n} \rightarrow \mathbb{R}^{m \times n}$ acts on any $\mathbf{X} \in \mathbb{R}^{m \times n}$ in the following way: $\left(\mathcal{P}_{\Omega}(\mathbf{X})\right)_{i j}=\mathbf{X}_{i j}$ if $(i, j) \in \Omega$ and 0 if $(i, j) \notin \Omega$. However, since the direct rank minimization problem \eqref{eq_lrmc_rank} is NP-hard, a standard approach is to replace the rank with a tractable surrogate or regularizer $R(\hat{\mathbf{X}})$ and solve
\begin{equation}\label{eq_lrmc_rank}
\begin{aligned}
\underset{\hat{\mathbf{X}}}{\operatorname{minimize}}~& R(\hat{\mathbf{X}})\\
    \text{subject to}~ &\mathcal{P}_{\Omega}(\hat{\mathbf{X}})=\mathcal{P}_{\Omega}(\tilde{\mathbf{X}})
\end{aligned}
\end{equation}
Popular examples for $R(\hat{\mathbf{X}})$ include the nuclear norm, Schatten-$p$ quasi-norm, etc. Table \ref{tab_regularizer} presents a summary of rank regularizers. Solving \eqref{eq_lrmc_rank} requires performing SVD on an $m\times n$ matrix in each iteration. Therefore, the time complexity in each iteration is $\mathcal{O}(m^2n)$ if using full SVD and can be reduced to $\mathcal{O}(mrn)$ if using truncated SVD, where $r<m$. The method given by \eqref{eq_lrmc_rank} can be generalized to noisy low-rank matrix completion \citep{candes2010matrix} if the low-rank matrix is corrupted by noise or the target data matrix is approximately low-rank. A general formulation is as follows
\begin{equation}\label{eq_lrmc_rank_reg}
\begin{aligned}
\underset{\hat{\mathbf{X}}}{\operatorname{minimize}}~& \|\mathcal{P}_{\Omega}(\tilde{\mathbf{X}}-\hat{\mathbf{X}})\|_F^2+\lambda\cdot R(\hat{\mathbf{X}})
\end{aligned}
\end{equation}
where $\lambda>0$ is a hyperparameter.

LRMC methods based on the regularizers in Table \ref{tab_regularizer} are not scalable to very large matrices, e.g., $m=10000$. Therefore, the second group of LRMC methods, which are based on factorizing the decision matrix $\hat{\mathbf{X}}$ as the product of two smaller matrices denoted as $\mathbf{A}\in\mathbb{R}^{m\times d}$ and $\mathbf{B}\in\mathbb{R}^{d\times n}$, received increasing attention. These methods are generally in the following form
\begin{equation}\label{eq_lrmc_mf}
\begin{aligned}
\underset{\mathbf{A},\mathbf{B}}{\operatorname{minimize}}~& \|\mathcal{P}_{\Omega}(\tilde{\mathbf{X}}-\mathbf{AB})\|_F^2+\lambda \cdot R(\mathbf{A},\mathbf{B})\\
\end{aligned}
\end{equation}
where $R$ denotes the regularization on the factors $\mathbf{A}$ and $\mathbf{B}$. Table \ref{tab_regularizer_2} summarizes the choices for $R$, where most of them are closely related to the nuclear norm and Schatten-$p$ quasi-norms. These regularizers are more efficient to implement than those in Table \ref{tab_regularizer} because they do not require SVD or only need to perform SVD on the smaller factors. Once $\mathbf{A}$ and $\mathbf{B}$ are found, the recovered matrix is $\hat{\mathbf{X}}=\mathbf{A}\mathbf{B}$.

\begin{table}
\centering
\caption{Popular rank regularizers ($\mathbf{X}\in\mathbb{R}^{m\times n}$, $m\leq n$; $\sigma_i(\mathbf{X})$ denotes the $i$-th largest singular value of $\mathbf{X}$).}\label{tab_regularizer}
\begin{tabular}{c|c|c}
\toprule
regularizer & definition & reference \\ \midrule
nuclear norm &  $\|\mathbf{X}\|_\ast=\sum_{i=1}^m\sigma_i(\mathbf{X})$ & \citep{CandesRecht2009}\\
truncated nuclear norm & $\|\mathbf{X}\|_{\text{TN}}=\sum_{i=r+1}^m\sigma_i(\mathbf{X})$  & \citep{hu2012fast}\\
weighted nuclear norm & $\|\mathbf{X}\|_{\text{WN}}=\sum_{i=1}^mw_i\sigma_i(\mathbf{X})$  &\citep{gu2014weighted}\\
Schatten-$p$ quasi-norm & $\|\mathbf{X}\|_{\text{Sp}}=(\sum_{i=1}^m\sigma_i^p(\mathbf{X}))^{1/p}$  & \citep{MC_ShattenP_AAAI125165}\\
weighted Schatten-$p$ quasi-norm & $\|\mathbf{X}\|_{\text{WSp}}=(\sum_{i=1}^m w_i\sigma_i^p(\mathbf{X}))^{1/p}$ & \citep{7539605}\\
generalized non-convex penalty  & $R(\mathbf{X})=\sum_{i=1}^m g_{\lambda}(\sigma_i(\mathbf{X}))$ & \citep{lu2014generalized}\\ 
\bottomrule
\end{tabular}
\end{table}

\begin{table}
\centering
\caption{Factored rank regularizers}\label{tab_regularizer_2}
\resizebox{\textwidth}{!}{
\begin{tabular}{c|c|c|c}
\toprule
regularizer & definition & property ($\mathbf{X}=\mathbf{AB}$) & reference \\ \midrule
factored nuclear norm &  $R(\mathbf{A},\mathbf{B})=\frac{1}{2}(\|\mathbf{A}\|_F^2+\|\mathbf{B}\|_F^2)$ & $\min R(\mathbf{A},\mathbf{B})= \|\mathbf{X}\|_\ast$ & \citep{srebro2005maximum}\\
max norm & $R(\mathbf{A},\mathbf{B})=(\max _i\left\|\mathbf{A}_{i:}\right\|_2)(\max _j\left\|\mathbf{B}_{:j}\right\|_2)$ & --- &\cite{srebro2005rank}\\
factored Schatten-$1/2$ & $R(\mathbf{A},\mathbf{B})=\frac{1}{2}(\|\mathbf{A}\|_*+\|\mathbf{B}\|_*)$  & $\min R^2(\mathbf{A},\mathbf{B})=\|\mathbf{X}\|_{\mathrm{S}_{1 / 2}}$& \citep{shang2016tractable}\\
factored Schatten-$1/2$ & $R(\mathbf{A},\mathbf{B})=\frac{1}{2}(\|\mathbf{A}\|_{2,1}+\left\|\mathbf{B}^T\right\|_{2,1})$  & $\min R^2(\mathbf{A},\mathbf{B})=\|\mathbf{X}\|_{\mathrm{S}_{1 / 2}}$& \citep{fan2019factor}\\
factored Schatten-$2q/(2+q)$ & $R(\mathbf{A},\mathbf{B})=\frac{1}{q}\|\mathbf{A}\|_{2, q}^q+\frac{\alpha}{2}\left\|\mathbf{B}^T\right\|_F^2$  & $\min R(\mathbf{A},\mathbf{B})=c\|\mathbf{X}\|_{S_{2 q /(2+q)}}^{2 q /(2+q)}$& \citep{fan2019factor}\\
factored Schatten-$p$ & $R(\mathbf{A},\mathbf{B})=\frac{1}{2^p} \sum_{i=1}^d\left(\left\|\mathbf{A}_{:i}\right\|_2^2+\left\|\mathbf{B}_{i:}\right\|_2^2\right)^p$& $\min R(\mathbf{A},\mathbf{B})=\|\mathbf{X}\|_{S_{p}}^p$& \citep{giampouras2020novel}\\
\bottomrule
\end{tabular}
}
\begin{minipage}{\textwidth}
\footnotesize
\textit{Note:} $c=(1 / 2+1 / q) \alpha^{q /(q+2)}, q=1,\frac{1}{2},\frac{1}{4},\ldots$.
\end{minipage}
\end{table}

Beyond general low-rank matrices, several LRMC methods are designed for specific types. Notable examples include distance matrix completion \citep{alfakih1999solving}, covariance matrix completion \citep{hippert2022robust}, kernel matrix completion \citep{tsuda2003algorithm}, and one-bit matrix completion \citep{davenport20141}. In the latter case, the observed data are binary (1-bit) measurements rather than precise numerical values. This approach is valuable in scenarios with highly quantized data or binary feedback, such as in recommender systems and social networks.

\subsubsection{High-Rank Matrix Completion}

LRMC relies on the assumption that the matrix to be recovered is either low-rank or can be effectively approximated by a low-rank matrix. However, this assumption may not hold when data are generated from a union of low-dimensional subspaces or nonlinear low-dimensional latent variable models. Specifically, let $\mathbf{U}_j\in\mathbb{R}^{m\times r_j}$ be the basis matrix of subspace $\mathcal{S}_j$ and $\mathbf{Z}_j\in\mathbb{R}^{r_j\times n_j}$ be a random matrix, where $j=1,\ldots,s$. We can obtain the following data matrix
\begin{equation}\label{eq_uos}
    \mathbf{X}=[\mathbf{U}_1\mathbf{Z}_1,\mathbf{U}_2\mathbf{Z}_2,\ldots,\mathbf{U}_s\mathbf{Z}_s]\in\mathbb{R}^{m\times n}
\end{equation}
where $n=\sum_{j=1}^s{n_j}$. The rank of $\mathbf{X}$ can be as high as $\min\{m,n,\sum_{j=1}^sr_j\}$. If $s$ is large and the $s$ subspaces are different, $\mathbf{X}$ will be full-rank. In this case, it is impossible to recover the missing values of $\mathbf{X}$ using LRMC methods.

Regarding the nonlinear latent variable model, the columns of $\mathbf{X}$ are assumed to be given by
\begin{equation}\label{eq_nlm}
    \mathbf{x}_i=f(\mathbf{z}_i),\quad \mathbf{z}_i\sim\mathcal{D}\subset \mathbb{R}^d
\end{equation}
where $f:\mathbb{R}^d\rightarrow\mathbb{R}^m$ is a nonlinear function, $d\ll m$, and $\mathcal{D}$ denotes some distribution. Due to the nonlinearity of $f$, $\mathbf{X}$ could be full-rank. For instance, suppose $f(\mathbf{z})=[z,z^2,z^3]^\top$, where the latent variable $z$ is one-dimensional, and the rank of $\mathbf{X}$ is three, although the latent dimension is only one. 

\begin{figure}[h]
    \centering
    \includegraphics[width=0.8\linewidth]{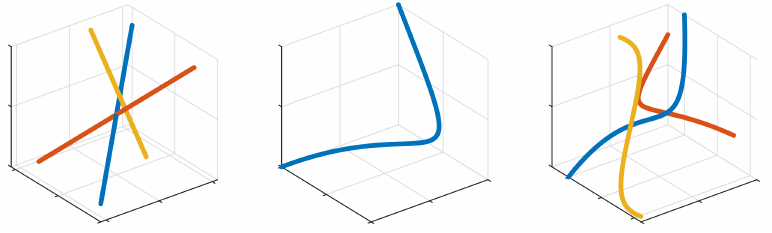}
    \caption{Examples \citep{fan2020polynomial} of data forming high-rank matrices in
3D space (left: union of subspaces; middle: one nonlinear manifold; right: union of nonlinear manifolds).}
    \label{fig_hrmc}
\end{figure}

\begin{figure}[h]
    \centering
    \includegraphics[width=1\linewidth]{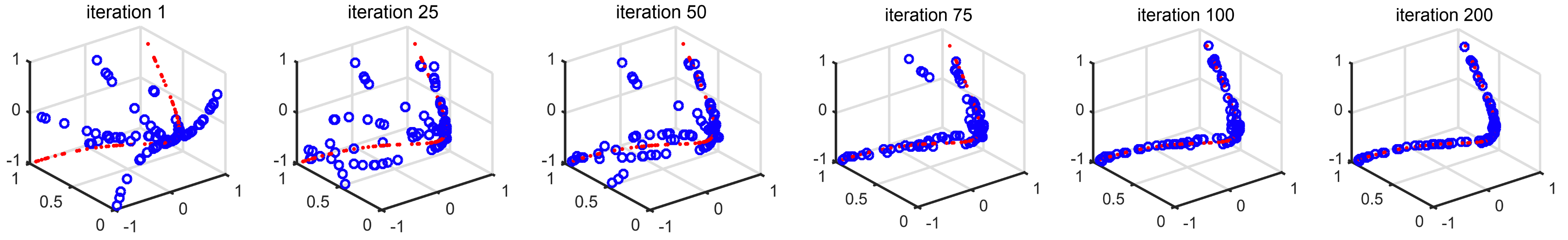}
    \caption{An example of HRMC on synthetic data. It shows the performance of the KFMC algorithm of \citep{fan2019online} in different iterations. The red points denote the true complete data, while the blue points denote the recovered data.}
    \label{fig_hrmc_kfmc}
\end{figure}

The nonlinear latent variable model \eqref{eq_nlm} can be extended to the following union of nonlinear manifolds:
\begin{equation}\label{eq_uonlm}
    \mathbf{x}_i^{(j)}=f^{(j)}(\mathbf{z}_i^{(j)}),\quad \mathbf{z}^{(j)}_i\sim\mathcal{D}^{(j)}\subset \mathbb{R}^{d_j},\quad j=1,2,\ldots,s
\end{equation}
where $f^{(1)},\ldots,f^{(s)}$ are different nonlinear functions.
This forms a matrix $\mathbf{X}=[\mathbf{X}^{(1)},\mathbf{X}^{(2)},\ldots,\mathbf{X}^{(s)}]$ of size $m\times n$, where $\mathbf{X}^{(j)}=[\mathbf{x}_1^{(j)},\ldots,\mathbf{x}^{(j)}_{n_j}]\in\mathbb{R}^{m\times n_j}$ and $n=\sum_{j=1}^s n_j$. It can be seen that \eqref{eq_uonlm} degrades to \eqref{eq_nlm} if $s=1$ and degrades to \eqref{eq_uos} if all $f^{(j)}$ are linear.
Figure \ref{fig_hrmc} provides synthetic examples for \eqref{eq_uos},\eqref{eq_nlm}, and \eqref{eq_uonlm}, respectively. The completion of the matrix produced by \eqref{eq_uonlm} poses the greatest challenge. Figure \ref{fig_hrmc_kfmc} shows an example of completing a matrix generated by \eqref{eq_nlm}.

In recent years, a few high-rank matrix completion (HRMC) methods have been proposed \citep{ErikssonBalzanoNowak2011,NLMC2016,NIPS2016_6357,fan2018matrix,fan2020polynomial,NEURIPS2020_42ae1544}.
They can be organized into three categories: subspace-based methods, kernel-based methods, and deep learning-based methods.

\paragraph{Subspace-Based Methods} These methods were proposed for the data given by \eqref{eq_uos} and usually learn the subspaces from the incomplete data \citep{ErikssonBalzanoNowak2011,8469061} or exploit the self-expressive model \citep{yang2015sparse,NIPS2016_6357,fan2017sparse,FAN2017290}, where each data point can be well represented by other data points drawn from the same subspace. The self-expressive model \citep{elhamifar2013sparse} can be formulated as
\begin{equation}
    \mathbf{X}=\mathbf{X}\mathbf{S}
\end{equation}
where $\mathbf{S}\in\mathbb{R}^{n\times n}$ is the coefficient matrix and is usually assumed to be sparse. Given an incomplete data matrix $\tilde{\mathbf{X}}$, \citet{FAN2017290} proposed the following matrix completion method:
\begin{equation}
\begin{aligned} 
\mathop{\text{minimize}}_{\hat{\mathbf{X}},\mathbf{S}}&~\|\mathbf{S}\|_{\ell_S}+
\frac{\lambda}{2}\|\hat{\mathbf{X}}-\hat{\mathbf{X}}\mathbf{S}\|_F^2\\ 
\text {subject to} &~\mathcal{P}_{\Omega}(\hat{\mathbf{X}})=\mathcal{P}_{\Omega}(\tilde{\mathbf{X}})
\end{aligned} 
\end{equation}
where $\|\cdot\|_{\ell_{S}}$ could be the $\ell_1$ norm, Frobenius norm, or nuclear norm.

\paragraph{Kernel-Based Methods} These methods \citep{NLMC2016,pmlr-v70-ongie17a,FANNLMC,fan2019online,fan2020polynomial,ongie2021tensor} take advantage of kernel techniques that can convert a nonlinear problem in the original data space to a linear problem in the high-dimensional feature space. \cite{pmlr-v70-ongie17a} and \cite{FANNLMC} proposed to solve
\begin{equation}
    \begin{aligned}
\mathop{\text{minimize}}_{\hat{\mathbf{X}}}&~\text{trace}(\mathbf{K}^{p/2})\\ 
\text {subject to} &~\mathcal{P}_{\Omega}(\hat{\mathbf{X}})=\mathcal{P}_{\Omega}(\tilde{\mathbf{X}})
    \end{aligned}
\end{equation}
where $\mathbf{K}\in\mathbb{R}^{n\times n}$ is a kernel matrix with $K_{i,j}=k(\hat{\mathbf{x}}_i,\hat{\mathbf{x}}_j)$ and $k$ is a kernel function such as the Gaussian kernel $k(\hat{\mathbf{x}}_i,\hat{\mathbf{x}}_j)=\exp(-\gamma\|\hat{\mathbf{x}}_i-\hat{\mathbf{x}}_j\|^2)$. The motivation is that the feature matrix $\phi(\hat{\mathbf{X}})$ induced by $k$ is approximately low-rank when $\hat{\mathbf{X}}$ has some nonlinear latent structure and $\|\phi(\hat{\mathbf{X}})\|_{S_p}^p=\text{trace}((\phi^\top(\hat{\mathbf{X}})\phi(\hat{\mathbf{X}}))^{p/2})=\text{trace}(\mathbf{K}^{p/2})$. 

\cite{fan2019online} showed that $\text{rank}(\phi(\mathbf{X}))=\min\{\binom{m+q}{q},n,s\binom{d+p q}{p q}\}$ if $f^{(j)}$ are $p$-order polynomials and $\phi$ is the feature map of a $q$-order polynomial kernel. When $n$ is sufficiently large and $p$ and $q$ are relatively small, the rank of $\phi(\mathbf{X})$ is low compared to its dimensions. \citet{fan2019online} introduced a kernel factorization method for HRMC and an online algorithm, which are more efficient than the methods proposed in \citep{pmlr-v70-ongie17a,FANNLMC}.

\paragraph{Deep Learning-Based Methods} These methods \citep{FAN2017540,fan2018matrix,seyedi2023elastic} are based on neural networks and will be detailed in Section \ref{sec_deep_impute}.

\subsection{Deep Learning-Based Missing Data Imputation}\label{sec_deep_impute}

Deep learning-based methods for missing data imputation \citep{KHAN2022278} can be categorized into three groups: autoencoder-based, deep matrix factorization, and deep generative model-based methods. These categories are detailed in the subsequent sections.

\subsubsection{Autoencoder-Based Methods}
An autoencoder (AE) \citep{hinton2006reducing} is a type of artificial neural network \citep{lecun2015deep} used for unsupervised learning of efficient codings. The goal of an AE is to learn a representation (encoding) for a set of data, typically for the purpose of dimensionality reduction or feature learning. Given a dataset with $n$ data points of $m$-dimension, a standard stacked or deep AE is composed of an encoder network $h_{\theta}:\mathbb{R}^m\rightarrow\mathbb{R}^d$ and a decoder network $g_{\phi}:\mathbb{R}^d\rightarrow\mathbb{R}^m$, which is often trained by solving the following optimization
\begin{equation}
    \mathop{\text{minimize}}_{\theta,\phi}~\frac{1}{n}\sum_{i=1}^n\|\mathbf{x}_i-g_{\phi}(h_\theta(\mathbf{x}_i))\|^2
\end{equation}
$\mathbf{z}=h_\theta(\mathbf{x})$ is usually called the representation of $\mathbf{x}$. Note that for tabular data, $h_{\theta}$ and $g_{\phi}$ are usually multilayer perceptrons, e.g., $h_{\theta}(\mathbf{x})=\mathbf{W}_2\sigma(\mathbf{W}_1\mathbf{x}+\mathbf{b}_1)+\mathbf{b}_2$, where $\theta=\{\mathbf{\mathbf{W}_1,\mathbf{b}_1,\mathbf{W}_2,\mathbf{b}_2}\}$ are the set of parameters to optimize and $\sigma$ denote the activation function. For image data, $h_{\theta}$ and $g_{\phi}$ are usually based on convolutional layers. Variants of the standard autoencoder include the denoising autoencoder (DAE) \citep{vincent2010stacked}, the sparse autoencoder \citep{ng2011sparse}, and the variational autoencoder (VAE) \citep{kingma2013auto}.

\begin{figure}[h]
     \centering
         \includegraphics[width=0.75\textwidth]{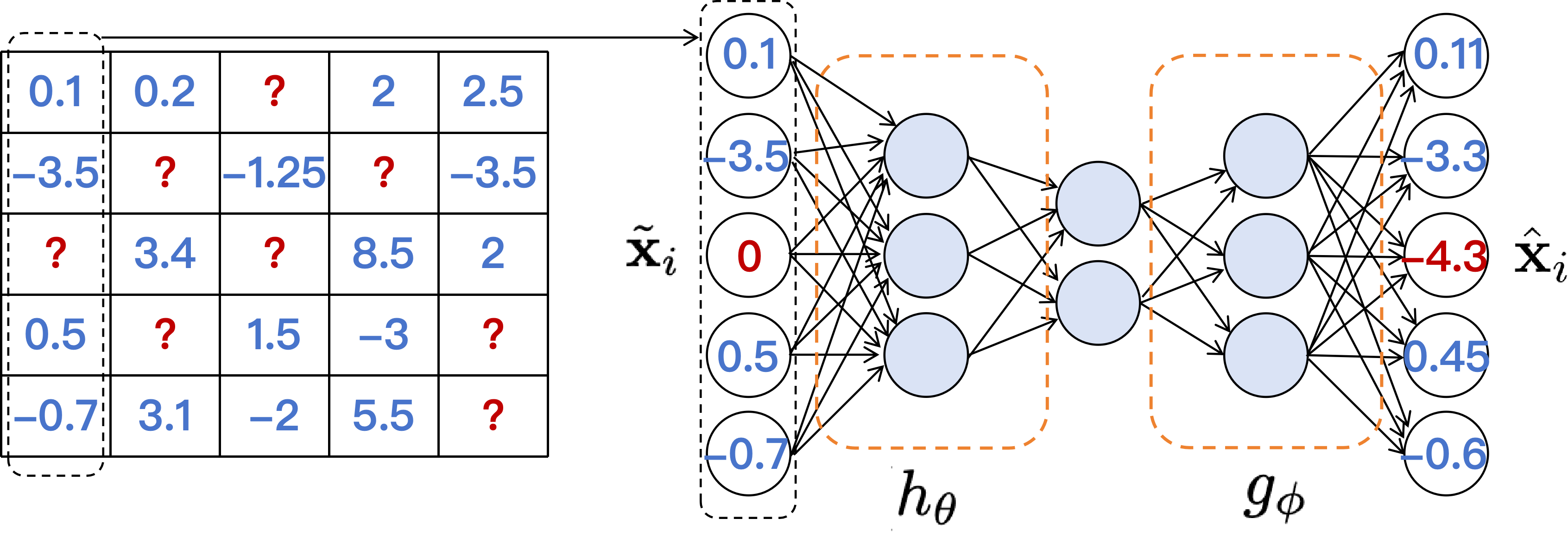}
\caption{A simple example of AE-based imputation. The rows of the matrix correspond to attributes. The neural network has 3 hidden layers. The missing values in the input layer are set to zero.}\label{fig_ae}
\end{figure}

\citet{vincent2008} used DAE to recover missing values of data, but it requires complete data during training. Applying AE or DAE to incomplete data directly, where the reconstruction errors for the observed values are minimized, works in practice \citep{beaulieu2017missing,ryu2020denoising}:
\begin{equation}
    \mathop{\text{minimize}}_{\theta,\phi}~\frac{1}{n}\sum_{i=1}^n\|(\tilde{\mathbf{x}}_i-g_{\phi}(h_\theta(\tilde{\mathbf{x}}_i)))\odot\boldsymbol{\omega}_i\|^2
\end{equation}
where $\boldsymbol{\omega}_i\in\{0,1\}^m$ is the missing value indicator for $\mathbf{x}_i$ and the missing values in the input are often temporarily filled with zeros. Figure \ref{fig_ae} shows an example of AE-based imputation.
However, the input bias caused by the missing values could be harmful. \citet{FAN2017540} proposed a deep learning based matrix completion method, which optimizes the missing values and the autoencoder parameters simultaneously and hence can reduce the input bias caused by the missing values. 
\citet{gondara2018mida} proposed a multiple imputation model based on DAE. 
\citet{pereira2020reviewing} wrote a review paper of AE and its variants based missing data imputation methods. \cite{fan2024neuronenhanced} proposed a neuron-enhanced autoencoder for missing data imputation, where the activation function of the output layer is learned adaptively by a neural network to model the complex response function in real data.
\citet{du2024remasker} proposed a simple yet effective imputation method called ReMasker. In ReMasker, besides the missing values (i.e., naturally
masked), another set of values is randomly re-masked, and the autoencoder is optimized by reconstructing the re-masked set. ReMasker performs on par with or outperforms many complex and strong competitors in terms of both imputation fidelity and utility under various missingness settings.

\subsubsection{Deep Matrix Factorization-Based Methods}

AE-based imputation models need to learn to reduce the negative impact of the missing values in the input of the encoder, which can be difficult in practice. To address the limitation, \citep{fan2018matrix} proposed deep matrix factorization (DMF) based matrix completion. Instead of a linear and shallow factorization like \eqref{eq_lrmc_mf}, they solve the following deep factorization problem
\begin{equation}\label{eq_dmf}
\begin{aligned}
\underset{\mathbf{Z},\mathbf{W}_1,\ldots,\mathbf{W}_L}{\operatorname{minimize}}~& \|\mathcal{P}_{\Omega}(\tilde{\mathbf{X}}-\mathbf{W}_L(\sigma(\mathbf{W}_{L-1}\cdots\sigma(\mathbf{W}_1\mathbf{Z})\cdots)))\|_F^2+\lambda \cdot R(\mathbf{Z},\mathbf{W}_1,\ldots,\mathbf{W}_L)\\
\end{aligned}
\end{equation}
where the bias terms of the neural network are omitted for simplicity and ${R}$ denotes some regularization on the latent matrix $\mathbf{Z}\in\mathbb{R}^{d\times n}$ and the weight matrices. One limitation of this model is that the latent dimension $d$ is difficult to determine in real-world applications. There have been a few extensions of DMF-based missing data imputation \citep{zhang2019deep,li2022regularised,li2023h2tf,ledent2024generalization,ji2025deep}.

\subsubsection{Deep Generative Model-Based Methods}

Deep generative models are a class of neural network architectures designed to learn the underlying distribution of a dataset and generate new data samples that resemble the training data. These models are particularly useful for tasks such as image synthesis, text generation, and data augmentation. They have gained significant attention in recent years due to their ability to produce high-quality and realistic data. Key types of deep generative models include Variational Autoencoders (VAEs) \citep{kingma2013auto}, Generative Adversarial Networks (GANs) \citep{goodfellow2014generative,gulrajani2017improved}, Normalizing Flows (NFs) \citep{dinh2014nice,kingma2016improved}, Autoregressive Models \citep{van2016pixel,salimans2017pixelcnn++}, and Diffusion models (DMs) \citep{sohl2015deep,ho2020denoising}.

\paragraph{VAE-Based Imputation}
The basic VAE \citep{kingma2013auto} solves the following problem
\begin{equation}
    \mathop{\text{maximize}}_{\theta,\phi}~\mathcal{L}(\theta, \phi)=\mathbb{E}_{\mathbf{z} \sim q_\phi(\mathbf{z} \mid \mathbf{x})}\left[\log p_\theta(\mathbf{x} \mid \mathbf{z})\right]-D_{\mathrm{KL}}\left(q_\phi(\mathbf{z} \mid \mathbf{x}) \| p(\mathbf{z})\right)
\end{equation}
The encoder $q_\phi(\mathbf{z} \mid \mathbf{x})$ approximates the posterior distribution over latent variables $\mathbf{z}\in\mathbb{R}^d$ given input $\mathbf{x}\in\mathbb{R}^m$ and outputs the parameters of a prior distribution (e.g., the mean and variance of a Gaussian).
The decoder $p_\theta(\mathbf{x} \mid \mathbf{z})$ reconstructs the input $\mathbf{x}$ from latent variable $\mathbf{z}$, since maximizing $\log p_\theta(\mathbf{x} \mid \mathbf{z})$ is equivalent to minimizing the reconstruction error.
The KL Divergence $D_{\mathrm{KL}}$ regularizes the latent space by pushing $q_\phi(\mathbf{z} \mid \mathbf{x})$ toward a prior $p(\mathbf{z})$ (e.g., standard Gaussian $\mathcal{N}(\boldsymbol{0}, \mathbf{I})$). $\mathcal{L}(\theta, \phi)$ is a lower bound of the log-likelihood $\log p_{\theta}(\mathbf{x})$. 
Once the model is trained, new data points can be generated by sampling a latent vector $\mathbf{z}$ from the prior $p(\mathbf{z})$
and passing it through the decoder. The core VAE framework has inspired several variants \citep{burda2015importance,higgins2017beta,pmlr-v97-mathieu19a}.

The MIWAE method proposed by \citep{mattei2019miwae} adapts the importance-weighted autoencoder (IWAE) \citep{burda2015importance} to a missing data imputation model, where IWAE shares the same architecture as VAE but has a different training objective and a tighter lower bound of the log-likelihood. Specifically, letting $\mathbf{x}_i^{+}$ and $\mathbf{x}_i^{-}$ be the observed and missing parts of $\mathbf{x}_i$, respectively, MIWAE solves
\begin{equation}\label{eq_MIWAE}
    \begin{aligned} 
    \mathop{\text{maximize}}_{\theta,\phi}~\mathcal{L}_K({\theta}, {\phi})=\sum_{i=1}^n \mathbb{E}_{\mathbf{z}_{i 1}, \ldots, \mathbf{z}_{i K} \sim q_{{\phi}}\left(\mathbf{z} \mid \mathbf{x}_i^{+}\right)}\left[\log \frac{1}{K} \sum_{k=1}^K \frac{p_{{\theta}}\left(\mathbf{x}_i^{+} \mid \mathbf{z}_{i k}\right) p\left(\mathbf{z}_{i k}\right)}{q_{{\phi}}\left(\mathbf{z}_{i k} \mid \mathbf{x}_i^{+}\right)}\right]\end{aligned}
\end{equation}
where $K\geq 1$ is a hyperparameter. When $K=1$, the model becomes a standard VAE on incomplete data. In \eqref{eq_MIWAE}, for example, one can let $q_\phi\left(\mathbf{z} \mid \mathbf{x}^{+}\right)=\mathcal{N}(\mathbf{z}; \boldsymbol{\mu}_\phi,\text{diag}(\boldsymbol{\sigma}^2_\phi))$, where $\boldsymbol{\mu}_\phi$ and $\boldsymbol{\sigma}^2_\phi$ are the outputs of the encoder network $h_\phi\left(\iota\left(\mathbf{x}^{+}\right)\right)$ and $\iota$ denotes a simple imputation (e.g., zero imputation) function transforming $\mathbf{x}^{+}$ into a complete input vector of $m$ dimension. In the inference phase, for each $\mathbf{x}^{+}$, it samples $\mathbf{z}_{(1)},\mathbf{z}_{(2)},\ldots,\mathbf{z}_{(L)}$ from $q_\phi\left(\mathbf{z} \mid \mathbf{x}^{+}\right)$, and for each $\mathbf{z}_{(l)}$, it samples an $\mathbf{x}_{(l)}^{-}$ from $p_{{\theta}}(\mathbf{x}^{-} \mid \mathbf{z}_{(l)})$. Finally, the missing values of $\mathbf{x}$ are imputed as $\hat{\mathbf{x}}^{-}=\sum_{l=1}^Lw_l\mathbf{x}_{(l)}^{-}$, where $w_l$ are the weights and will not be detailed here.

There are more VAE-based imputation methods. For example, \citet{nazabal2020handling} proposed HI-VAE for heterogeneous (mixed continuous and discrete) data imputation. \citet{collier2020vaes} proposed a VAE-based imputation method that explicitly models missing values via a latent corruption process, enabling principled handling of both MCAR and MNAR data by conditioning the encoder and decoder on missingness indicators. \citet{fortuin2020gp} proposed a method called GP-VAE for time series imputation.


\paragraph{GAN-Based Imputation}

The basic GAN \citep{goodfellow2014generative} solves the following problem
\begin{equation}
    \min _G \max _D V(G,D)=\mathbb{E}_{\mathbf{x} \sim p_{\text{data}}(\mathbf{x})}[\log D(\mathbf{x})]+\mathbb{E}_{\mathbf{z} \sim p_{\mathbf{z}}(\mathbf{z})}[\log (1-D(G(\mathbf{z})))]
\end{equation}
where $G$ and $D$ are the generator and discriminator, respectively. 
$G$ is a neural network that learns to generate fake data from random noise $\mathbf{z}$, usually drawn from a Gaussian distribution. It tries to minimize $V(D, G)$ (fool the discriminator).
$D$, usually a binary classifier, learns to distinguish real data ($\mathbf{x}$, labeled as $1$) from fake data ($G(\mathbf{z})$, labeled as $0$). It tries to maximize $V(D, G)$. Representative variants of GAN include cycle-GAN \citep{zhu2017unpaired}, Wasserstein GAN \citep{arjovsky2017wasserstein}, etc.

A wide range of GAN-based methods have been developed for missing data imputation \citep{shang2017vigan,yoon2018gain,luo2018multivariate,li2019misgan,yang2020adversarial,gupta2020time,zhang2021missing,dong2021generative,xia2021recovering,kazemi2021igani,wang2022sta,qin2023imputegan,kang2023cm,wang2024time,wang2025generative}. \cite{shahbazian2023generative} provided a review of GAN-based imputation methods, including a numerical performance evaluation of several models. A common paradigm in these approaches is to use the generator as an imputation network, which takes as input either a random latent vector $\mathbf{z}$, the incomplete data $\mathbf{x}$, or a combination of both with additional contextual information.
Taking the GAIN proposed by \citep{yoon2018gain} as an example, let $\tilde{\mathbf{x}}$, $\boldsymbol{\omega}$, and $\mathbf{z}$ be the incomplete data vector, binary mask vector, and random noise, respectively. The generator in GAIN is designed to impute missing data, i.e., $\hat{\mathbf{x}}=G(\tilde{\mathbf{x}}, \boldsymbol{\omega},(\mathbf{1}-\boldsymbol{\omega}) \odot \mathbf{z})$, while the discriminator is tasked with determining whether a particular element is observed or imputed, i.e., $\hat{\boldsymbol{\omega}}=D(\bar{\mathbf{x}},\mathbf{h})$,
where $\bar{\mathbf{x}}=\boldsymbol{\omega} \odot \tilde{\mathbf{x}}+(\mathbf{1}-\boldsymbol{\omega}) \odot \hat{\mathbf{x}}$ and $\mathbf{h}$ is a hint vector generated by letting one or more randomly selected entries of $\boldsymbol{\omega}$ be $0.5$. The loss functions for the discriminator and generator are as follows: 
\begin{align}
    \mathcal{L}_D(\boldsymbol{\omega}, \hat{\boldsymbol{\omega}}, \mathbf{h})=&\sum_{j: h_j=0.5} {\left(\omega_j \log \left(\hat{\omega}_j\right)\right.}\left.+\left(1-\omega_j\right) \log \left(1-\hat{\omega}_j\right)\right)\\
    \mathcal{L}_G(\boldsymbol{\omega}, \hat{\boldsymbol{\omega}}, \mathbf{h},\tilde{\mathbf{x}},\hat{\mathbf{x}})=&-\sum_{j: h_j=0.5} \left(1-\omega_j\right) \log \left(\hat{\omega}_j\right)+\alpha\sum_{j=1}^m\omega_j\ell(\tilde{{x}}_j,\hat{{x}}_j)
\end{align}
where $\ell$ denotes a reconstruction loss such as the square loss and $\lambda>0$ is a trade-off hyperparameter. One of the key improvements of GAIN over traditional GANs is that the discriminator's output in GAIN is an estimated mask matrix that matches the size of the original data, rather than a simple binary value. Additionally, the GAIN structure incorporates a hint matrix to provide supplementary information about the mask matrix, ensuring that the generated fake data adheres to the actual data distribution.

\paragraph{Normalizing Flow-Based Imputation} Normalizing flows \citep{dinh2014nice,kingma2016improved}, which transform a simple base distribution $q(\mathbf{z})$ (e.g., a standard Gaussian) into a complex target distribution $p(\mathbf{x})$ through a bijective function $\mathbf{x}=g(\mathbf{z})$:
\begin{equation}
    p(\mathbf{x})=q\left(g^{-1}(\mathbf{x})\right) \cdot\left|\operatorname{det}\left(\frac{\partial g^{-1}(\mathbf{x})}{\partial \mathbf{x}}\right)\right|
\end{equation}
It solves the following log-likelihood maximization problem:
\begin{equation}
    \mathop{\text{maximize}}_\theta~ \mathbb{E}_{\mathbf{x} \sim p_{\text{data}}(\mathbf{x})}\left[\log q\left(g_\theta^{-1}(\mathbf{x})\right)+\log \left|\operatorname{det}\left(\frac{\partial g_\theta^{-1}(\mathbf{x})}{\partial \mathbf{x}}\right)\right|\right]
\end{equation}
where $\theta$ denotes the parameters of the invertible neural network $g$.

Normalizing flows are used for density estimation, generative modeling, and variational inference, offering exact likelihood evaluation. It has been adapted to missing data imputation \citep{richardson2020mcflow,ma2021emflow}. For instance, \citet{richardson2020mcflow} introduced MCFlow, an effective deep imputation approach that employs normalizing flow generative models and Monte Carlo sampling. Specifically, let $\tilde{\mathbf{x}}$ be an incomplete data and $\boldsymbol{\omega}$ be the corresponding mask vector. Let $\hat{\mathbf{x}}=g_\theta(\hat{\mathbf{z}})$ be the imputation for $\tilde{\mathbf{x}}$, where $\hat{\mathbf{z}}=h_{\phi}(\bar{\mathbf{z}})$ denotes the post-processing for $\bar{\mathbf{z}}$ by a neural network $h_\phi$. $\bar{\mathbf{z}}$ is given by $\bar{\mathbf{z}}=g_\theta^{-1}(\bar{\mathbf{x}})=f_{\theta}(\bar{\mathbf{x}})$, where
$\bar{\mathbf{x}}=\boldsymbol{\omega} \odot \tilde{\mathbf{x}}+(\mathbf{1}-\boldsymbol{\omega}) \odot \hat{\mathbf{x}}$.
MCFlow alternately refines the density estimate and imputes missing values by 
\begin{align}
&\mathop{\text{maximize}}_\theta~\frac{1}{n} \sum_{i=1}^{n}\left[\log q\left(f_\theta\left(\bar{\mathbf{x}}_i\right)\right)+\log \left|\operatorname{det} \left(\frac{\partial f_\theta\left(\bar{\mathbf{x}}_i\right)}{\partial \bar{\mathbf{x}}_i}\right)\right|\right]\\
&\mathop{\text{minimize}}_\phi~\frac{1}{n} \sum_{i=1}^{n}\left[\|(\bar{\mathbf{x}}_i- \hat{\mathbf{x}}_i)\odot \boldsymbol{\omega}_i\|^2-\lambda \log p\left(\hat{\mathbf{x}}_i\right)\right]
\end{align}
where $\lambda>0$ is a hyperparameter. Please note that the notation used here differs from that in \citep{richardson2020mcflow}. This change was made to maintain consistency with the mathematical formulations of the previously introduced methods.

\paragraph{Diffusion Model-Based Imputation}
Diffusion models \citep{sohl2015deep,ho2020denoising,song2020score,yang2023diffusion} are generative models that learn to generate data (e.g., images, audio) by gradually denoising random noise into structured samples. They consist of two key processes. The first is the forward (diffusion) process, which gradually corrupts the training data with Gaussian noise over many steps, transforming it into pure noise, i.e., at each step $t$, 
\begin{equation}
q\left(\mathbf{x}_t \mid \mathbf{x}_{t-1}\right)=\mathcal{N}\left(\mathbf{x}_t ; \sqrt{1-\beta_t} \mathbf{x}_{t-1}, \beta_t \mathbf{I}\right)
\end{equation}
where $\beta_t$ is the noise schedule controlling how fast data is destroyed, $t=1,2,\ldots, T$, $T$ is usually a large number such as 1000, and $\mathbf{x}_0$ is the original clean data point.
The second process is the reverse process (denoising), in which a neural network with parameters $\theta$ learns to reverse the noise addition, i.e., 
\begin{equation}
p_\theta\left(\mathbf{x}_{t-1} \mid \mathbf{x}_t\right)=\mathcal{N}\left(\mathbf{x}_{t-1} ; \mu_\theta\left(\mathbf{x}_t, t\right), \Sigma_\theta\left(\mathbf{x}_t, t\right)\right)
\end{equation}
The model is trained by maximizing the evidence lower bound of the log-likelihood, i.e.,
\begin{equation}
   \mathop{\text{maximize}}_{\theta}~\mathbb{E}_{q\left(\mathbf{x}_{1: T} \mid \mathbf{x}_0\right)}\left[\log p_\theta\left(\mathbf{x}_0 \mid \mathbf{x}_1\right)-\sum_{t=2}^T D_{\mathrm{KL}}\left(q\left(\mathbf{x}_{t-1} \mid \mathbf{x}_t, \mathbf{x}_0\right) \| p_\theta\left(\mathbf{x}_{t-1} \mid \mathbf{x}_t\right)\right)\right]
\end{equation}
After training, the model can generate new data by starting from random noise and iteratively denoising it.

Since diffusion models are often more effective than other generative models, such as VAEs and GANs, in generating high-quality new samples, a few researchers have attempted to extend them to missing data imputation. For instance, 
\citet{tashiro2021csdi} proposed CSDI, a diffusion model-based time series imputation method, by modeling the following distribution
\begin{equation}\label{eq_CSDI}
    p_\theta\left(\mathbf{x}_{t-1}^{\text{ta}} \mid \mathbf{x}_t^{\text{ta}}, \mathbf{x}_0^{\text{co}}\right)=\mathcal{N}\left(\mathbf{x}_{t-1}^{\text{ta}} ; \mu_\theta\left(\mathbf{x}_t^{\text{ta}}, t \mid \mathbf{x}_0^{\text{co}}\right), \Sigma_ \theta\left(\mathbf{x}_t^{\text{ta}}, t \mid \mathbf{x}_0^{\text{co}}\right)\right)
\end{equation}
In \eqref{eq_CSDI}, $\mathbf{x}_0^{\text{ta}}$ and $\mathbf{x}_0^{\text{co}}$ were constructed by splitting the observed entries of $\mathbf{x}$ into a target part and a condition part, where the missing values were replaced by zeros. The neural network parameterized by $\theta$ learns to remove the random noise added to $\mathbf{x}_{t-1}^{\text{ta}}$. After learning $\theta$, the imputation for the missing entries of $\mathbf{x}$ is implemented by iteratively denoising the missing entries (initialized by a Gaussian) conditioned on the observed entries of $\mathbf{x}$.

Inspired by CSDI, \citet{zheng2022diffusion} applied diffusion models to tabular data imputation.
\citet{ouyang2023missdiff} proposed MissDiff for tabular data imputation.
\citet{chen2024rethinking} introduced negative entropy-regularized Wasserstein gradient flow for diffusion model-based imputation. \citet{jolicoeur2024generating} introduced
an approach for generating and imputing mixed-type (continuous and categorical) tabular data utilizing score-based diffusion and conditional flow matching. 
\citet{liu2024self} demonstrated that self-supervised pretraining enhances diffusion models for tabular data imputation by leveraging unlabeled data to learn richer feature representations before fine-tuning on missing-value tasks. 
\citet{zhang2025diffputer} proposed DiffPuter, which iteratively trains a diffusion model to learn the joint distribution of missing and observed data and performs an accurate conditional sampling to update the missing values using a tailored reversed sampling strategy.
\citet{ahamed2025refidiff} proposed RefiDiff that combines local machine learning predictions with a novel Mamba-based denoising network capturing interrelationships among distant features and samples, which improved the imputation performance of diffusion models.

\subsubsection{Graph Neural Network-Based Methods}
Graph neural networks (GNNs) are popular models for graph data. A graph is a collection of nodes (or vertices) connected
by edges (or links), and is usually used to represent
complex relationships and interactions between objects or
entities. A graph is usually denoted as $G=(V, E)$, where $V=\{v_1,v_2,\ldots,v_n\}$ denotes the vertex set and $E\subseteq V\times V$ denotes the edge set. Sometimes, the nodes have attributes, forming an attribute matrix $\mathbf{X}\in\mathbb{R}^{n\times d}$. The adjacency matrix of $G$ is usually denoted as $\mathbf{A}\in\{0,1\}^{n\times n}$, with $A_{ij}=1$ if $(i,j)\in E$ and $0$ otherwise. In some cases, the edges are weighted, meaning that $\mathbf{A}$ is not binary. Examples of graphs include chemical
molecules, biological networks, social networks, user-item interaction networks, and transportation
networks. 

Representative architectures of GNNs include graph convolutional network (GCN) \citep{kipf2017semi}, graph attention network GAT) \citep{velivckovic2017graph}, graph isomorphism network (GIN) \citep{xu2019powerful}, and graph transformer (GT) \citep{yun2019graph}. 
The hidden layers of a GCN can be formulated as 
\begin{equation}
    \mathbf{H}^{(l)}=\sigma\left(\tilde{\mathbf{D}}^{-\frac{1}{2}} \tilde{\mathbf{A}} \tilde{\mathbf{D}}^{-\frac{1}{2}} \mathbf{H}^{(l-1)} \mathbf{W}^{(l)}\right)
\end{equation}
where $\tilde{\mathbf{A}}=\mathbf{A}+\mathbf{I}$ is the adjacency matrix with self-connections, $\tilde{\mathbf{D}}$ is its degree matrix, $\mathbf{H}^{(l)}$ is the node feature matrix at layer $l$, $\mathbf{H}^{(0)}=\mathbf{X}$, $\mathbf{W}^{(l)}$ is the trainable weight matrix, and $\sigma$ is a nonlinear activation function. GNNs have shown impressive performance in many graph learning tasks such as node classification/clustering and graph classification/clustering.

Several studies have explored using Graph Neural Networks (GNNs) for missing data imputation \citep{zhang2019inductive,you2020handling,spinelli2020missing,telyatnikov2023egg}. A key challenge, however, is that GNNs typically require a pre-defined similarity graph across the entire dataset, which is difficult to construct from incomplete data. \citet{zhang2019inductive} proposed a GNN-based matrix completion method for collaborative filtering. \citet{you2020handling} proposed GRAPE, a graph-based framework for feature imputation as well as label prediction. GRAPE tackles the missing data problem using a graph representation, where the observations and features are viewed as two types of nodes in a bipartite graph, and the observed feature values as edges. Specifically, in GRAPE, for the incomplete data matrix $\tilde{\mathbf{X}}$ (with mask matrix $\mathbf{M}$) containing $n$ observations and $d$ attributes, a bipartite graph $G=(V_n\cup V_d, E)$ is constructed, where $V_n=\{u_1,u_2,\ldots,u_n\}$ denotes the set of nodes corresponding to the $n$ observations, $V_d=\{v_1,v_2,\ldots,v_d\}$ denotes the set of nodes corresponding to the $d$ attributes, and $E=\left\{\left(u_i, v_j,e_{u_i,v_j}\right): u_i\in V_n, v_j\in V_d, e_{u_i,v_j}=\tilde{x}_{ij},M_{i j}=1\right\}$ denotes the edge set. The prediction for the missing values in $\tilde{\mathbf{X}}$ is given by
\begin{equation}
    \hat{x}_{ij}=f_{\theta}(u_i,v_j;G,\mathbf{H}_n,\mathbf{H}_d), \quad\forall (i,j) \text{ with } M_{ij}=0
\end{equation}
where $f_{\theta}$ is a GNN such as GraphSAGE \citep{hamilton2017inductive}. $\mathbf{H}_n\in\{1\}^{n\times d}$ is the input embedding matrix for the $n$ sample nodes and $\mathbf{H}_d=\mathbf{I}_d$ is the input embedding matrix for the $d$ attribute nodes; they are updated in the subsequent layers via neighbor aggregation over $G$. The parameters $\theta$ of the GNN are then optimized using a loss function such as $\sum_{(i,j):M_{ij}=1}(\tilde{x}_{ij}-\hat{x}_{ij})^2$.

More recently, \citet{telyatnikov2023egg} proposed EGG-GAE, a GCN-based method for tabular data that adaptively constructs the adjacency matrix from a neural network's representations of the incomplete data. In a different domain, \citet{li2024stmcdi} combined a GNN with a diffusion model to impute missing values in spatial transcriptomics data. More methods of GNN-based imputation can be found in \citep{huang2022graph,wu2023overview,eskandari2024gn2di,zhang2025enhancing}.

It is worth noting that GNNs were also applied to link prediction and graph completion, which will be detailed in Section \ref{sec_lp_gc}.

\subsection{Large Language Model for Missing Data Imputation}\label{sec_LLM}

Large Language Models (LLMs) \citep{achiam2023gpt,naveed2025comprehensive} are billion-parameter neural networks that learn universal linguistic patterns by next-token prediction on web-scale text. Their recent ascendancy has redefined state-of-the-art across conversational AI, code synthesis, and scientific discovery. It is also promising to exploit LLMs in missing data imputation tasks. Indeed, the pretrained language models are already effective enough in predicting the missing words in pure text owing to the training strategy of next-token prediction, of which a typical application is code completion \citep{wei2023copiloting,izadi2024language,husein2025large}. For non-text data such as tabular data, a rudimentary approach is to feed the LLM an incomplete table as raw text and prompts it to hallucinate the blanks; while clever prompting can yield plausible guesses, this tactic barely scratches the surface of LLMs’ expressive power.

\citet{9458712} proposed a semantic-aware imputation framework, called IPM, that leverages pre-trained language models to fill missing categorical data. The key idea of IPM follows the regression imputation discussed in Section \ref{sec_regimpu}. Specifically, 
for each attribute $x_j$, letting $\mathcal{S}_j:=\{v_1,v_2,\ldots,v_{K_j}\}$ be the set of all possible values, IPM treats the imputation for $x_j$ as a $K_j$-class classification task. Let $\mathcal{D}$ be the set of data samples without missing values and $\bar{\mathbf{x}}^{(j)}:=\mathbf{x}\backslash x_j$ be data sample with attribute $x_j$ removed, where $\mathbf{x}\in\mathcal{D}$. The numerical embedding for each $\bar{\mathbf{x}}^{(j)}$ is obtained by a language model (LM) with parameters $\theta$, i.e., 
\begin{equation}\label{eq_textualization}
    \mathbf{z}^{(j)}=\text{LM}_{\theta}\left(\text{Textualize}\left(\bar{\mathbf{x}}^{(j)}\right)\right)
\end{equation}
where $\text{Textualize}(\cdot)$ denotes the operation to convert the input data into text that can be directly handled by the LM. In \citep{9458712}, a very simple textualization strategy is used. For example, if $\bar{\mathbf{x}}^{(j)}$ has three attributes including \textit{city}, \textit{hobby}, and \textit{age}, its textualization is [[CLS]; paris; football; 25; [SEP]], where [CLS] and [SEP] are special tokens in BERT \citep{devlin2019bert} or its variants. In \eqref{eq_textualization}, $\mathbf{z}^{(j)}$ is the embedding of [CLS] that summarizes the embeddings of the input tokens. 

Then IPM fine-tunes the LM and trains a classifier $f^{(j)}_\phi$ by
\begin{equation}
    \mathop{\text{minimize}}_{\theta,\phi}~\frac{1}{|\mathcal{D}|}\sum_{\mathbf{x}\in\mathcal{D}}\text{CE}\left(\text{Softmax}\left(f^{(j)}_\phi(\mathbf{z}^{(j)})\right),\mathbf{y}^{(j)}\right)
\end{equation}
where $\text{CE}$ denotes the cross-entropy loss and $\mathbf{y}^{(j)}$ is a one-hot label vector with $y_k^{(j)}=1$ if $x_j=v_k$. Once the models are well trained, the missing attribute $x_v$ in a test sample $\mathbf{x}$ is imputed as $v_k$ if the $k$-th element of $\text{Softmax}(f^{(j)}_\phi(\mathbf{z}^{(j)}))$ is largest. This method is called IPM-Multi. IPM has another variant called IPM-Binary, which models it as a binary candidate selection problem to mitigate over-fitting in low-redundancy datasets. 
By utilizing fine-tuned models like RoBERTa \citep{liu2019roberta} and neighbor-based negative sampling, IPM captures fine-grained token semantics, outperforming traditional and generative baselines across multiple real-world datasets. One limitation of IPM is that it requires a sufficient number of complete samples to train the classifiers. Another limitation is that it is not effective in inputting numerical data. Note that larger LMs and more sophisticated textualization and embedding summarization methods can be used for \eqref{eq_textualization}, which may improve the imputation performance.

More recently, \citet{zhang2023large} explored the use of LLMs like GPT-4 for data preprocessing tasks such as missing data imputation. The paper introduced a framework that combines prompt engineering techniques—such as zero-shot, few-shot, and batch prompting—with contextualization and feature selection to guide LLMs in imputing missing values. 
\citet{wang2024treb} introduces TREB, a BERT-based framework for imputing missing values in tabular data. By transforming numerical rows into tokenized text and fine-tuning RoBERTa with a custom 4-digit tokenizer, TREB captures feature interdependencies without relying on categorical semantics. 
In \citep{huang2024missing}, the authors proposed a missing data imputation framework called PRPMI. PRPMI enhances pre-trained language models by incorporating trainable prompt embeddings and retrieval-augmented knowledge from external data sources. It serializes tabular data into natural language format, fine-tunes the model using both original and retrieved contextual information, and treats imputation as a multiclass classification task. 
\citet{ding2024data} proposed to fine-tune a pre-trained GPT-2 model using complete data and prompting it to predict missing values in recommender systems. The method generates semantically meaningful imputations and outperforms a few traditional imputation methods on some benchmarks.
\citet{hayat2024context} presented a method called CRILM for context-aware missing data imputation that leverages pre-trained language models to generate meaningful, contextually relevant descriptors for missing data instead of relying on traditional numerical estimation. By transforming tabular data into natural language and fine-tuning lighter LLMs on these enriched datasets, CRILM effectively handles various missingness mechanisms (MCAR, MAR, and MNAR), and provides promising imputation performance.
\citet{fang2025spatiotemporal} proposed SPLLM, a spatiotemporal pretrained LLM designed for forecasting with missing values. SPLLM integrates a spatiotemporal fusion GCN module to extract spatial and temporal correlations, employs an FFN fine-tuning strategy to adapt pretrained LLMs to incomplete data, and uses a final fusion layer to combine learned and pretrained knowledge.

\subsection{Other Imputation Methods}
Some imputation methods fall outside the categories discussed in previous sections. For instance, \citet{muzellec2020missing} proposed a method based on optimal transport \citep{villani2008optimal}. Their core idea is that two randomly drawn batches from the same dataset should share the same distribution; this principle forms a loss function for imputation by matching distributions. Specifically, let $\theta$ be the missing values of the incomplete dataset $\tilde{\mathbf{X}}$, the method solves the following problem
\begin{equation}\label{eq_ot_miss}
    \mathop{\text{minimize}}_{\theta}~\mathbb{E}_{\tilde{\mathbf{X}}_A,\tilde{\mathbf{X}}_B\sim \mathbf{X}}[\text{Sinkhorn}(\tilde{\mathbf{X}}_A,\tilde{\mathbf{X}}_B)]
\end{equation}
where $\tilde{\mathbf{X}}_A$ and $\tilde{\mathbf{X}}_B$ are two batches drawn from $\tilde{\mathbf{X}}$ and Sinkhorn is the Sinkhorn divergence \citep{genevay2018learning}, an extension of the Wasserstein distance (see \eqref{eq_WD2}). Two variants of \eqref{eq_ot_miss} were also proposed in \citep{muzellec2020missing}.

Inspired by \citep{muzellec2020missing}, \citet{pmlr-v202-zhao23h} proposed imputing missing values by transforming two sample batches into a latent space using deep invertible functions and then matching their distributions. Traditional missing data imputation methods are simple and flexible but require precise model specification, while deep generative approaches are efficient but harder to optimize and rely on stronger assumptions. Motivated by this, \citet{jarrett2022hyperimpute} introduced HyperImpute, a framework that combines their strengths by adaptively configuring column-wise models.
The proposed method is implemented with ready-to-use tools and validated through extensive experiments, demonstrating superior accuracy over benchmarks and reinforcing the effectiveness of iterative imputation. 

Interestingly, \citet{awan2022reinforcement} proposed a reinforcement learning \citep{sutton1998reinforcement,kaelbling1996reinforcement} method for missing data imputation, by learning a policy through an action-reward-based experience. The method was compared with a few baselines such as GAIN and showed competitive performance in the experiments. One limitation of the method is that it performs imputation for each variable separately and hence cannot effectively utilize the correlation between variables.

\section{Missing Data Imputation for Special Data Format or Type}

\subsection{Tensor Completion}
\begin{figure}[h!]
    \centering
    \includegraphics[width=1\linewidth]{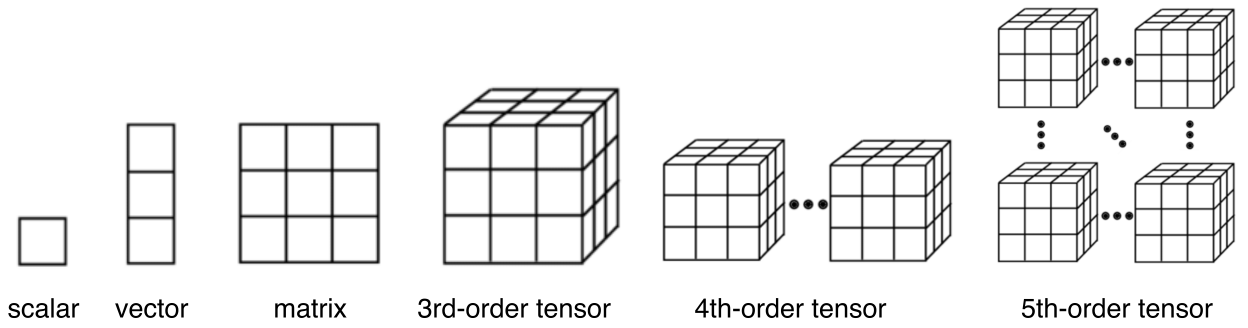}
    \caption{Toy examples of tensors. Each grid or cell denotes a scalar. Actually, it is impossible to directly visualize a 4th-order or higher-order tensor. Here we place different 3rd-order tensors in `different spaces' to ease the visualization.}
    \label{fig_tensors}
\end{figure}

Tensors in this paper are referred to as multidimensional arrays \citep{kolda2009tensor} that, as shown by Figure \ref{fig_tensors}, generalize scalars, vectors, and matrices to higher dimensions. We denote $\bm{\mathcal{X}}\in\mathbb{R}^{n_1\times n_2\times \cdots\times n_d}$ an order-$d$ or $d$-way real tensor. Higher-order ($d\geq 3$) tensors widely exist in science and engineering. Examples of 3rd-order tensors include RGB images and hyperspectral images. Examples of 4th-order tensors include a batch of RGB images and videos. Examples of 5th-order tensors include 3D MRI over time and subjects.

Tensor completion aims to recover the missing entries of a tensor satisfying some properties. 
Let $\Omega$ be the set of locations of observed entries of $\tilde{\bm{\mathcal{X}}}$.
Tensor completion usually solves the following problem
\begin{align}\label{problem_LRTC_0}
\mathop{\text{minimize}}_{\hat{\bm{\mathcal{X}}}}\dfrac{1}{2}\Vert \mathcal{P}_{\Omega}\left(\tilde{\bm{\mathcal{X}}}-\hat{\bm{\mathcal{X}}}\right)\Vert_F^2+\lambda\cdot R(\hat{\bm{\mathcal{X}}}),
\end{align}
where $R$ denotes a regularizer ensuring a certain property on $\hat{\bm{\mathcal{X}}}$. 
Similar to matrix completion, the most common property used in tensor completion is the low-rankness. Low-rank tensor completion (LRTC), as a higher-order extension of LRMC, has been extensively studied in the past decades. A naive approach to LRTC is to unfold a tensor to a matrix first and then perform LRMC or other missing data imputation methods \citep{liu2012tensor}. However, unfolding will destroy the higher-order structure of tensors and lead to lower imputation performance. \citet{yuan2016tensor} pointed out that the unfolding-based tensor completion algorithms require much more observed entries for exact recovery than direct tensor completion without matricization.

LRTC methods can be organized into different categories according to the types of factorization models. 
Representative tensor factorization models include CP (CANDECOMP/PARAFAC) decomposition \citep{carroll1970analysis}, Tucker decomposition \cite{tucker1966some}, t-SVD decomposition \citep{kilmer2011factorization}, tensor train decomposition \citep{oseledets2011tensor}, and tensor ring decomposition \citep{zhao2016tensor}. In this paper, we take the CP decomposition of a tensor $\bm{\mathcal{X}}\in\mathbb{R}^{n_1\times n_2\times \cdots\times n_d}$ as an example:
\begin{equation}
 \bm{\mathcal{X}} \approx \sum_{i=1}^r \mathbf{a}_i^{(1)} \circ \mathbf{a}_i^{(2)} \circ \cdots \circ \mathbf{a}_i^{(d)}
\end{equation}
where $\mathbf{a}_i^{(j)}\in\mathbb{R}^{n_j}$ is the 
$i$-th factor vector for mode $j$ and $\circ$ denotes the outer product. 
The CP rank of a tensor is defined as the minimum number of rank-one tensors that sum to $\bm{\mathcal{X}}$:
$$
\textup{rank}(\bm{\mathcal{X}})=\min\left\{r\in\mathbb{N}:\ \bm{\mathcal{X}}=\sum_{i=1}^r\mathbf{a}_i^{(1)}\circ\mathbf{a}_i^{(2)}\cdots\circ\mathbf{a}_i^{(d)}\right\}.
$$
It is known that the CP rank is NP-hard to compute.

CP decomposition based LRTC methods \citep{acar2011scalable,jain2014provable,zhao2015bayesian,NEURIPS2020_dab1263d,fan2023euclideannorminduced} often solve
 \begin{align}\label{problem_LRTC_cp}
\mathop{\text{minimize}}_{\{\mathbf{a}_{i}^{(j)}\}}~\dfrac{1}{2}\left\Vert \mathcal{P}_{\Omega}\left(\tilde{\bm{\mathcal{X}}}-\sum_{i=1}^r \mathbf{a}_i^{(1)} \circ \mathbf{a}_i^{(2)} \circ \cdots \circ \mathbf{a}_i^{(d)}\right)\right\Vert_F^2+\lambda\cdot R\left(\{\mathbf{a}_{i}^{(j)}\}\right)
\end{align}
where the recovered tensor is $\hat{\bm{\mathcal{X}}}=\sum_{i=1}^r \mathbf{a}_i^{(1)} \circ \mathbf{a}_i^{(2)} \circ \cdots \circ \mathbf{a}_i^{(d)}$.
The nuclear norm \citep{friedland2018nuclear} of a tensor $\bm{\mathcal{X}}$, though still NP-hard to compute, is defined as
\begin{equation}\label{eq_lrtc_cp1}
 \Vert \bm{\mathcal{X}}\Vert_\ast=\inf\left\lbrace\sum_{i=1}^r\vert s_i\vert:\ \bm{\mathcal{X}}=\sum_{i=1}^r s_i\mathbf{a}_i^{(1)}\circ\mathbf{a}_i^{(2)}\cdots\circ\mathbf{a}_i^{(d)},~\Vert\mathbf{a}_i^{(j)}\Vert=1, \ r\in\mathbb{N}\right\rbrace.   
\end{equation}
Based on the nuclear norm, a variant of \eqref{eq_lrtc_cp1} can be derived as
\begin{align}\label{problem_LRTC_cp_1}
\mathop{\text{minimize}}_{\{s_i,\mathbf{a}_{i}^{(j)}\}}~\dfrac{1}{2}\left\Vert \mathcal{P}_{\Omega}\left(\tilde{\bm{\mathcal{X}}}-\sum_{i=1}^r s_i\cdot \mathbf{a}_i^{(1)} \circ \mathbf{a}_i^{(2)} \circ \cdots \circ \mathbf{a}_i^{(d)}\right)\right\Vert_F^2+\lambda\cdot \|\mathbf{s}\|_1
\end{align}
where $\Vert\mathbf{a}_i^{(j)}\Vert=1$, $i=1,\ldots,r$, $j=1,\ldots,d$. Note that the $\ell_1$ norm in \eqref{problem_LRTC_cp_1} can be replaced by $\ell_p$ quasi-norms, leading to tensor Schatten-$p$ quasi-norm regularized LRTC \citep{fan2023euclideannorminduced}. In \citep{bazerque2013rank,pmlr-v51-cheng16,yang2016tensor,lacroix2018canonical,fan2023euclideannorminduced}, $s_i$ are absorbed into the factor vectors $\mathbf{a}_i^{(j)}$, which makes the optimization more efficient. In particular, \citep{fan2023euclideannorminduced} presented
a class of tensor rank regularizers based on the Euclidean norms of the CP component vectors of a tensor and showed that these regularizers are monotonic transformations of tensor Schatten-$p$ quasi-norms. This connection enables us to minimize the Schatten-$p$ quasi-norm in LRTC implicitly on big tensors and provides an arbitrarily sharper rank proxy for low-rank tensor recovery compared to the nuclear norm.

The Tucker decomposition of a tensor $\bm{\mathcal{X}}\in\mathbb{R}^{n_1\times n_2\times \cdots\times n_d}$ can be formulated as
\begin{equation}\label{tucker_decomp}
 \bm{\mathcal{X}} \approx \bm{\mathcal{G}} \times_1 \mathbf{U}^{(1)} \times_2 \mathbf{U}^{(2)} \times_3 \cdots \times_d \mathbf{U}^{(d)}
\end{equation}
where $\bm{\mathcal{G}} \in \mathbb{R}^{r_1 \times r_2 \times \cdots \times r_d}$ is the core tensor (a compressed version of $\bm{\mathcal{X}}$), $\mathbf{U}^{(j)} \in \mathbb{R}^{n_j \times r_j}$ is the factor matrix for mode $j$ (analogous to singular vectors in matrix SVD), and $\times_j$ denotes the mode-$j$ product (a tensor-matrix multiplication along mode $j$ ). See \citep{kolda2009tensor} for more details. The rank of $\bm{\mathcal{X}}$ based on Tucker decomposition is a tuple $(r_1,r_2,\ldots,r_d)$, which is the size of the minimum core tensor $\bm{\mathcal{G}}$ ensuring equality in \eqref{tucker_decomp}. Examples of Tucker decomposition-based LRTC methods are \citep{xu2013parallel,xu2013block,kressner2014low,kasai2016low,8000407,kong2018t}.

For convenience, Table \ref{tab_TC} summarizes tensor completion approaches based on different decomposition models. The table shows that only a few deep-learning-based methods exist. For instance, \citet{fan2021multi} proposed a multi-mode deep tensor factorization method for tensor completion and provided a generalization error analysis. Their idea is to perform further factorization on the Tucker factor matrices using neural networks. Separately, \citet{ahn2024neural} proposed NeAT, which applies neural networks to each latent component in an additive fashion. This approach not only captures diverse patterns and complex structures in sparse tensors but also offers direct and intuitive interpretation by remaining close to the multi-linear tensor model.

The aforementioned models are for regular tensors, where every mode has a fixed size and all slices are aligned. However, there exist irregular tensors in real applications. For example, in a patient-time-feature tensor, each patient may have a different number of visits. In recent years, a few researchers have proposed irregular tensor decomposition and completion methods \citep{ren2020robust,jang2022accurate,kwon2024compact,xie2025irtf,han2025irregular}.

\begin{table}[]
    \centering
    \begin{tabular}{c|c}
    \toprule
    Decomposition model & Tensor completion papers\\ \midrule
       CP  & \makecell{\citep{acar2011scalable,jain2014provable,zhao2015bayesian}\\\citep{cai2019nonconvex,NEURIPS2020_dab1263d,fan2023euclideannorminduced} }\\ \midrule
       Tucker  & \makecell{\citep{gandy2011tensor,balazevic-etal-2019-tucker,huang2015provable,8000407}\\\citep{chen2019non,pan2020low,tong2022scaling,wang2023implicit}}\\ \midrule
       tensor-train/ring & \makecell{\citep{bengua2017efficient,ding2019low,wang2017efficient,yuan2019tensor}\\ \citep{huang2020robust,qiu2022noisy,long2021bayesian}}\\ \midrule
       t-SVD & \makecell{\citep{zhang2016exact,lu2018exact,zhang2020low,jiang2020framelet}\\\citep{he2022tensor,wang2023guaranteed,wu2024smooth}}\\ \midrule
       deep learning & \makecell{\citep{liu2019costco,wu2019neural,ijcai2020p339,9338416}\\\citep{fan2021multi,ahn2024neural,10978100}}\\
       \bottomrule
    \end{tabular}
    \caption{Categorization of tensor completion methods}
    \label{tab_TC}
\end{table}

\subsection{Link Prediction and Graph Completion}\label{sec_lp_gc}

Given a graph $G=(V,E)$, link prediction is to use the observed nodes and topological structure to estimate whether an unobserved edge currently exists but is missing or will appear in the future. Specifically, let $U=\{(u, v) \mid u, v \in V, u \neq v\}$ be the universal set of possible links, link prediction aims to find a scoring function $s: V \times V \rightarrow \mathbb{R}$ that assigns a likelihood score to each node pair $(u,v)$, particularly those in $U \backslash E$. A higher score $s(u,v)$ indicates a higher probability that a link should exist or will form between $u$ and $v$.
Link prediction finds applications in recommender systems, knowledge graph completion, etc. There are several review papers on link prediction \citep{hasan2011survey,7046907,martinez2016survey,daud2020applications,zhou2021progresses,kumar2020link,10.1007/978-3-031-77954-1_1}. \citet{martinez2016survey} partitioned link prediction methods into the following categories: 1) similarity-based methods; 2) probabilistic and statistical methods; 3) algorithmic methods; 4) preprocessing methods. In recent years, graph neural networks have shown promising performance in link prediction tasks \citep{zhang2018link,cai2021line,zhu2021neural,wang2021surveylink,yun2021neo,li2023evaluating,shomer2024lpformer}, especially in the application of recommendation systems \citep{he2020lightgcn}. More recently, there have been a few attempts to extend large language models to link prediction \citep{shu2024knowledge,bi2024lpnl,he2024linkgpt}.

Graph completion, which is more general than link prediction, aims to enrich the entire graph by adding not only edges but also missing nodes, attributes, or labels so that the resulting graph is more complete and useful for downstream tasks. A general graph is formulated as $G=(V,E,\mathbf{X})$, where $\mathbf{X}$ denotes the matrix of node attributes. Graph completion may recover the missing edges as well as the missing values in $\mathbf{X}$, or predict where a new node should be inserted into the graph \citep{wu2023overview}. Within the tasks of graph completion, knowledge graph completion (KGC) is best known. KGC is the task of predicting missing facts in a knowledge graph, typically by inferring new relationships between entities to answer queries like (subject, predicate, ?) or (?, predicate, object).
There have been many review papers on KGC \citep{9220143,SHEN2022109597,zamini2022review,cai2022temporal,10.1007/s00521-023-09286-2,LI2025100809}. In \citep{LI2025100809}, KGC methods are organized into embedding-based methods, path-based methods, neural network-based methods, and LLM-based methods \citep{wang2022simkgc,wei2024kicgpt,xu2024multi,li2024contextualization,zhang2024making,yao2025exploring}.

\subsection{Time Series Imputation}
The task of time series imputation is to recover the missing values in a single multivariate time series (e.g., a matrix $\mathbf{X}\in\mathbb{R}^{d\times T}$) or a union of independent multivariate time series (e.g., a union of matrices $\mathbf{X}_i\in\mathbb{R}^{d\times T_i}$, $i=1,\ldots,K$). Compared to the standard missing data imputation, time series data imputation should consider the auto-correlation, trend/seasonality, and lagged cross-correlation. Time-series imputation keeps critical real-time systems (power grids, ICU monitors, high-frequency trading) from blinding when sensors drop out, preventing cascading failures or false alarms.
It also unlocks the full value of expensive historical records—satellite climate data, industrial IoT (internet of things) logs, retail POS streams—so forecasting, digital twins and compliance audits can run on complete, audit-ready data instead of discarding gap-ridden sequences. Recent advances in time series imputation are based on recurrent neural networks \citep{cao2018brits} and diffusion models \citep{tashiro2021csdi}.
There have been a few review papers on time series imputation \citep{pratama2016review,weerakody2021review,ribeiro2022missing,zainuddin2022time,fang2020time,wang2024deep,qian2025deep}. In the past ten years, deep learning methods, especially deep generative models, have shown impressive performance in time-series imputation.   \citet{fang2020time} wrote a survey paper of deep learning-based imputation methods for time series \citep{luo2018multivariate,gupta2020time,park2023long}.
\citet{kazijevs2023deep} focused on time series health data and provided a review with numerical comparisons for a few deep learning imputation methods. \citet{jin2024survey} reviewed graph neural network-based methods for time series imputation. \citet{du2024tsi} provides a standardized pipeline to evaluate imputation performance and offers a systematic way to adapt forecasting algorithms for imputation. It involved over 34,000 experiments, 28 algorithms, and 8 datasets.

\subsection{Online Imputation for Streaming Data}
Online data imputation refers to the process of estimating and filling in missing values in a continuous, streaming data environment where new data arrives sequentially over time, and the imputation model must update its estimates incrementally without reprocessing the entire dataset. Compared to batch or offline imputation, online imputation should adapt and learn from new incomplete data dynamically. That means online imputation enables real-time decision-making, monitoring, and analytics in dynamic environments. To be more specific, suppose we have one observation $\mathbf{x}_t\in\mathbb{R}^m$ at time point $t$, of which the missing entries are indicated by $\bm{\omega}_t\in\{0,1\}^m$. We need to construct a model $\mathcal{M}_t$ to impute the missing values of $\mathbf{x}_{t}$ using $\{\mathbf{x}_{1},\mathbf{x}_{2},\ldots,\mathbf{x}_{t-1}\}$ quickly, where $\mathcal{M}_t$ evolves with $t$ since the distribution of the complete data may change over time, usually slowly. Online missing data imputation finds critical applications in domains such as IoT and sensor networks \citep{deng2022online}, chemical process monitoring, financial trading \citep{nishanth2013computational}, and patient health monitoring.

A straightforward approach to adapting the aforementioned offline imputation methods for an online setting is to update the model incrementally, for instance via gradient descent, as new data arrives. Such an approach, however, requires careful design to be effective.
Over the past decades, several online imputation algorithms have been proposed. For instance, \cite{brand2003fast} developed a fast online SVD revision method for missing data imputation, which was applied to collaborative filtering, allowing users to asynchronously join, add ratings, add movies, revise ratings, get recommendations, and delete themselves from the model. \citet{MC_GROUSE_2010} provided an algorithm of online subspace tracking on incomplete data, which can be used for online missing data imputation. \cite{ravi2014new} developed an auto-associative neural network method for online missing data imputation.
\citet{guo2015online} developed an online LRMC method based
on low-rank factorization. \citet{devooght2015dynamic} developed an algorithm of dynamic matrix factorization for collaborative filtering.
\citet{Fan_2019_CVPR} proposed an online KFMC method to recover high rank matrices online. \citet{zhao2022online} provided a Gaussian Copula-based online imputation algorithm for data of mixed types, including ordinal, boolean, and continuous variables.
\citet{fan2022dynamic} presented a dynamic nonlinear matrix completion for online data imputation based on fast eigenvalue decompositions of the kernel matrices constructed by a sliding window of the data stream. \citet{LIU2023107822} proposed an online imputation algorithm that is able to handle binary, count and continuous variables simultaneously. \citet{zhan2025online} developed a mask asymmetric transformer GAN for online imputation in edge computing.

\subsection{Categorical Data Imputation}\label{sec_categorical}
Categorical data imputation \citep{van1992imputation,chen2003deal,cranmer2013we} is fundamentally more intricate than numerical data imputation due to the following reasons. First,
unlike numerical variables, categorical data lack intrinsic ordering or meaningful distance, so mean-based or regression-style imputations are invalid and even model-based solutions must first convert categories to dummies whose continuous outputs then have to be rounded while preserving mutual exclusivity, a step that introduces bias. 
Second, missingness mechanisms are harder to specify or test for categorical data, since the unobserved truth is discrete. 
Lastly, categorical imputation is harder to evaluate: a single misclassification counts as a full error, so we must rely on high-variance metrics, like hit-rate, 
F1-score, or Cohen's $\kappa$ coefficient, instead of the stable RMSE/MAE used for numerical data.

Classical methods for categorical data imputation include kNN with Gower or Hamming distance \citep{schwender2012imputing,faisal2022nearest}, CART (classification and regression trees \citep{breiman2017classification}, MICE \citep{MICE}, and graphical models \citep{geng2003bayesian}.
\citet{akande2017empirical} and \citet{memon2023comparison} conducted empirical comparisons of several classical methods for categorical data. 
\citet{nishanth2016probabilistic} proposed a probabilistic neural network for categorical data imputation. The method outperformed classical methods such as kNN. Note that by converting all categorical variables into numerical form using techniques like one-hot encoding, we can perform numerical imputation methods, especially those based on deep neural networks. It is also possible to use LLMs to impute categorical data, as discussed in Section \ref{sec_LLM}.

\subsection{Multimodal Data Imputation}
Multimodal data, characterized by the integration of diverse data types such as images, text, tabular measurements, and graphs, is ubiquitous in modern artificial intelligence applications. Examples include electronic health records (EHRs) combining clinical measurements (tabular), medical notes (text), and X-ray images (vision); social media profiles with user demographics (tabular), posts (text), and profile pictures (vision); and product catalogs with specifications (tabular), descriptions (text), and customer interaction graphs. Learning from multimodal data is more effective than from a single modality \citep{ngiam2011multimodal,baltruvsaitis2018multimodal}. However, missing data in multimodal settings presents a uniquely complex challenge. The missingness can occur not only within individual features but across entire modalities for certain data instances. For instance, one patient in an EHR may have complete lab tests but no associated medical image, while another may have the reverse. 

\begin{figure}
    \centering
    \includegraphics[width=0.5\linewidth]{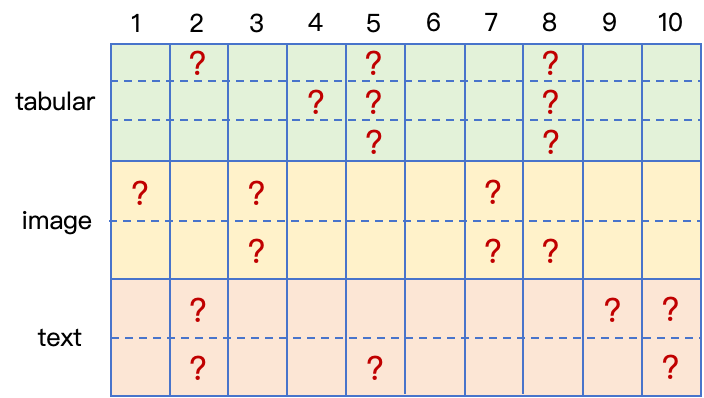}
    \caption{An intuitive example of multimodal data imputation (three modalities, ten subjects). Each cell represents a numerical (or categorical) value, an image, or text. The content in every cell marked by the red cross is missing.}
    \label{fig_mdi}
\end{figure}

Figure \ref{fig_mdi} presents a toy example of multimodal data imputation. As can be seen, multimodal data imputation is significantly more intricate than unimodal imputation. The former may require an integration of multiple machine learning models or techniques, e.g., a combination of a convolutional neural network and a recurrent neural network for handling image and text simultaneously. \citet{zhan2025systematic} provided a literature review on incomplete multimodal learning. It categorizes the related methods into two groups: internal information-based methods and external information-based methods. Since our paper focuses on missing data imputation, we here detail the internal information-based methods, particularly how to impute the multimodal data, without using any external information.

\citet{tran2017missing} presented a multimodal imputation method based on a cascaded residual autoencoder.
\citet{shang2017vigan} proposed combining GANs and AEs for multimodal imputation, where the AEs for different modalities share a common representation space to exploit cross-modal relationships.
\citet{chen2023disentangle} proposed a method that disentangles brain images into inter-modality relevant and intra-modality specific features, and uses knowledge distillation to impute missing modality representations, enabling multimodal Alzheimer's disease diagnosis using only single-modality data at inference.
\citet{cohen2023joint} developed a machine learning model based on joint VAEs for single-cell multimodal data imputation and embedding. Similarly, \citet{tang2024modal} combined contrastive learning within modality-specific AEs for single-cell data imputation.
More work about multi-omics data imputation can be found in \citep{du2022robust,chandrashekar2023deepgami,wang2024inclust+,jiang2024xmint,sadia2025depmicrodiff}.
\citet{poudel2025multimodal} proposed an imputation method for multimodal federated learning and evaluated it on three medical datasets containing text and image modalities.
\citet{hassanzadeh2025generative} provided a cycle-GAN imputation method for Alzheimer’s disease diagnosis based on two modalities. \citet{zhang2025hyper} applied diffusion models to federated multimodal graph completion, utilizing image, text, and graph information simultaneously. There is additional work on handling multimodal data, primarily focused on heuristic-based applications such as medical diagnosis, which we will not elaborate on in this paper.

The principle followed by most multimodal data imputation methods is that different modalities are correlated or share some common information; otherwise, it is impossible to recover one modality from others. Specifically, let $\mathbf{x}^{(j)}\in\mathcal{X}_j$ be the data of a subject or sample in modality $j$; $\mathbf{x}^{(j)}$ could be a vector, an image, a text, a graph, or a time series, etc, where $j=1,2,\ldots,M$ and $M$ is the number of modalities. Let $\mathbf{x}=(\mathbf{x}^{(1)},\mathbf{x}^{(2)},\ldots,\mathbf{x}^{(M)})$ and let $\mathcal{D}=\{\mathbf{x}_1,\mathbf{x}_2,\ldots,\mathbf{x}_N\}$ be the whole dataset of the multimodal data consisting of $N$ samples. Suppose $\mathcal{D}$ is incomplete, where an $\mathbf{x}_i^{(j)}$ may have some missing elements or is completely missing. Most deep learning-based missing data imputation methods we introduced in Section \ref{sec_deep_impute} can be extended to impute $\mathcal{D}$. For instance, here we consider an autoencoder-type imputation strategy. Let $f:\mathcal{X}_1\times\mathcal{X}_2\times\cdots\mathcal{X}_M\rightarrow\mathbb{R}^d$ be an encoder with $M$ different types of input and $g:\mathbb{R}^d\rightarrow\mathcal{X}_1\times\mathcal{X}_2\times\cdots\mathcal{X}_M$ be a decoder. Then we have the following framework for multimodal data imputation:
\begin{equation}
    \mathop{\text{minimize}}_{f,g}~\frac{1}{N}\sum_{i=1}^N\ell(\mathbf{x}_i,g(f(\mathbf{x}_i))|\bm{\omega}_i)
\end{equation}
where $\ell(\mathbf{x},\hat{\mathbf{x}}|\bm{\omega})$ denotes a function to compare $\mathbf{x}$ with its reconstruction $\hat{\mathbf{x}}$ only at the observed locations indicated by $\bm{\omega}$. It should be pointed out that $\ell$ is not a simple loss function such as the square loss; it should be able to compare two images, two texts, or even two graphs; $\ell$ itself could be a multimodal neural network, pretrained, or jointly optimized with $f$ and $g$. This idea can be adapted to other approaches, such as GAN and diffusion models, to impute multimodal data. 

\section{Supervised and Unsupervised Learning on Incomplete Data}
Missing data imputation often has a significant impact on downstream tasks such as classification and clustering, especially when the missing rates are high \citep{ghahramani1995learning,saar2007handling,EM}. Performing downstream task-agnostic imputation and then conducting the downstream task may not provide satisfactory performance.
The conventional approach of employing task-agnostic imputation strategies followed by separate downstream analysis frequently yields suboptimal performance outcomes. To address the limitation of this two-stage approach, one may need to ensure that the assumption on the data distribution or structure is the same as that in the downstream task. Moreover, a few researchers proposed end-to-end approaches that integrate missing data imputation with downstream tasks.

\subsection{Classification and Regression on Incomplete data}
The best-known strategy for classification and regression on incomplete data is the EM algorithm \citep{dempster1977maximum}, which was introduced in Section \ref{sec_likelihood}. There have been more methods for classification and regression on incomplete data. \citet{EM} reviewed different ways to handle missing data in classification tasks, grouping methods into categories like deleting incomplete cases, imputing (estimating) missing values, and using models that can work with incomplete data directly. The following discussion organizes the methods of classification and regression on incomplete data by model type, progressing from linear to non-linear.

\paragraph{Linear Models on Incomplete Data}
\citet{williams2005incomplete} presented a logistic regression-based classification method for datasets with missing features, avoiding imputation by analytically integrating over the missing values using a Gaussian Mixture Model (GMM). \citet{NIPS2010_3932} concatenated the feature matrix and label matrix, treated the unknown labels as a part of missing values, and used low-rank matrix completion to predict the unknown labels directly, where the classifier is linear. 
\citet{hazan2015classification} proposed an efficient, theoretically grounded algorithm that learns a linear classifier directly from incomplete data without requiring matrix completion. The method uses a kernel-based approach to embed observed data into a high-dimensional space, enabling competitive classification performance with the best possible linear classifier trained on full data, even when most features are missing.

\paragraph{kNN on Incomplete Data}
\citet{garcia2009k} proposed a kNN method with mutual information for simultaneous classification and missing data imputation. \citet{sell2024nonparametric} introduced a nonparametric framework for classification problems in the presence of missing data and derived the minimax rate for the excess risk of classification. It also proposed a Hard-thresholding Anova Missing data (HAM) classifier based on kNN and a thresholding step. 

\paragraph{Tree Methods on Incomplete Data}
\citet{TWALA2008950} proposed a method called MIA for handling missing data in decision trees by treating missingness as a separate, informative group during the splitting process. The authors demonstrated that MIA performs competitively with more complex methods like the EM-based method, while offering advantages in simplicity and computational efficiency. \citet{hapfelmeier2014new} introduced a new variable importance measure that is applicable to data with or without missing values, thereby enabling the use of random forests on incomplete datasets.

\paragraph{Kernel Machine on Incomplete Data}
\citet{smola2005kernel} introduced a framework for handling missing data in kernel methods like Support Vector Machines (SVM) \citep{boser1992training} and Gaussian Processes \citep{williams1995gaussian} by modeling them as estimation problems within exponential families. The key idea is to treat estimation with missing variables as a problem of computing marginal distributions, which leads to non-convex optimization problems. \citet{chechik2006max} developed a max-margin classifier that works directly with incomplete data, avoiding the need for a separate imputation step. Their method uses a geometry-based approach with different optimization strategies for separable and non-separable cases. 
\citet{fan2020polynomial} proposed to use polynomial matrix completion to do nonlinear classification on incomplete data, showing improved performance compared to the method of \citep{NIPS2010_3932}, SVM, and LRMC+SVM.

\paragraph{Deep Learning on Incomplete Data}
Although deep learning-based imputation methods (Section \ref{sec_deep_impute}) are often adaptable to joint imputation and classification, some research focuses specifically on classification without imputation or prioritizes the supervised learning objective. For instance, \citet{ghorbani2018embedding} proposed an embedding approach to learn representations for missingness in parallel with the network's prediction task. The approach bypasses the need to impute the missing attributes.
For incomplete image classification, \citet{danel2020processing} proposed to convert images with missing values to graphs, and then use GCN to do graph classification, which avoids imputation.
\citet{chang2022neural} presented GapNet, a neural network method that can be applied to highly incomplete medical datasets.
\citet{kim2023probabilistic} developed a probabilistic framework for classifying multivariate time series data with missing values by combining a deep generative model for imputation with a classifier. To prevent the model from taking the easy way out and ignoring the imputed values, the authors introduce a novel regularization technique called `obsdropout', which randomly drops observed values during training to force the classifier to rely more on the generated imputations.



\subsection{Clustering on Incomplete Data}

Clustering is an unsupervised learning task. Therefore, handling missing data in clustering is more challenging than in classification and regression. In recent decades, many clustering algorithms have been proposed. Typical clustering algorithms include k-means \citep{ball1965isodata,IKOTUN2023178}, DBSCAN \citep{ester1996density}, Gaussian mixture models, spectral clustering \citep{shi2000normalized,ng2001spectral}, subspace clustering \citep{parsons2004subspace,elhamifar2013sparse}, and deep learning-based clustering \citep{xie2016unsupervised,ji2017deep,cai2022efficient,ren2024deep}. When the data are incomplete, we can use some missing data imputation methods to fill the missing values and then perform these clustering algorithms. However, such a two-stage approach may not work well if the imputation algorithm does not align with the cluster structure in the data. There have been a few efforts in designing more effective approaches for clustering on incomplete data. 

\paragraph{K-means Clustering on Incomplete Data}

\citet{sarkar2001fuzzy} proposed an algorithm for fuzzy k-means on incomplete data, where the missing values are repaired incrementally in each iteration.
\citet{wagstaff2004clustering} presented a method for encoding partially observed features as a set of supplemental soft constraints and introduced an algorithm of k-means with soft constraints, which incorporates constraints into the clustering process, thereby eliminating the need for imputation.
\citet{chi2016k} developed an algorithm called k-POD for k-means with missing data. k-POD minimizes the sum of the squared differences between the data and the centers over the observed entries only, remaining unhindered by the need for accurate imputations.
\citet{8693952} proposed a variant of k-means that unifies clustering and imputation into one single objective, where missing values are optimized dynamically. The method outperformed k-means with kNN, EM, and mean imputations on a few datasets. \citet{armah2024k} extended the idea to k-means with the Mahalanobis distance. There are also studies on kernel k-means for incomplete data such as \citep{liu2020efficient} and Gaussian mixture models with missing data \citep{zhang2021gaussian}.

\paragraph{Spectral Clustering on Incomplete Data}
Spectral clustering first constructs the similarity matrix and then performs k-means on the representations given by the eigenvalue decomposition of the Laplacian matrix \citep{ng2001spectral}. A simple strategy for handling missing data is computing the similarity between two data points using only the observed values, especially when the Gaussian kernel is used. However, this may not work well when the missing rate is high. 
\citet{lokse2017spectral} developed a robust kernel function that is specifically designed to handle incomplete data and applied it to spectral clustering.
\citet{10586869} presented a spectral embedding fusion method for incomplete
multi-view clustering.
\citet{wen2022survey} wrote a survey paper of incomplete multi-view clustering, which included a few extension methods of spectral clustering for incomplete data.
\citet{NEURIPS2023_e5aa7171} proposed
an imputation-free framework with two approaches to improving spectral
clustering on incomplete data. The first approach is based on a kernel correction method that finds the best positive semi-definite approximation for the kernel matrix estimated on incomplete data. The second one is an extension of the first one, by using kernel self-expressive learning.

\paragraph{Subspace Clustering on Incomplete Data} Most subspace clustering methods can be organized into two categories. The first category is based on learning the subspaces directly \citep{bradley2000k,fan2021large}, while the second category is based on self-expressive models \citep{elhamifar2013sparse} followed by spectral clustering. 
\cite{yang2015sparse} developed two methods for subspace clustering on incomplete data. The first one generalizes the sparse subspace
clustering algorithm so that it can obtain a sparse representation of the data using only the observed entries. The second one estimates a suitable kernel matrix by assuming a random model for the
missing entries and obtains the sparse representation from this kernel.
\citet{pimentel2016group} also proposed two methods for subspace clustering on incomplete data: (a) group-sparse subspace clustering (GSSC), which is based on group-sparsity and alternating minimization, and (b) mixture subspace clustering (MSC), which models each data point as a convex combination of its projections onto all subspaces in the union. \citet{pimentel2016information} analyzed the information-theoretic requirements of subspace clustering with missing data. \citet{fan2017sparse} proposed to alternately compute the matrix of sparse representation coefficients and recover the missing entries of a data matrix. In particular, the algorithm recovers missing entries by minimizing the representation coefficients, representation errors, and matrix rank.
\citet{tsakiris2018theoretical} provided theoretical
guarantees for sparse subspace clustering \citep{elhamifar2013sparse} with incomplete data, showing that projecting the
zero-filled data onto the observation pattern of the point being expressed can lead to substantial improvement in performance.
\citet{lane2019classifying} provided a comprehensive comparison of methods for subspace clustering with missing data. 
\citet{wang2020icmsc} proposed a method for incomplete cross-modal subspace clustering.
\citet{liu2021self} developed an algorithm for self-representation subspace clustering on incomplete multi-view data. \citet{soni2025integer} proposed to dynamically determine a set of candidate subspaces and optimally assign points to selected subspaces. The empirical study showed that the proposed approach can achieve high clustering accuracy even when the data are high rank, the percentage of missing data is high, or the subspaces are similar.

\paragraph{Deep Clustering on Incomplete Data}
\citet{de2019deep} developed a variational deep embedding model for clustering of multivariate clinical patient trajectories with missing values.
\citet{wen2020dimc} proposed a deep learning model for incomplete multi-view clustering. \citet{xu2024deep} provided a deep variational incomplete multi-view clustering algorithm that is able to explore shared clustering structures of incomplete data from different views. More work on deep clustering with missing data can be found in \citep{ZHAO20181053,wen2021structural,xu2022deep,lin2022incomplete,tang2022deep}.

\subsection{Anomaly/Outlier/Novelty Detection on Incomplete Data}

Anomaly detection \citep{chandola2009anomaly}, sometimes known as outlier detection \citep{boukerche2020outlier}, novelty detection \citep{pimentel2014review}, or even fault detection (in the manufacturing industry) \citep{venkatasubramanian2003review}, is usually an unsupervised learning task that aims to identify the samples deviating from common patterns.  Typical algorithms include kernel density estimation, one-class SVM \citep{scholkopf1999support}, Isolation Forest \citep{liu2008isolation}, Deep SVDD \citep{ruff2018deep}, etc. Note that most of the dimensionality reduction methods \citep{van2009dimensionality}, such as PCA \citep{abdi2010principal}, KPCA \citep{scholkopf1998nonlinear}, and autoencoder \citep{hinton2006reducing}, can be used for anomaly detection or outlier detection. More papers about anomaly detection, especially those based on deep learning, can be found in \citep{pang2021deep}.

\citet{dietterich2018anomaly} evaluated the detection performance of a few anomaly detection methods combined with different data imputation techniques. The experimental results showed that implementations contribute positively to the detection performance of unsupervised anomaly detection methods on incomplete data. 
\citet{fantii2022} investigated statistical process monitoring in the presence of missing data and introduced a fast incremental nonlinear matrix completion technique for online and sequential imputation. \citet{xu2021conformal} developed ECAD, a method tailored for incomplete spatio-temporal data like traffic flow sequences, while \citet{ma2021probabilistic} utilized Probabilistic Principal Component Analysis (PPCA) to recover missing values and identify anomalies. \citet{sarda2023unsupervised} conducted a comparative analysis of existing unsupervised anomaly detection techniques applied to GAN-imputed datasets. These approaches follow a two-stage framework, in which the imputation models are trained exclusively on normal data or datasets with very few unlabeled outliers. Consequently, when applied to abnormal data during inference, the imputation model tends to reconstruct missing values based on learned normal patterns, thereby making anomalies appear more normal and reducing detection accuracy.

\citet{NEURIPS2024_f99f7b22} proposed an end-to-end method called ImAD for unsupervised anomaly detection in the presence of missing values. It integrates data imputation with anomaly detection into a unified optimization objective and automatically generates pseudo-abnormal samples to alleviate the imputation bias. They also proved the effectiveness of ImAD theoretically. \citet{zhang2025learning} proposed an anomaly detection method for incomplete time series that flags unusual patterns by checking how consistently a two-step, patch-based imputation network fills the same series under different random masks.

\subsection{Other Tasks on Incomplete Data}
Beyond the tasks discussed above, numerous other applications require careful handling of missing data. For brevity, we summarize key references in Table \ref{tab_other_tasks} without further elaboration.
\begin{table}[h!]
    \centering
    \begin{tabular}{c|c} \toprule
    Task & Papers \\ \midrule
    dimensionality reduction  & \makecell{\citep{sanguinetti2006missing,nguyen2008robust,hu2010semiparametric}\\ \citep{de2018nonlinear,gilbert2018unsupervised,ling2021dimension,yan2024inference}}  \\ \midrule
       change point detection  &  \makecell{\citep{xie2012change,londschien2021change,zhao2022online,follain2022high}\\ \citep{liu2025high,xu2025online,enikeeva2025change}}\\ \midrule
       causal inference \& discovery  & \makecell{\citep{kallus2018causal,yang2019causal,pmlr-v89-tu19a,nguyen2020bayesian}\\ \citep{wang2020causal,huang2020causal,hillis2021causal}\\ \citep{gao2022missdag,levis2025robust,landsiedel2025causal}}\\ 
              \midrule
       reinforcement learning & \makecell{\citep{yamaguchi2020model,mei2023reinforcement,chasalow2025missing}} \\ \bottomrule
    \end{tabular}
    \caption{Other tasks on incomplete data}
    \label{tab_other_tasks}
\end{table}

\section{Theoretical Guarantees for Missing Data Imputation}\label{sec_theory}
Most research on missing data imputation is empirical, as providing theoretical guarantees for complex models is highly challenging. Specifically, it is difficult to ensure the accurate recovery of missing values or the convergence to optimal parameters. Existing theoretical analyses can be broadly categorized into three types: guarantees for effective parameter estimation, exact recovery of missing values, and generalization error bounds.

\subsection{Effective Parameter Estimation}
\citet{dempster1977maximum} proved that the EM algorithm for handling missing data converges to a local maximum point of the likelihood function. Although this does not directly guarantee accurate missing data imputation, it guarantees the effectiveness of the EM algorithm in estimating the parameters of many models such as mixture models and robust regression. 
\citet{chandrasekher2020imputation} studied the problem of single imputation for high-dimensional sparse linear regression and obtained optimal consistency rates of both the LASSO \citep{ranstam2018lasso} and the square-root LASSO without modification.
\citet{bertsimas2024simple} studied the performance of impute-then-regress pipelines by contrasting theoretical and empirical evidence. It established the asymptotic consistency of such pipelines for a broad family of imputation methods and showed that mean-impute is asymptotically optimal.
\citet{verchand2024high} presented an asymptotically exact characterization of the risk of ridge-regularized logistic regression when the observed covariates stem from a Gaussian error-in-variables model.

\subsection{Exact Recovery Guarantee}
\cite{CandesRecht2009} proved that a rank-$r$ matrix of size $n\times n$ can be recovered exactly via nuclear norm minimization when the number of observed entries (sampled uniformly at random) is larger than $Cn^{1.2}r\log n$ for some positive numerical constant $C$. The bound was further extended or improved by a few studies such as \citep{candes2010matrix,MC_icml2014c1_chenc14}. 
Regarding the low-rank factorization model for matrix completion, nonconvex but more efficient than nuclear norm minimization, a few scholars proposed alternating minimization and gradient descent algorithms that can exactly recover low-rank matrices \citep{keshavan2012efficient,jain2013low,hardt2014understanding,ruoyu15}.
Compared to low-rank matrix completion, exact recovery guarantees for low-rank tensor completion is more difficult to obtain. Nevertheless, there have been a few guaranteed algorithms for tensor completion under CP decomposition \citep{krishnamurthy2013low,mu2014square,jain2014provable,pmlr-v65-potechin17a,NEURIPS2020_dab1263d}, Tucker decomposition \citep{huang2014provable}, and t-SVD \citep{zhang2016exact}.

\subsection{Generalization Error Bound}
Most of the missing data imputation methods train a model using the observed values and then use the model to predict the missing values. Therefore, it is important to understand the gap between the training error (for observed values) and the expected error (for missing values). The gap is usually related to the number of observed values and the property of the imputation model. A small gap means that the model has a strong generalization ability. Generalization analysis is a typical topic in the area of learning theory. 

\citet{srebro2005rank} analyzed the generalization error bounds of rank, nuclear norm, and max-norm constrained matrix completion methods under a uniform sampling distribution. \citet{shamir14a} provided a generalization bound for matrix completion with nuclear norm minimization under a milder assumption. \citep{negahban2012restricted,fan2019factor} analyzed the recovery error bound for Schatten-$p$ quasi-norm based matrix completion.
\citet{fan2021multi} provided generalization bounds for deep matrix and tensor factorization models on incomplete data. \citet{fan2023euclideannorminduced} analyzed the generalization bounds of Schatten-$p$ norm minimization for LRTC and showed that for LRTC with Schatten-$p$ quasi-norm regularizer on $d$-order tensors, $p = 1/d$ is always better than any $p > 1/d$ in terms of the generalization ability. \citet{wu2024smooth} analyzed the generalization bound of total-variation regularized tensor completion. \citet{ledent2024generalization} provided a tighter bound for matrix completion with Schatten-$p$ quasi-norm constraint.

\section{Benchmarks, Toolboxes, and Evaluation Metrics}\label{sec_benchmarks}
\subsection{Benchmarks and Toolboxes}\label{sec_bench}
Despite the extensive literature on missing data imputation, comprehensive toolboxes and benchmarks remain relatively limited. Several studies have sought to fill this gap. For instance, \cite{folch2016missing} provided a MATLAB toolbox for missing data imputation and the imputation methods are mainly based on PCA and linear regression.
\cite{josse2016missmda} developed an R package for handling missing data. It includes principal component analysis for continuous variables, multiple correspondence analysis for categorical variables, factorial analysis on mixed data for both continuous and categorical variables, and multiple factor analysis for multi-table data.
\citet{fancyimpute} developed fancyimpute, a Python toolbox of missing data imputation, that included simple imputation, kNN, and a few LRMC methods. 
\citet{woznica2020does} evaluated the empirical effectiveness of 7 imputation algorithms for 5 predictive models (logistic regression, classification tree, random forest, etc.) on 13 datasets. 
\citet{jager2021benchmark} compared 6 imputation methods on 69 datasets under MCAR, MAR, and MNAR missingness patterns (with different missing rates) and evaluated by regression and classification downstream tasks.  \citet{miao2022experimental} experimentally compared 19 imputation algorithms on 15 real-world benchmark datasets. \citet{perez2022benchmarking} compared 6 imputation methods in 10 classification and 3 regression tasks on 4 publicly available health databases. \citet{pereira2024imputation} tested 6 imputation methods in an experimental setup that covers 10 datasets and 5 missingness levels (10\% to 80\%) in different MNAR settings. Empirically, they found that, for most missing rates and datasets, multiple imputation by chained equations performed best, whereas autoencoders showed promising results at higher missingness rates. \citet{cabrera2024benchmark}
provided a comparison of classical and deep learning imputation methods on 21 heterogeneous datasets in various areas.
\citet{pons2024iti} developed a toolbox to assess the reliability of
various imputation methods, select the best imputer for any feature or group of features, and filter out features that do not meet quality criteria.
\citet{toye2025benchmarking} evaluated 12 imputation methods on two time-series datasets only. \citet{richter2025imputebench} proposed ImputeBenc, which includes 4 imputation methods only, though more methods can be inserted into the workflow. There are more benchmarking papers on missing data imputation, most of which focus on specific data types or problems. For convenience, we summarize those published after 2020 in Table \ref{tab_bench}. 

In practice, a common and rigorous approach to benchmarking imputation methods involves the amputation of complete datasets. In this procedure, missing values are generated under controlled conditions, dictating the mechanism (MCAR, MAR, MNAR), the proportion of missing data, and the dependency structures among features. This simulation framework allows for objective performance comparison against the known ground truth.

The simulation-based benchmarking paradigm is supported by a relatively mature ecosystem of tools and studies. For instance, specialized libraries (e.g., pyampute \citep{Schouten2018,schouten_rianne_m_2022_6946887}, missMethods \citep{missmethods2022}) facilitate controlled data amputation and method implementation. Reviews and frameworks (e.g., MissMecha \citep{zhou2025missmecha}) provide structured guidance for generating and studying missing data under different mechanisms. Additionally, auxiliary tools for visualizing missingness patterns (e.g., missingno \citep{bilogur2018missingno}) and testing assumptions (e.g., MissMech \citep{jamshidian2014missmech} for MCAR testing) further strengthen the experimental pipeline.

\begin{table}[h!]
    \centering
    \begin{tabular}{c|c|c|c}\toprule
    Reference & Field& No. of methods & No. of datasets\\ \midrule
    \citep{woznica2020does} & general & 7 & 13 \\
      \citep{jager2021benchmark} & general & 6 & 69 \\
      \citep{ahn2022comparison} & time series & 6 & 4\\
      \citep{miao2022experimental} &general  &19 & 15 \\
      \citep{perez2022benchmarking}  &healthcare   & 6& 4\\
      \citep{sun2023deep} & general & 6 & 10 \\
      \cite{kazijevs2023deep} & time series & 7 & 5 \\
      \citep{pereira2024imputation} & general & 6 & 10 \\
      \citep{cabrera2024benchmark} & general & 8 & 21 \\
      \citep{gama2024imputation} & geological & 8 & 3 \\
      \citep{prakash2024benchmarking} & health survey &4 & 10 \\
      \citep{gendre2024benchmarking} & biological & 6 & 2\\
      \citep{du2024tsi} & time series & 28 & 8 \\
      \citep{toye2025benchmarking} & time series & 12 & 2\\
      \citep{zhang2025comparison} & healthcare & 5& 3\\
      \citep{richter2025imputebench} & general & 4 &-- \\ \bottomrule
    \end{tabular}
    \caption{Literature on missing data imputation benchmarks (2020-present)}
    \label{tab_bench}
\end{table}

\subsection{Evaluation Metrics}
The evaluation metrics for missing data imputation can be organized into four groups, shown in Table \ref{tab_metrics}. The first group is composed of various imputation errors, based on the difference between the imputed value and the true value. Let $x_{i,j}$ and $\hat{x}_{i,j}$ be the true value and imputed value respectively, where $(i,j)\in\Omega$ and $\Omega$ denotes the index set of all missing values. The mean absolute error is defined as $\text{MAE}:=\frac{1}{|\Omega|}\sum_{(i,j)\in\Omega}|x_{i,j}-\hat{x}_{i,j}|$. Similarly, the root mean square error is $\text{RMSE}:=\sqrt{\frac{1}{|\Omega|}\sum_{(i,j)\in\Omega}(x_{i,j}-\hat{x}_{i,j})^2}$. MAE is less sensitive to outliers than RMSE. It is worth noting that both MAE and RMSE cannot be compared across different datasets unless the data in each dataset are standardized. For instance, when the data are standardized into $[0,1]$, we can compare the imputation difficulty between two datasets. One common limitation of MAE and RMSE is that they cannot indicate how large the imputation error is relative to the ground truth; although standardization to $[0,1]$ helps, it is still sensitive to outliers. Therefore, we should use the relative MAE and relative RMSE, defined as $r\text{MAE}:=\sum_{(i,j)\in\Omega}|x_{i,j}-\hat{x}_{i,j}|/\sum_{(i,j)\in\Omega}|x_{i,j}|$ and $r\text{RMSE}:=\sqrt{\sum_{(i,j)\in\Omega}(x_{i,j}-\hat{x}_{i,j})^2/\sum_{(i,j)\in\Omega}x_{i,j}^2}$, respectively. A practical consideration for evaluating imputation methods on real-world tabular data is the handling of heterogeneous feature types. While metrics like RMSE and MAE are standard for continuous variables, categorical, ordinal, and binary variables necessitate different evaluation criteria (e.g., accuracy, F1-score, or weighted Cohen's kappa). Therefore, a complete benchmarking framework for such data should employ a hybrid evaluation strategy that combines type-specific metrics or their aggregated scores to reflect overall imputation quality across all feature types.

The second group is based on distance metrics between distributions such as the Wasserstein distance (WD) \citep{villani2008optimal} and maximum mean discrepancy (MMD) \citep{gretton2012kernel}. Let $\mathbf{X}\in\mathbb{R}^{d\times n_x}$ and $\mathbf{Y}\in\mathbb{R}^{d\times n_y}$ be two discrete distributions in $\mathbb{R}^{d}$, WD$_2$ (Wasserstein-2 distance) is calculated by
\begin{equation}\label{eq_WD2}
    \text{WD}_2^2(\mathbf{X}, \mathbf{Y}):=\min _{\mathbf{P} \in \Pi(\mathbf{a}, \mathbf{b})} \sum_{i=1}^{n_x} \sum_{j=1}^{n_y} P_{i j}\left\|x_i-y_j\right\|^2
\end{equation}
where $\mathbf{a}\in\mathbb{R}^{n_x}$ and $\mathbf{b}\in\mathbb{R}^{n_y}$ are the probability weights vectors often set as $a_i=1/n_x$ and $b_j=1/n_y$. MMD is defined by
\begin{equation}
    \text{MMD}^{2}(\mathbf{X}, \mathbf{Y}):=\frac{1}{n_x^2} \sum_{i=1}^{n_x} \sum_{j=1}^{n_x} k\left(\mathbf{x}_i, \mathbf{x}_j\right)+\frac{1}{n_y^2} \sum_{i=1}^{n_y} \sum_{j=1}^{n_y} k\left(\mathbf{y}_i, \mathbf{y}_j\right)-\frac{2}{n_x n_y} \sum_{i=1}^{n_x} \sum_{j=1}^{n_y} k\left(\mathbf{x}_i, \mathbf{y}_j\right)
\end{equation}
As can be seen, the computations of WD and MMD are much more time-consuming than the imputation errors, but they can be applied to cases where the ground truth of the missing values is not available. For example, let $\tilde{\mathbf{X}}_a$ be an incomplete data matrix, for which the ground truth of the missing values is not available. Let $\hat{\mathbf{X}}_a$ be the imputed data matrix given by some imputation method. If there is another complete data matrix ${\mathbf{X}}_b$ from the same distribution as $\mathbf{X}_a$, we can measure the WD between $\hat{\mathbf{X}}_a$ and $\mathbf{X}_b$, where a smaller WD means a better imputation performance. WD has been utilized in a few papers such as \citep{muzellec2020missing,jarrett2022hyperimpute}.

The third group of evaluation metrics assesses performance on downstream tasks—specifically regression, classification, and clustering.
For regression, the most common measures are the Mean Absolute Error MAE and RMSE.
For classification, standard metrics include accuracy, precision, recall, F1-score, AUROC, and AUPRC; \citet{grandini2020metrics} provided a comprehensive survey of these measures. Clustering metrics fall into two families: internal indices, which do not require ground-truth labels (e.g., the Davies–Bouldin index and Silhouette coefficient), and external indices, which do (e.g., clustering accuracy, normalized mutual information (NMI), and the adjusted rand index (ARI)).

It is also important to compare the theoretical space and time complexity, as well as the empirical running time of different imputation methods, especially when dealing with large-scale datasets that involve a high number of features or samples. Additionally, the number of hyperparameters should be taken into account, as hyperparameter tuning can be particularly challenging and time-consuming in real-world applications. 

\begin{table}[]
    \centering
    \begin{tabular}{c|c}\toprule
    Group&Metric\\ \midrule
    \multirow{4}{*}{imputation error}     & mean absolute error (MAE)   \\
    & root mean absolute error (RMSE)  \\
    & relative MAE \\
    & relative RMSE \\ \midrule
    \multirow{2}{*}{distribution distance}     & Wasserstein distance (WD)  \\ 
    & maximum mean discrepancy (MMD)  \\ \midrule
    \multirow{4}{*}{downstream performance} &  regression metrics (e.g., MAE, RMSE)   \\ 
    & classification metrics (e.g., precision, recall, AUROC)\\
    & clustering metrics (e.g., NMI, ARI) \\
    & others\\ \midrule
    \multirow{4}{*}{complexity} & space complexity\\
    & time complexity\\
    & running time\\
    & number of hyperparameters\\
    \bottomrule
    \end{tabular}
    \caption{Evaluation metrics in handling missing data}
    \label{tab_metrics}
\end{table}

\section{Challenges and Future Directions}\label{sec_challenges}

\subsection{Model Selection and Hyperparameter Tuning for Missing Data Imputation}
Hundreds of missing data imputation methods have been proposed and no single method performs the best in all cases. The reason is that different imputation methods usually rely on different assumptions about the data. For instance, low-rank matrix completion assumes low-rankness in the data matrix and will not perform well on high-rank matrices; autoencoder-based imputation methods assume that the data are lying on a low-dimensional manifold and will not work well when the data structure is more complex; kNN imputation, though nonparametric, is not effective on high-dimensional data and high-missing-rate cases; missForest, also nonparametric, is time-consuming for high-dimensional data and cannot provide extrapolated values. Therefore, it is necessary and important to select the most appropriate method or model for each dataset. 

Method or model selection for missing data imputation is challenging. Almost every imputation method has at least one hyperparameter and the best values for the hyperparameters vary across different datasets. For each dataset, we need to split it into two subsets, one for training and the other for validation, to find the best hyperparameters. This will be time-consuming when the number of hyperparameters is large and the imputation method has a high-time complexity. For instance, for a deep learning-based imputation method, we need to determine the number of layers of the neural network, the width of each layer, the learning rate, the number of iterations, the coefficient of weight decay, and algorithm-specific hyperparameters; if each has 4 candidate values, we need to use grid search over at least 4096 combinations, meaning training 4096 neural networks, which is extremely costly when the dataset is large. To accelerate the hyperparameter tuning, we may use more effective strategies such as Bayesian optimization \citep{snoek2012practical} or other automated machine learning (AutoML) techniques \citep{hutter2019automated} instead of grid search. Although some works, such as \citep{jarrett2022hyperimpute}, have used AutoML to tune the hyperparameters of imputation methods, the related studies are limited, and there is a large room for improvement. Future work may consider the following points:
\begin{itemize}
    \item Develop hyperparameter-free or hyperparameter-light imputation methods
    \item Develop more advanced nonparametric imputation methods
    \item Establish AutoML specifically tailored for missing data imputation
    \item Use domain knowledge or historical datasets to accelerate model selection and hyperparameter tuning
\end{itemize}

\subsection{Privacy Protection in Missing Data Imputation}

Data privacy is the cornerstone of trustworthy data science: without it, individuals can be re-identified from supposedly anonymized models, exposing medical records, financial details, or location histories. For missing data imputation, we also need to consider the problems of data leakage and privacy protection. Privacy-preserving techniques like differential privacy \citep{dwork2006differential} and federated learning \citep{kairouz2021advances} enable organizations to unlock collective insights across silos—think hospitals training shared cancer-detection models without ever moving patient data—turning privacy from a legal checkbox into a competitive advantage. In federated-learning scenarios, the training data are distributed across multiple clients, and any sharing or transmission between clients—or between a client and the central server, if one exists—is forbidden, thereby protecting data privacy to some extent. This protection is often further strengthened by differential privacy, with a trade-off between accuracy and privacy.

Recently, there have been a few studies considering federated learning for missing data imputation \citep{10.14778/3503585.3503598,abbasi2023fast,lian2025federated,fang2025msfgan,10.1016/j.eswa.2025.126543}, but there are still a few challenges. 
\begin{itemize}
    \item It is difficult to extend nonparametric imputation methods such as kNN and tree-based methods to federated learning scenarios \citep{zhang2024fedknn,Li_Wen_He_2020}. For example, regarding kNN, we cannot directly compute the distance between two data points located on different clients. 
    \item The theoretical analysis for the trade-off between accuracy and privacy of federated imputation methods, especially when the missing mechanism is not MCAR, is limited. 
    \item In some scenarios, the attributes are distributed across different clients, and the correspondence of their data indices is only partially known or even completely unknown, posing a major challenge for missing-data imputation.
\end{itemize}
Subsequent research is encouraged to systematically tackle these open issues.

\subsection{Extensive and Fair Numerical Comparison of Missing Data Imputation Methods}
As highlighted in Section \ref{sec_bench}, the current benchmarks for missing-data imputation are inadequate. The studies surveyed in Table \ref{tab_bench} either evaluate only a handful of imputation techniques or rely on a very limited number of datasets. Furthermore, several methods have never been subjected to systematic numerical comparison. For example, nearly every paper that proposes a new imputation technique omits any comparison with state-of-the-art matrix-completion algorithms.
To address these shortcomings, we propose the following two directions:
\begin{itemize}
\item Conduct a large-scale empirical study that compares more than 30 imputation algorithms on over 100 datasets drawn from a wide range of domains. The evaluation must cover multiple missingness mechanisms, varying missing-rate regimes, and extensive hyper-parameter tuning for every method.
\item Assess the utility of sophisticated imputation techniques for downstream analyses, benchmarking them against end-to-end learning approaches that operate directly on incomplete data.
\end{itemize}

\subsection{Pretrained and Universal Models for Missing Data Imputation}
In recent years, pretrained models, such as BERT \citep{devlin2019bert}, CLIP \citep{radford2021learning}, and GPTs \citep{achiam2023gpt}, have become the centerpiece of AI research, delivering breakthrough performance on a wide spectrum of real-world problems. A pretrained model is a large neural network whose parameters have already been learned from massive, general-domain data—text, images, audio, protein sequences, or code—so it arrives packed with broad patterns and world knowledge; instead of training from scratch, practitioners simply fine-tune or prompt it with a handful of examples, slashing data, compute, and time by orders of magnitude. It is natural to consider using pretrained models in missing data imputation or even establishing a universal model for missing data imputation. Here, a universal model for missing data imputation refers to one that, once well trained, can impute missing values for any dataset—or at least a broad class of datasets from different domains—without requiring further tuning or with only light fine-tuning. 

As discussed in Section \ref{sec_LLM}, a few scholars have explored the use of LLMs to assist in missing data imputation \citep{huang2024missing,hayat2024context,fang2025spatiotemporal} and achieved some inspiring results in general missing data imputation tasks. However, there are still limitations:
\begin{itemize}
    \item The performance improvements given by LLM-assisted imputation methods over conventional imputation methods are not significant or haven't been sufficiently verified. It is necessary to systematically compare LLM-assisted imputation methods with state-of-the-art imputation methods on diverse datasets.
    \item Since existing methods heavily rely on the textualization strategy for the tabular data or other data types, it is important to invent more elegant textualization strategies. It is also interesting to use LLM to improve missing data imputation without textualization.
    \item A versatile or universal model for missing data imputation does not necessarily rely on LLMs. It is possible to invent new neural network architectures for missing data imputation. 
\end{itemize}

\section{Conclusion}\label{sec_conclude}

Missing data is a pervasive challenge across a wide spectrum of scientific and engineering disciplines, from social science and healthcare to bioinformatics, e-commerce, and industrial monitoring. This review has provided a comprehensive and interdisciplinary overview (involving more than 500 references) of the theories, methodologies, and applications of missing data imputation, aiming to bridge the gaps between different research communities and foster cross-domain innovation. 

We began by introducing the fundamental concepts of missing data, including the mechanisms of missingness—MCAR, MAR, and MNAR—and the distinction between single and multiple imputation. We also discussed the diverse goals of imputation, whether as a preprocessing step, a final objective (as in recommendation systems or image inpainting), or a cost-reduction strategy. The review systematically categorized and examined a broad range of imputation methods, from classical approaches such as mean imputation, regression, hot-deck, and likelihood-based methods, to modern techniques including low-rank and high-rank matrix completion, tensor completion, and a variety of deep learning models. Notably, we highlighted the growing role of autoencoders, generative adversarial networks, normalizing flows, diffusion models, and graph neural networks in handling complex and high-dimensional incomplete data. More recently, the emergence of large language models has opened new avenues for semantic-aware imputation, particularly for categorical and mixed-type data. Furthermore, we explored how missing data imputation interacts with downstream tasks such as classification, clustering, and anomaly detection, emphasizing the importance of integrated or end-to-end approaches that jointly optimize imputation and task performance. Theoretical guarantees—including parameter estimation consistency, exact recovery conditions, and generalization bounds—were also reviewed to provide a rigorous foundation for evaluating imputation methods.

Despite significant advances, several challenges remain. Model selection and hyperparameter tuning are often cumbersome and dataset-specific, necessitating more automated and adaptive solutions. Privacy concerns in sensitive domains like healthcare call for the integration of federated learning and differential privacy into imputation pipelines. Moreover, while pretrained and universal models (including LLMs) show promise for general-purpose imputation, their performance gains over traditional methods require further validation, and more efficient adaptation strategies are needed. Looking forward, we anticipate that future research will focus on developing more robust, scalable, and interpretable imputation systems that can handle diverse data types and missingness patterns across domains. The integration of domain knowledge, the advancement of federated and privacy-preserving imputation, and the pursuit of universal imputation models represent exciting directions that will further solidify the role of missing data imputation as a cornerstone of reliable data science and artificial intelligence.

\section*{Acknowledgments}
This work was supported by the National Natural Science Foundation of China under Grant No.62376236 and the General Program of Natural Science Foundation of Guangdong
Province under Grant No.2024A1515011771.

I sincerely thank the researchers in all areas covered by this review. I have striven to be comprehensive, but I extend my apologies to any authors whose valuable work I may have inadvertently overlooked.



\bibliographystyle{plainnat}
\bibliography{MDIRef}

\end{document}